\documentclass[10pt,twocolumn,letterpaper]{article}

\usepackage{cvpr}              

\usepackage[dvipsnames]{xcolor}

\definecolor{cvprblue}{rgb}{0.21,0.49,0.74}
\usepackage[pagebackref,breaklinks,colorlinks,citecolor=cvprblue]{hyperref}
\usepackage[accsupp]{axessibility}
\usepackage{makecell}
\usepackage{multirow}
\usepackage{colortbl}
\usepackage{utfsym}

\definecolor{mygray}{gray}{.9}
\definecolor{mypink}{rgb}{.99,.91,.95}
\definecolor{mycyan}{cmyk}{.3,0,0,0}

\def\paperID{4707} 
\def\confName{CVPR}
\def\confYear{2024}

\title{Rethinking Diffusion Model for Multi-Contrast MRI Super-Resolution}

\author{Guangyuan Li, Chen Rao, Juncheng Mo, Zhanjie Zhang, Wei Xing$^{*}$, Lei Zhao$^{*}$ \\
College of Computer Science and Technology, Zhejiang University, China\\
{\tt\small $\{$cslgy, raochen, csmjc, cszzj, wxing, cszhl$\}$@zju.edu.cn}
}

\begin{document}
\maketitle
\footnotetext{* Corresponding author}
\footnotetext{$^1$ Code:  \url{https://github.com/GuangYuanKK/DiffMSR}}
\begin{abstract}
Recently, diffusion models (DM) have been applied in magnetic resonance imaging (MRI) super-resolution (SR) reconstruction, exhibiting impressive performance, especially with regard to detailed reconstruction. However, the current DM-based SR reconstruction methods still face the following issues: (1) They require a large number of iterations to reconstruct the final image, which is inefficient and consumes a significant amount of computational resources. (2) The results reconstructed by these methods are often misaligned with the real high-resolution images, leading to remarkable distortion in the reconstructed MR images. To address the aforementioned issues, we propose an efficient diffusion model for multi-contrast MRI SR, named as DiffMSR. Specifically, we apply DM in a highly compact low-dimensional latent space to generate prior knowledge with high-frequency detail information. The highly compact latent space ensures that DM requires only a few simple iterations to produce accurate prior knowledge. In addition, we design the Prior-Guide Large Window Transformer (PLWformer) as the decoder for DM, which can extend the receptive field while fully utilizing the prior knowledge generated by DM to ensure that the reconstructed MR image remains undistorted. Extensive experiments on public and clinical datasets demonstrate that our DiffMSR$^1$ outperforms state-of-the-art methods.
\end{abstract}    
\begin{figure}[t!]
  \centering
   \includegraphics[width=0.9\linewidth]{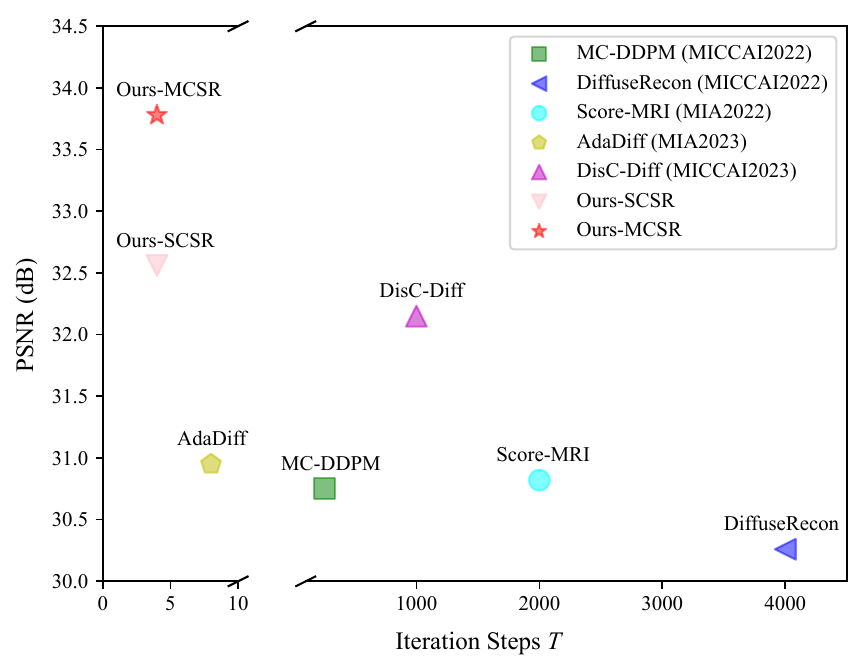}
   \caption{Comparison with DM-based MRI reconstruction methods on FastMRI dataset. Note that the experimental settings used by these methods are the same as those in Sec. \ref{exper}. As can be seen, our method has the best reconstruction metric and it only requires 4 iteration steps. Note that DiffuseRecon \cite{peng2022towards}, MC-DDPM \cite{xie2022measurement}, Score-MRI \cite{chung2022score}, and AdaDiff \cite{gungor2023adaptive} are employed for single-contrast SR (SCSR) reconstruction. DisC-Diff \cite{mao2023disc} is specifically designed for multi-contrast SR (MCSR) reconstruction.}
   \label{}
\end{figure}

\section{Introduction}
\label{sec:intro}
Magnetic resonance imaging (MRI) is a widely utilized in clinical imaging technology as it can provide clear information on tissue structure and function while being non-invasive and radiation-free.
However, obtaining high-resolution (HR) magnetic resonance (MR) images is challenging due to acquisition limitations \cite{Li2021HighResolutionPM, feng2021brain}.
Super-resolution (SR) technology can handle this challenge by reconstructing low-resolution (LR) images into their corresponding high-resolution (HR) version.
Traditional methods typically utilized model-based \cite{park2003super,manjon2010non} and learning-based \cite{yang2010image,zeyde2012single} SR reconstruction approaches. Nonetheless, these methods have insufficient reconstruction capabilities under high upsampling factors due to the complicated anatomical structures in MR images\cite{li2023rethinking}.

%
Following the advent of deep learning, convolutional neural networks (CNNs) have found extensive application in MRI SR reconstruction tasks \cite{Lyu2019Multi, Feng2021Multi, zhang2021mr, kang2022super, shin2023improving}.
Besides, Transformer-based approaches \cite{Li2022Trans, Li2022Wav, forigua2022superformer, lyu2023region, lyu2023multicontrast, huang2023accurate, li2023rethinking} are introduced as an alternative to CNNs for modeling long-range dependencies in MR images.
%
%
While these approaches enhance the performance of SR reconstruction, they tend to yield images that lack detail \cite{saharia2022image}, and for MR images with complicated anatomical structures, they fail to reconstruct some high-frequency details satisfactorily.
Apart from CNN- and Transformer-based methods, 
deep generative models like generative adversarial networks (GANs) \cite{goodfellow2014generative} provide different strategies for creating intricate details.
Very recently, diffusion models (DMs) \cite{ho2020denoising, gong2023semi} have shown impressive performance in MR image synthesis \cite{jiang2023cola, muller2023multimodal, ozbey2023unsupervised, yoon2023sadm}, reconstruction tasks \cite{peng2022towards, xie2022measurement, gungor2023adaptive, mao2023disc, wang2023inversesr}.
DMs generate high-fidelity images through a stochastic iterative denoising process employing a pure white Gaussian noise. In contrast to GANs, DMs yield a more accurate target distribution without facing optimization instability or mode collapse issues.

However, 
DM-based approaches encounter several issues when applied to MRI reconstruction.
On the one hand, DMs are required to perform a large number of iteration steps to generate samples, aiming to simulate the accurate details of the data. However, unlike MRI synthesis tasks that generate every pixel from scratch, MRI SR reconstruction tasks only require adding details on the given LR MR images.
%
Therefore, in theory, the SR reconstruction tasks only require a small number of iterations to generate satisfactory results, which will greatly save computational resources \cite{xia2023diffir, chen2023hierarchical}.
%
%
On the other hand, DMs tend to introduce artifacts in the generated MR images that are not present in the original HR images. Furthermore, some details of complicated anatomical structures may also be misaligned with the real target, resulting in image distortion.

These challenges lead us to rethink DM for MRI super-resolution reconstruction from a new perspective. 
%
First, we consider compressing the latent features in the DM into a low-dimensional latent space to reduce computational complexity and the number of iteration steps.
Second, given the advantage of the Transformer in capturing long-range dependencies, we integrated DM and Transformers to solve the distortion problem of DM.
Third, MRI can acquire images with different contrasts but the same anatomical structure by adjusting scanning parameters. Numerous studies \cite{Lyu2019Multi, Feng2021Multi, Li2022Trans, Li2022Wav, lyu2023multicontrast, huang2023accurate,li2023dudoinet,li2023rethinking, mao2023disc} have demonstrated that utilizing HR T1 contrast images with shorter scan times can offer valuable supplementary information for LR T2 contrast images with longer scan times. Therefore, we integrate HR T1 contrast images as reference images into the network.

Based on the above motivations, we propose an efficient diffusion model for multi-contrast MRI SR reconstruction, which it integrates the Prior-Guide Large Window Transformer (PLWformer) as the decoder. We call it \emph{DiffMSR}.
PLWformer can benefit from large window self-attention while having less computational burden, effectively enhancing the performance of DM in terms of generating accurate details.
Following previous practice \cite{xia2023diffir, chen2023hierarchical,esser2021taming}, we divide the training process into two stages to achieve prior extraction and training of the DM.
In the first stage, we employ the Prior Extraction (PE) module to compress the original HR MR image into highly compact latent features as prior knowledge, which is utilized to guide the PLWformer.
To fully utilize the prior knowledge, we design Prior-Guide Large Window Multi-Head Self-Attention (PL-MSA) and Prior-Guide Feed-Forward Network (PG-FFN) in PLWformer to utilize the compressed prior knowledge while expanding the attention receptive field.
We jointly optimize the PE and PLWformer to effectively obtain reliable prior knowledge.
In the second stage, we employ the PE module trained in the first stage to train the DM, allowing it to generate latent prior features in the latent space from Gaussian noise to guide the PLWformer for accurate reconstruction. Since the latent prior features are low-dimensional, the DM can estimate accurate details with simple iterations.

The main contributions are summarized as follows:

(1) We propose an efficient diffusion model for multi-contrast MRI SR, named as DiffMSR. Our method leverages diffusion models to generate effective prior knowledge, which is integrated into the SR process to reconstruct more satisfactory MR images with accurate details.

(2) We design the PLWformer with the aim of fully leveraging the prior knowledge generated by DM and having less computational burden while expanding the receptive field, which ensures that the reconstructed MR image is undistorted.

(3) Extensive experiments conducted on public and clinical datasets demonstrate the superior performance of the DiffMSR in comparison to state-of-the-art methods.

\begin{figure*}[t!]
  \centering
   \includegraphics[width=\linewidth]{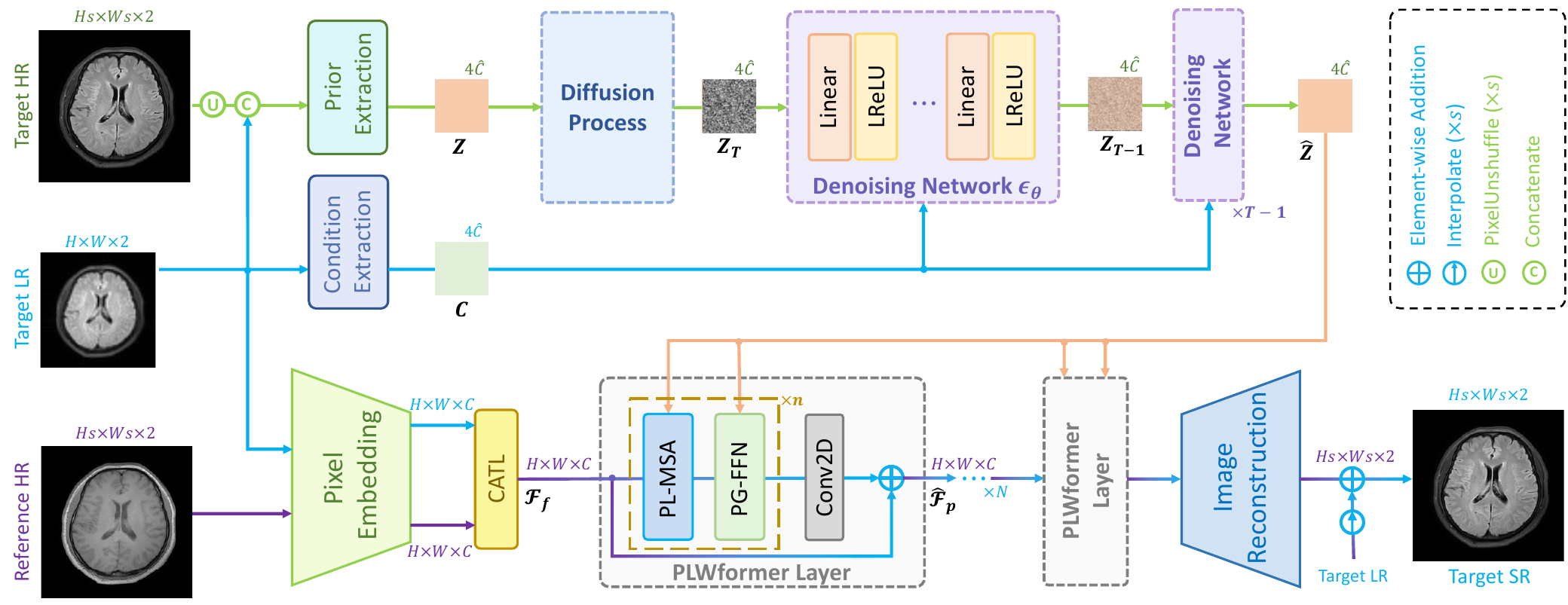}
   \caption{The overall architecture of our proposed DiffMSR, which mainly divided into two parts: (1) Diffusion Model; (2) Prior-guide Large Window Transformer (PLWformer) including $N$ PLWformer layers and a image reconstruction module. CATL: Cross-Attention Transformer Layer.}
   \label{fig_main}
\end{figure*}

\section{Related Works}
\subsection{Multi-Contrast MRI SR}
Multi-contrast MRI super-resolution (MCSR) involves using multiple contrasts of MR images in the SR process. Specifically, MCSR utilizes T1 contrast images with shorter repetition time and echo time as reference images to provide valuable high-frequency information during SR for T2 contrast images with longer scan times.
For instance, 
Li \etal \cite{Li2022Trans} were the first to introduce Transformer in MCSR tasks and proposed the Transformer-empowered multi-scale contextual matching and aggregation network.
Lyu \etal \cite{lyu2023multicontrast} introduced a texture-preserving branch and a contrastive constraint in MCSR.
Lei \etal \cite{lei2023decomposition} proposed a decomposition-based variational network for MCSR task.
Li \etal \cite{li2023rethinking} designed the reference-aware implicit attention to extend MCSR to arbitrary scale upsampling.
Mao \etal \cite{mao2023disc} proposed a disentangled conditional diffusion model for MCSR. 
The majority of the aforementioned MCSR methods primarily employ Transformer-based models, while few methods employ diffusion models to perform MCSR tasks.

\subsection{MRI Diffusion Model}
Diffusion models (DM) \cite{ho2020denoising}, as a type of probabilistic generative model, can construct the required data samples from Gaussian noise through a stochastic iterative denoising process. DM has exhibited impressive performance in MRI reconstruction tasks \cite{xie2022measurement, peng2022towards, chung2022score, gungor2023adaptive, mao2023disc}.
For instance,
Xie \etal \cite{xie2022measurement} proposed a measurement-conditioned denoising diffusion probabilistic model for under-sampled medical image reconstruction.
Peng \etal \cite{peng2022towards} introduced a novel diffusion model-based MR reconstruction method.
Chung \etal \cite{chung2022score} designed a score-based diffusion models for accelerated MRI. 
Mao \etal \cite{mao2023disc} designed a disentangled conditional diffusion model for multi-contrast brain MRI SR. 
Nonetheless, the above DM-based methods require a significant number of iteration steps to generate samples, which costs many computing resources. Additionally, they tend to introduce artifacts into the generated MR SR images that are not present in the original HR images. 
To address the above issues, we propose to combine DM and Transformer to reconstruct artifact-free and non-distorted MR images.

\subsection{MRI Transformer}
The Transformer architecture is widely applied in MCSR \cite{Li2022Trans, Li2022Wav, lyu2023multicontrast, li2023rethinking} as it can model long-range dependencies effectively, allowing for the reconstruction of complicated anatomical structures in MR images.
However, due to computational burdens, these methods typically limit the window size to 8$\times$8. Expanding the window size can effectively increase the receptive field, capture longer-range dependencies, and enhance the quality of reconstructed images.
Inspired by \cite{zhou2023srformer}, to maintain a smaller computational burden while expanding the window size, we design a prior-guide large window transformer (PLWformer). 
PLWformer can utilize the prior knowledge generated by DM while possessing a larger receptive field to better reconstruct details.


\section{Methodology}
\subsection{Overall Architecture}
The overall architecture of the proposed DiffMSR is shown in Figure \ref{fig_main}. As can be seen, DiffMSR mainly comprises two components: the diffusion model (DM) and the prior-guide large window transformer (PLWformer). We design DiffMSR with the goal of integrating DM and Transformer to overcome the shortcomings of DM and achieve better reconstruction results.
Specifically, DM is employed to generate highly compact latent features as prior knowledge, which includes detailed information from real MR images, guiding the PLWformer for reconstruction to produce artifact-free and non-distorted MR images. Meanwhile, PLWformer utilizes large window attention to model longer-range dependencies in MR images.

\begin{figure}[t!]
  \centering
   \includegraphics[width=\linewidth]{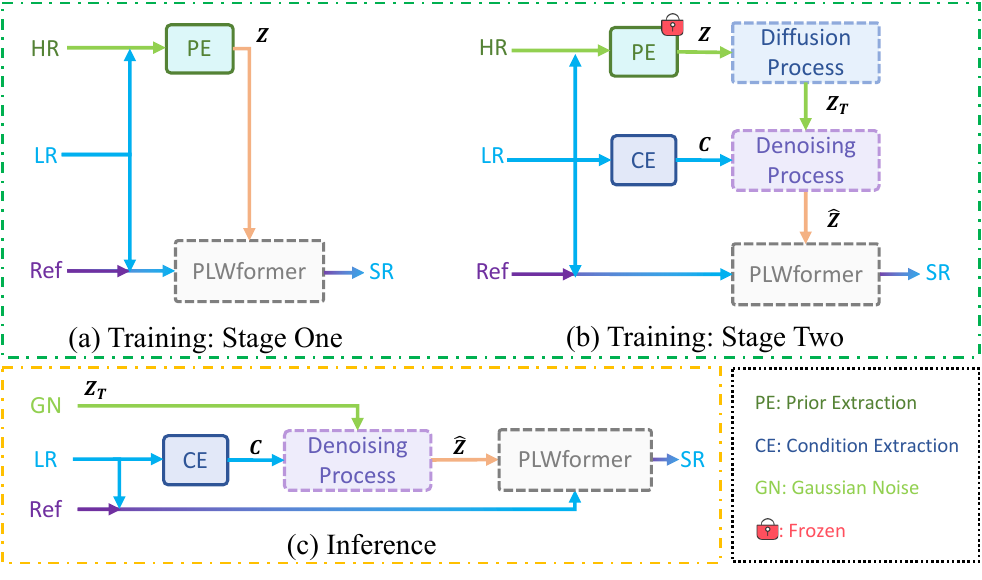}
   \caption{The process of training stage and inference stage.}
   \label{fig_T2I}
\end{figure}

We follow the previous practice \cite{xia2023diffir, chen2023hierarchical,esser2021taming} and split the training process into two stages, as shown in Figure \ref{fig_T2I}. In the first stage, we utilize the Prior Extraction (PE) module to compress the target HR image $I_{HR} \in \mathbb{R}^{H s \times W s \times 2}$ into a highly compact latent space, obtaining prior knowledge $Z \in \mathbb{R}^{4 \hat{C}}$, and employ it to guide the PLWformer for high-quality reconstruction.
Here, $H$, $W$, and $s$ represent the height, width, and upsampling scale. $\hat{C}$ denotes the latent feature channel number. We jointly optimize the PE and PLWformer to obtain more reliable prior knowledge $Z$.
In the second stage, we first employ the PE module trained in the first stage to generate the target sample $Z$. Then, we train DM to learn to generate prior knowledge $\hat{Z} \in \mathbb{R}^{4 \hat{C}}$ and subsequently utilize $\hat{Z}$ to enhance the reconstruction capabilities of the PLWformer.
Additionally, we employ Pixel Embedding and Cross-Attention Transformer Layer (CATL) \cite{Li2022Wav,lyu2022dudocaf} to fuse target LR image $I_{LR} \in \mathbb{R}^{H \times W \times 2}$ and reference HR image $I_{Ref} \in \mathbb{R}^{H s \times W s \times 2}$, obtaining $\mathcal{F}_f$ as input for the PLWformer.


\begin{figure*}[t!]
  \centering
   \includegraphics[width=\linewidth]{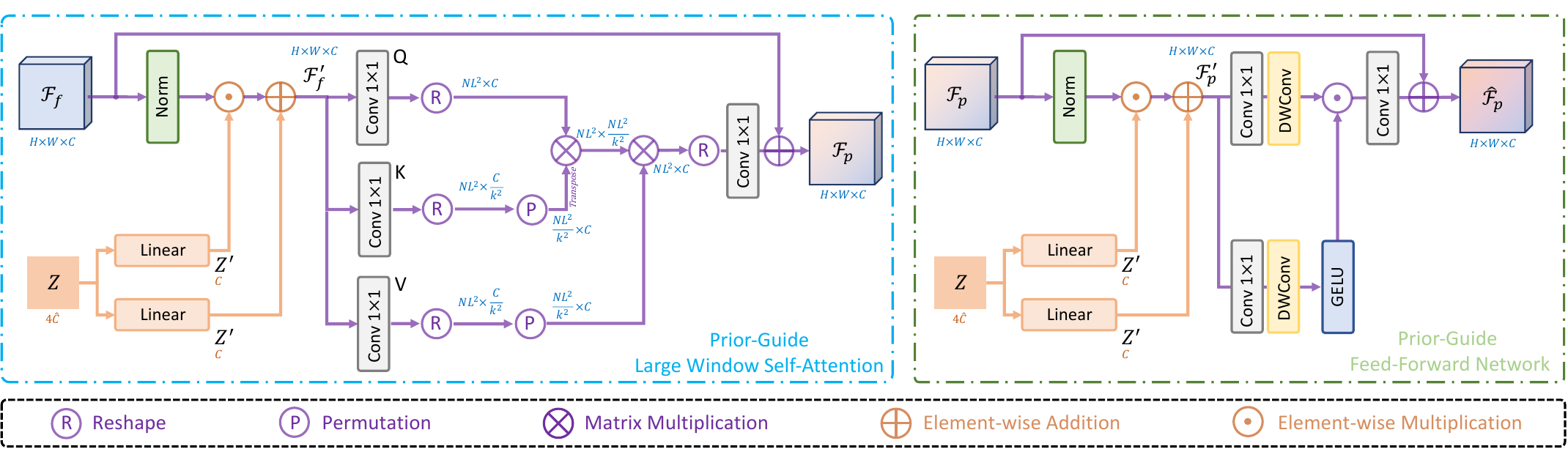}
   \caption{The architecture of prior-guide large window self-attention and prior-guide feed-forward network.}
   \label{fig_sub2}
\end{figure*}

\subsection{Training: Stage One} \label{S1}
The purpose of the first stage is to jointly train Prior Extraction (PE) and PLWformer to obtain more reliable prior knowledge $Z$, as shown in Figure \ref{fig_T2I}(a).
Hence, we primarily utilize two modules: PE and PLWformer. 
PE compresses HR images to obtain a compact representation $Z$, which serves as prior knowledge.
PLWformer employs $Z$ as a guide to model long-range dependencies using large window attention without increasing computational burden, thereby enhancing the accuracy of reconstructed details.

\noindent \textbf{Prior Extraction.}
The structure of PE is primarily composed of residual blocks and linear layers stacked together. For more detailed architecture, please see \emph{Supplementary Material}.
Given the target LR image $I_{LR}$ and its corresponding HR image $I_{HR}$, we first perform a PixelUnshuffle operation on $I_{HR}$, then concatenate them along the channel dimension and feed them into the PE to generate the prior knowledge $Z$:
\begin{equation}
Z=PE(Concat(PixelUnshuffle(I_{HR}), I_{LR})).
\end{equation}
Since $Z$ is a highly compact feature, this effectively reduces the computational burden for the subsequent DM.

\noindent \textbf{PLWformer.}
The vast majority of Transformer-based MCSR methods \cite{Li2022Trans, Li2022Wav, lyu2023multicontrast, li2023rethinking} employ multi-head self-attention (MSA) with a window size of 8$\times$8. Appropriately expanding the window size of MSA in Transformer can effectively enhance reconstruction performance, but this will result in higher computational burden\cite{zhou2023srformer}.
Inspired by \cite{zhou2023srformer}, we introduce the Prior-Guide Large Window Transformer (PLWformer) to enjoy the benefits of the large window while utilizing prior knowledge. PLWformer consists of two components, the Prior-Guide Large Window Multi-Head Self-Attention (PL-MSA) and the Prior-Guide Feed Forward Network (PG-FFN), as shown in Figure \ref{fig_main}.
Next, we will introduce how PLWformer employs prior knowledge Z as guidance and how to expand the window size without increasing the computational burden.

After obtaining the prior knowledge $Z$ through PE compression, it is input into PL-MSA and PG-FFN as dynamic modulation parameters to guide the reconstruction, as shown in Figure \ref{fig_sub2}:
\begin{equation}
\begin{aligned}
\mathcal{F}_f^{\prime} & =\mathbb{L}(Z) \odot \mathbb{N} \left(\mathcal{F}_f\right)+\mathbb{L}(Z), \\
\mathcal{F}_p^{\prime} & =\mathbb{L}(Z) \odot \mathbb{N} \left(\mathcal{F}_p\right)+\mathbb{L}(Z),
\end{aligned}
\end{equation}
where $\mathcal{F}_f^{\prime} \in \mathbb{R} ^{H \times W \times C}$, $\mathcal{F}_p^{\prime} \in \mathbb{R} ^{H \times W \times C}$, C means the channel dimension of the feature map, $\odot$ indicates element-wise multiplication, $\mathbb{N}$ denotes layer normalization, $\mathbb{L}$ represents linear layer.

In PL-MSA, $\mathcal{F}_f^{\prime}$ is first splitted into $N$ non-overlapping square windows $\mathcal{F}_f^{s\prime} \in \mathbb{R}^{NL^2 \times C}$, where $L$ is the length of each window.
Then, linear layers $\mathbb{L}_Q$, $\mathbb{L}_K$, and $\mathbb{L}_V$ are used to embed $\mathcal{F}_f^{s\prime}$ to obtain $Q \in \mathbb{R}^{NL^2 \times C}$, $K \in \mathbb{R}^{NL^2\times C/k^2}$, and $V \in \mathbb{R}^{NL^2\times C/k^2}$, where $k$ is the token reduction factor. Next, permutation is applied to $K$ and $V$, obtaining in permuted tokens $K_p \in \mathbb{R}^{NL^2/k^2\times C}$ and $V_p \in \mathbb{R}^{NL^2/k^2\times C}$.
This way, a factor of $k$ ($k$=2) reduces the window size for $K$ and $V$, but their channel dimension is still unchanged to ensure the expressiveness of the attention map generated by each attention head \cite{touvron2021going}. Finally, self-attention (SA) computation is performed:
\begin{equation}
SA(Q, K, V)=softmax(Q K_p'/{\sqrt{d_k}}+B) V_p,
\end{equation}
where $B$ is the aligned relative position embedding, $K_p'$ is the transpose matrix of $K_p$.
We reduce the channel dimensions of the $K$ and $V$ matrices and employ permutation operations to transfer some spatial information to the channel dimensions, which effectively reduces the computational burden in the case of large windows.

In PL-FFN, we first employ 1$\times$1 convolution to aggregate information from different channels. Then, 3$\times$3 depthwise convolution is utilized to aggregate information from spatially adjacent pixels, and a gating mechanism is employed to enhance information encoding:
\begin{equation}
\hat{\mathcal{F}}_p=Conv(GELU(\Phi(\mathcal{F}_p^{\prime})) \odot \Phi(\mathcal{F}_p^{\prime}))+\mathcal{F}_p,
\end{equation}
where $\Phi$ means 1$\times$1 convolution and 3$\times$3 depthwise convolution.

\noindent \textbf{Optimization function.}
To generate more reliable prior knowledge, we jointly optimize PE and PLWformer in the first training stage using an image-domain reconstruction loss $\mathcal{L}_{img}$ and frequency-domain data consistency loss $\mathcal{L}_{dc}$:
\begin{equation}
\mathcal{L}_{stage}^1=\lambda_{1}\mathcal{L}_{img}+\lambda_{2}\mathcal{L}_{dc},
\end{equation}
where 
$\mathcal{L}_{img} = \left\|I_{S R}-I_{H R}\right\|_1, 
\mathcal{L}_{dc} = \left\|K_{D C}-K_{H R}\right\|_2. $
$I_{SR}$ and $I_{HR}$ represent the reconstructed T2 image and original HR T2 image, respectively. $K_{DC}$ and $ K_{HR}$ are the frequency domain data after fidelity and the original frequency domain data, respectively. We set $\lambda_{1}$=1 and $\lambda_{2}$ =0.001 to balance the contributions of the two losses. 
%

\subsection{Training: Stage Two}\label{S2}
The purpose of the second stage is to train DM to learn how to generate prior knowledge consistent with the distribution of real MR images for guiding and enhancing the reconstruction process of the PLWformer, as shown in Figure \ref{fig_T2I}(b). DM includes the forward diffusion process and the reverse denoising process.

\noindent \textbf{Diffusion Process.}
We employ the PE trained in the first stage to capture the prior knowledge $Z$. After that, we apply the diffusion process on $Z$ to sample $Z_T \in \mathbb{R}^{4 \hat{C}}$, which can be described as:
\begin{equation}
q\left(\mathbf{Z}_T \mid \mathbf{Z}\right)=\mathcal{N}\left(\mathbf{Z}_T ; \sqrt{\bar{\alpha}_T} \mathbf{Z},\left(1-\bar{\alpha}_T\right) \mathbf{I}\right),
\end{equation}
where $T$ is the total number of iterations, $\mathcal{N}$ denotes the Gaussian distribution, $\alpha$ and $\bar{\alpha}_T$ are defined as:
$
\alpha=1-\beta_t, \quad \bar{\alpha}_t=\prod_{i=1}^t \alpha_i,
$
where $t=(1,\dots,T)$, $\beta_{1:T} \in (0,1)$ are hyperparameters that control the variance of the noise.

\noindent \textbf{Denoising Process.}
The reverse process is a Markov chain running backwards from $Z_T$ to $\hat{Z}$.
Taking the reverse step from $Z_t$ to $Z_{t-1}$ as an example:
\begin{equation}
q\left(\mathbf{Z}_{t-1} \mid \mathbf{Z}_t, \mathbf{Z}_0\right)=\mathcal{N}(\mathbf{Z}_{t-1} ; \boldsymbol{\mu}_t\left(\mathbf{Z}_t, \mathbf{Z}_0\right), \frac{1-\bar{\alpha}_{t-1}}{1-\bar{\alpha}_t} \beta_t \mathbf{I}),
\end{equation}
\begin{equation}
\boldsymbol{\mu}_t\left(\mathbf{Z}_t, \mathbf{Z}_0\right)=\frac{1}{\sqrt{\alpha_t}}(\mathbf{Z}_t-\frac{1-\alpha_t}{\sqrt{1-\bar{\alpha}_t}} \boldsymbol{\epsilon}),
\end{equation}
where $\boldsymbol{\epsilon}$ represents the noise in $Z_t$ and a denoising network $\boldsymbol{\epsilon}_\theta$ is employed to estimate the noise $\boldsymbol{\epsilon}$ for each step.
Inspired by \cite{rombach2022high}, we design a Condition Extraction (CE) module, as shown in Figure \ref{fig_main}. The structure of this module is consistent with PE, except that it only takes the target LR image $I_{LR}$ as input and outputs the conditional latent feature $C \in \mathbb{R}^{4 \hat{C}}$.
Therefore, the denoising network predicts noise conditioned on $Z_t$ and $C$:
\begin{equation}
\boldsymbol{Z}_{t-1}=\frac{1}{\sqrt{\alpha_t}}(\boldsymbol{y}_t-\frac{1-\alpha_t}{\sqrt{1-\bar{\alpha}_t}} \boldsymbol{\epsilon}_\theta\left(\mathbf{Z}_t, \mathbf{C}, t\right))+\sqrt{1-\alpha_t} \boldsymbol{\epsilon}_t,
\end{equation}
where $\boldsymbol{\epsilon}_t \sim \mathcal{N}(0, \mathbf{I})$. After $T$ iterations of sampling, the DM can generate the predicted prior knowledge $\hat{Z}$, and then it can be utilized to guide the PLWformer, as shown in Figure \ref{fig_main}.
Since the prior knowledge $Z$ is highly compact, DM in the second stage can employ fewer iterations to obtain considerably better estimations than traditional DMs \cite{ho2020denoising, rombach2022high, mao2023disc}.

\noindent \textbf{Optimization function.}
We employ $\mathcal{L}_{stage}^2$ to joint train CE, denoising network, and PLWformer:
\begin{equation}
\mathcal{L}_{stage}^2=\lambda_{1}\mathcal{L}_{img}+\lambda_{2}\mathcal{L}_{dc}+\mathcal{L}_{diff},
\end{equation}
where 
$\mathcal{L}_{diff}=\frac{1}{4 C^{\prime}} \sum_{i=1}^{4 C^{\prime}}|\hat{\mathbf{Z}}(i)-\mathbf{Z}(i)|$.

\subsection{Inference}\label{IS}
In the inference, CE is first used to compress the target LR image $I_{LR}$ into a conditional latent $C$.
Then, we randomly sample Gaussian noise $Z_T$. The denoising network employs $Z_T$ and $C$ to generate prior knowledge $\hat{Z}$ after $T$ iterations ($T$=4). $\hat{Z}$ is then utilized to guide PLWformer to reconstruct the final SR image, as shown in Figure \ref{fig_T2I}(c). Here, the PLWformer takes $I_{LR}$ and the reference HR image $I_{ref}$ as input and employs CAT to fuse $I_{LR}$ and $I_{ref}$ to make full use of the valuable supplementary information in $I_{ref}$.

\section{Experiments} \label{exper}
\subsection{Datasets and Baselines}
\noindent \textbf{Public Dataset.}
The public dataset employed is the FastMRI Knee \cite{zbontar2018fastmri}, where the reference contrast is PD, and the target contrast is FS-PD. We selected 1600 slices with a training, validation, and test set split ratio of 7:1:2.

\noindent \textbf{Clinical datasets.}
Clinical datasets include Brain (with T1 reference contrast and T2-FLAIR target contrast) and Pelvic (with T1 reference contrast and T2 target contrast). 
Specifically, the brain dataset consists of 637 slices from healthy subjects and 305 slices from tumor subjects. Among them, 512 slices of healthy subjects are utilized for training, 125 slices of healthy subjects are employed for validation and testing, and 305 slices of tumor subjects are used for additional testing. 
The pelvic dataset comprises 1600 slices, with a training, validation, and test split ratio of 7:1:2.
The raw clinical datasets are generated by scanning with a 3T Philips Ingenia MRI Scanner. The scanning parameters for the brain are TE (T1): 2.3ms, TE (T2-FLAIR): 120ms. The scanning parameters for the pelvic are TE (T1): shortest, TE (T2): 130ms. The sum-of-squares (SOS) method is used for coil combination. 

\noindent \textbf{Baselines.}
We compared our DiffMSR with several recent state-of-the-art methods, including MCSR \cite{Lyu2019Multi}, MINet \cite{Feng2021Multi}, MASA \cite{lu2021masa}, WavTrans \cite{Li2022Wav}, McMRSR \cite{Li2022Trans}, MC-VarNet \cite{lei2023decomposition} and DisC-Diff \cite{mao2023disc}.
Note that we focus on 4$\times$ upscaling reconstruction in our experiments.

\subsection{Implementation Details}
We implement our proposed approach in PyTorch \cite{paszke2019pytorch} with a single NVIDIA RTX4090 GPU. 
For PLWformer, we set the number of Transformer blocks as [6, 6, 6, 6], the attention heads as [4, 4, 4, 4], the number of channels $C$ as 64, the window size as 16$\times$16.
For the diffusion model, the channel dimension $\hat{C}$ is 64, the linear layers in the denoising network are set to 5, and the total time-step $T$ is 4.
The Adam \cite{kingma2014adam} optimizer is adopted for network training with iterations of 500K. We set the batch size as 4 and the learning rate is 2e-4 and decayed by factor 0.5 at [250K, 400K, 450K, 475K].
Furthermore, the complex data [$H \times W$] is divided into two channels, \emph{real} and \emph{imag}, \emph{i.e.}, [$H \times W\times 2$].

\begin{figure*} [h]
	\centering
	\captionsetup[subfloat]{labelformat=empty}
	\subfloat[\footnotesize MCSR \cite{Lyu2019Multi}]{
	\begin{minipage}[b]{0.1\textwidth}
    	\includegraphics[scale=0.17]{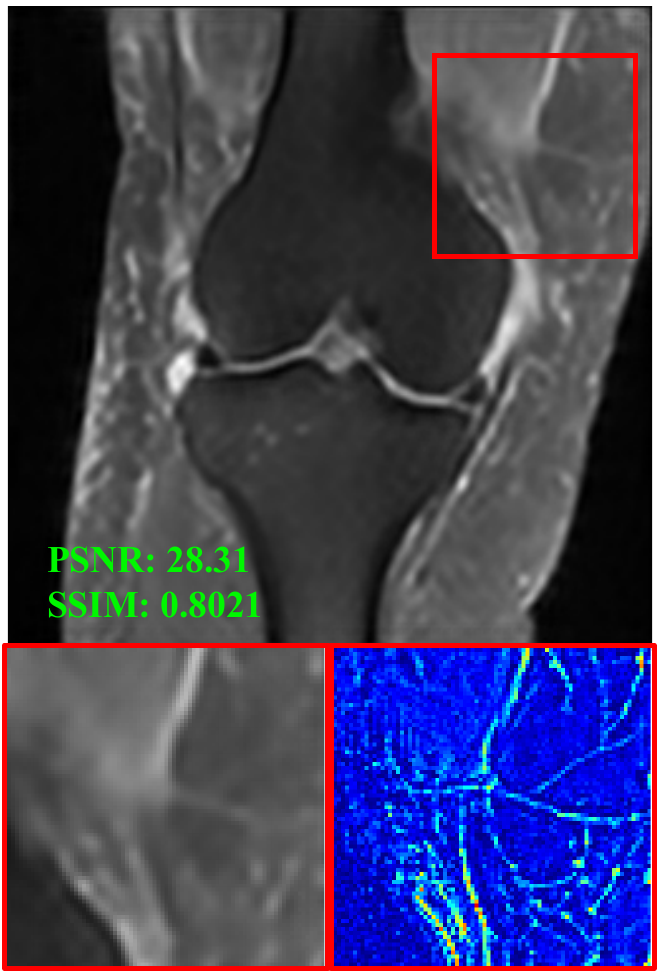} \\
        \includegraphics[scale=0.17]{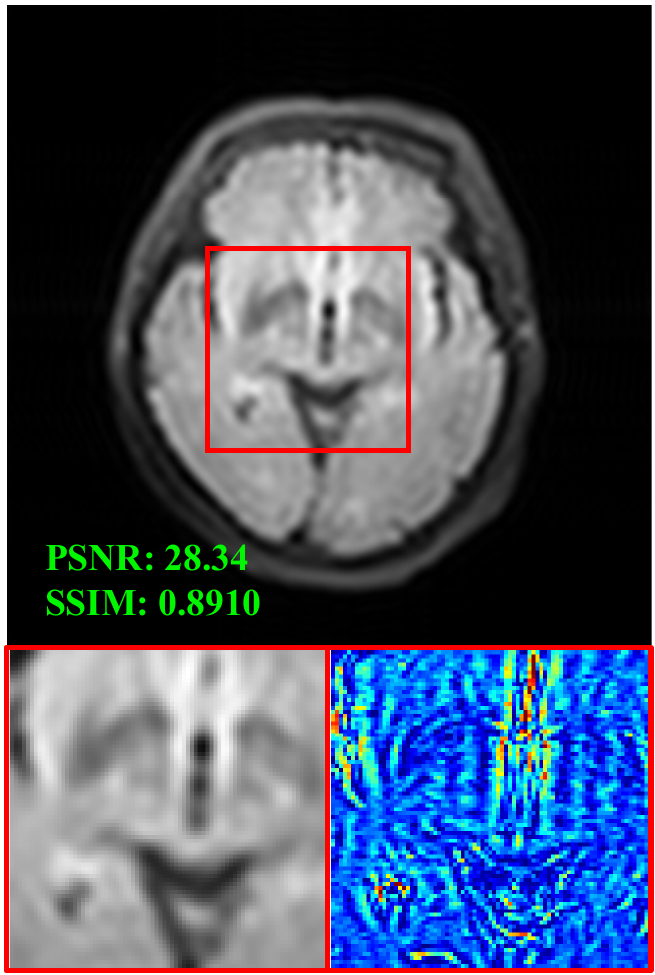}\\
        \includegraphics[scale=0.17]{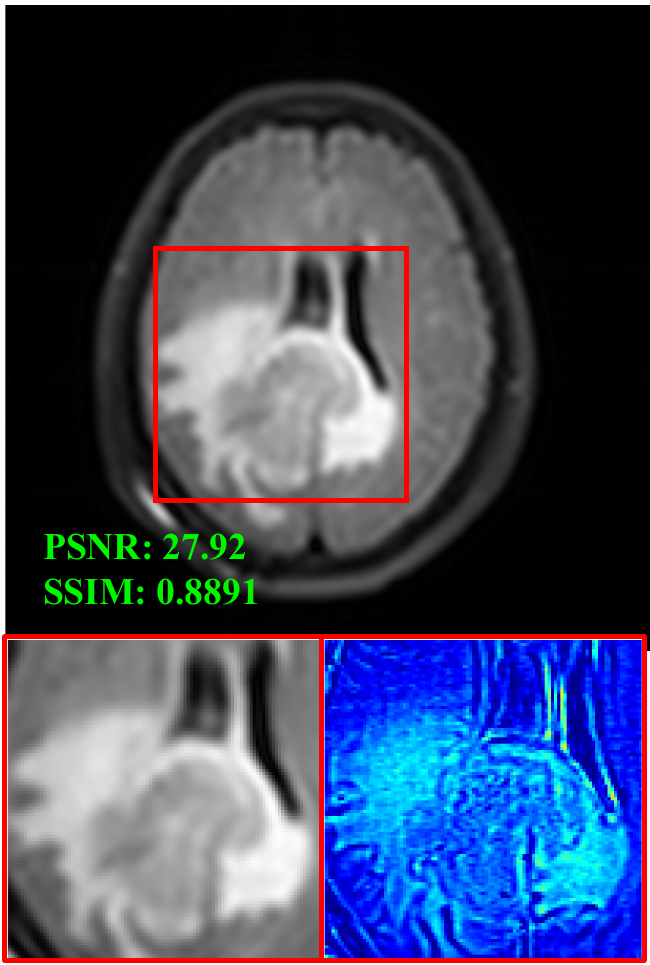}\\
        \includegraphics[scale=0.17]{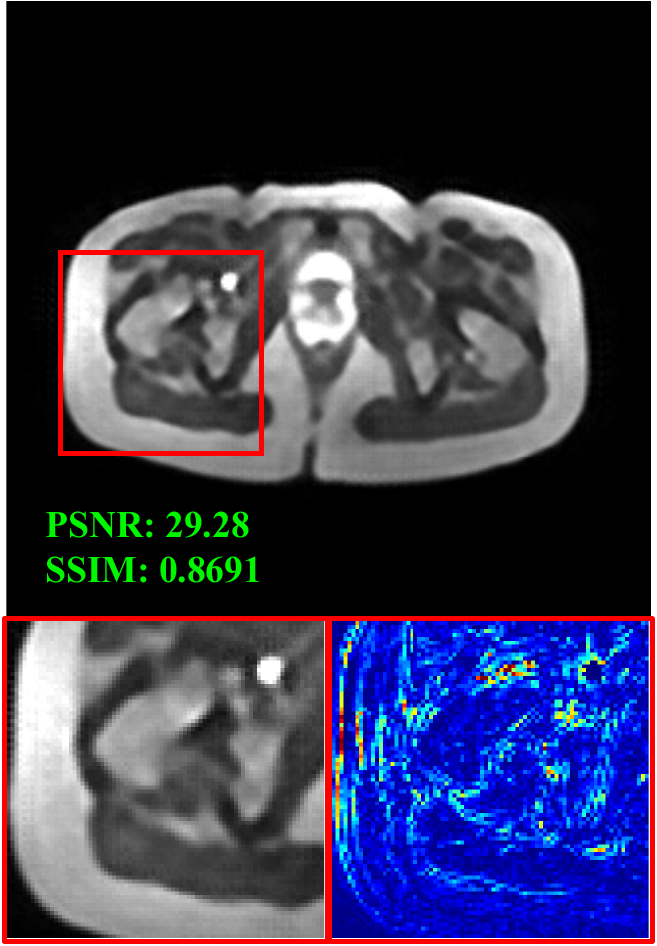}
	\end{minipage}
}
\hfill
	\subfloat[\footnotesize MINet \cite{Feng2021Multi}]{
	\begin{minipage}[b]{0.1\textwidth}
		\includegraphics[scale=0.17]{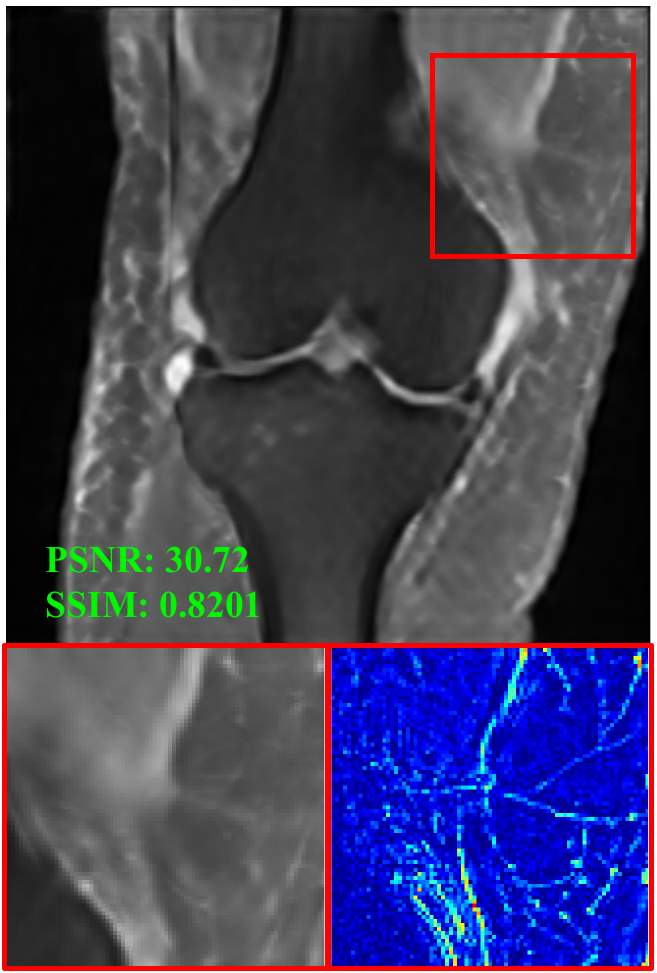} \\
        \includegraphics[scale=0.17]{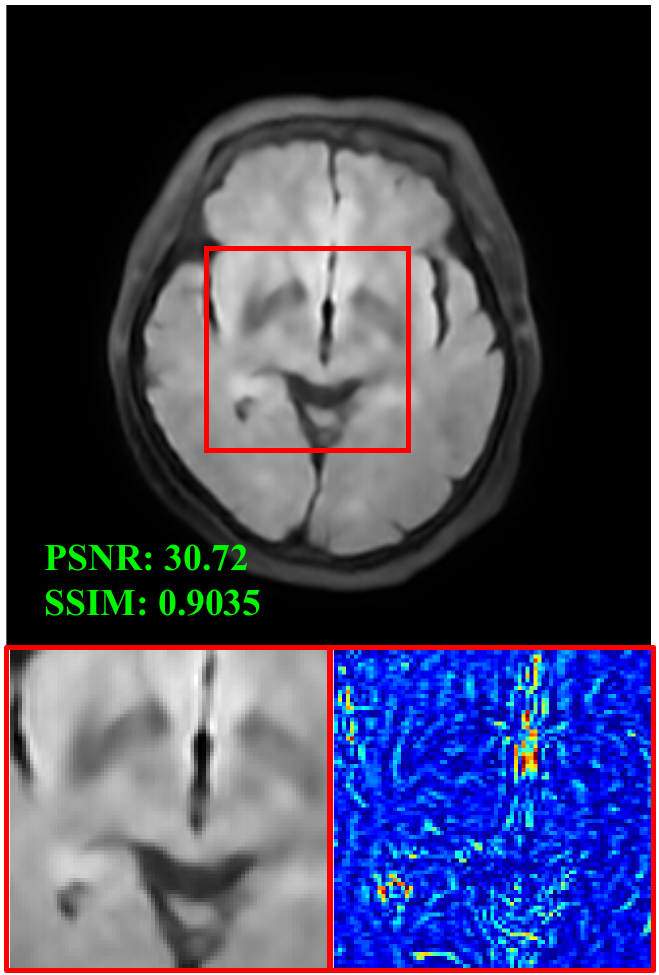}\\
        \includegraphics[scale=0.17]{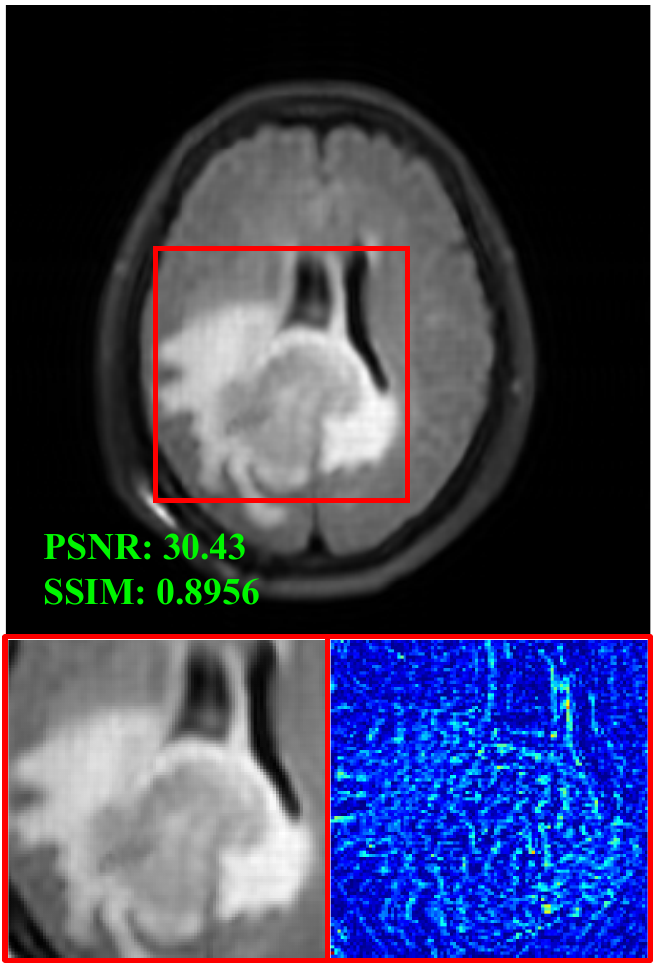}\\
        \includegraphics[scale=0.17]{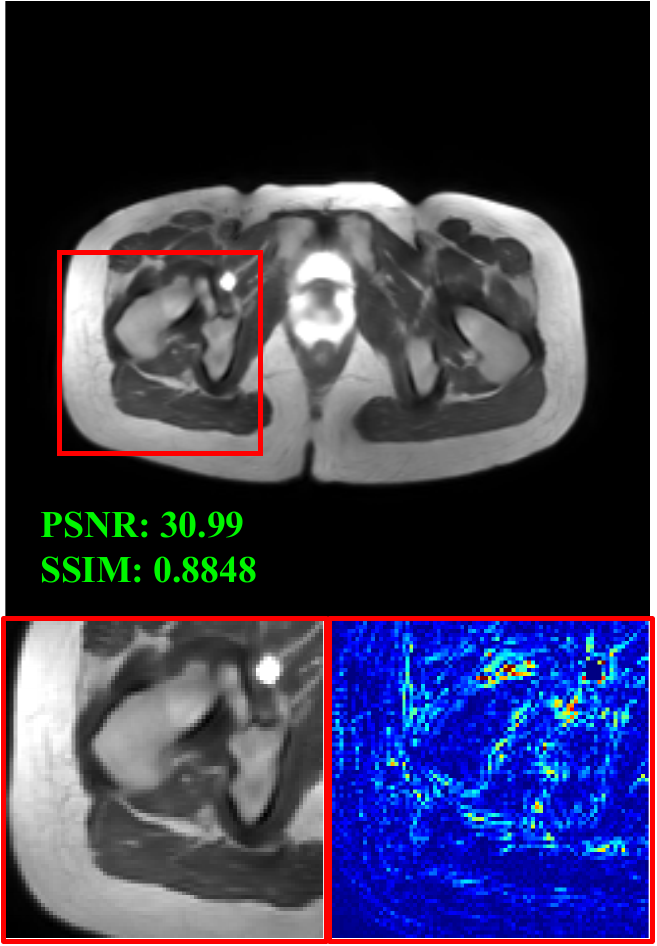}
	\end{minipage}
}
\hfill
	\subfloat[\footnotesize MASA \cite{lu2021masa}]{
	\begin{minipage}[b]{0.1\textwidth}
		\includegraphics[scale=0.17]{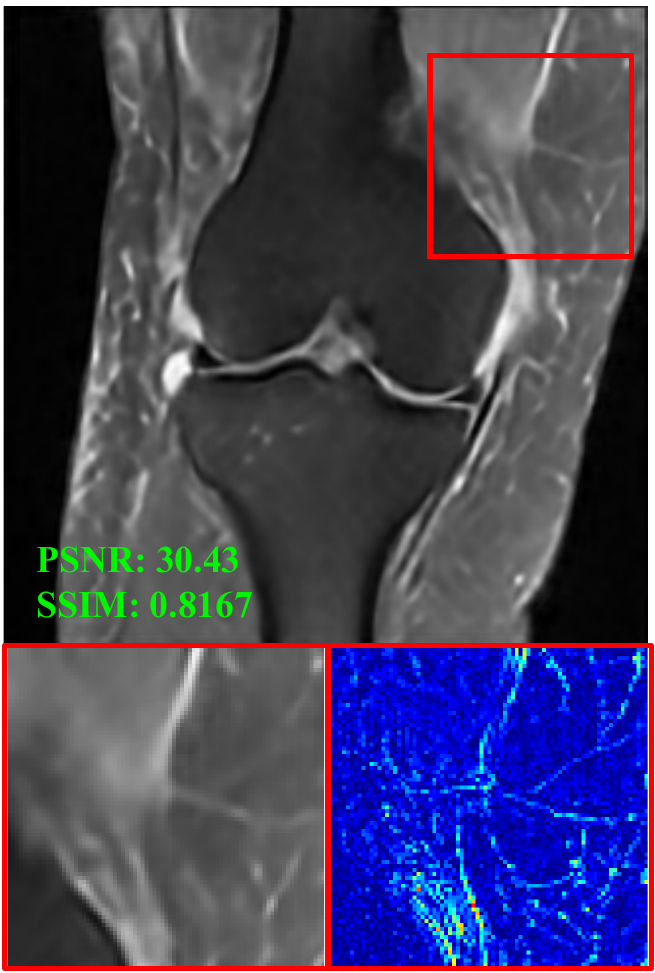} \\
        \includegraphics[scale=0.17]{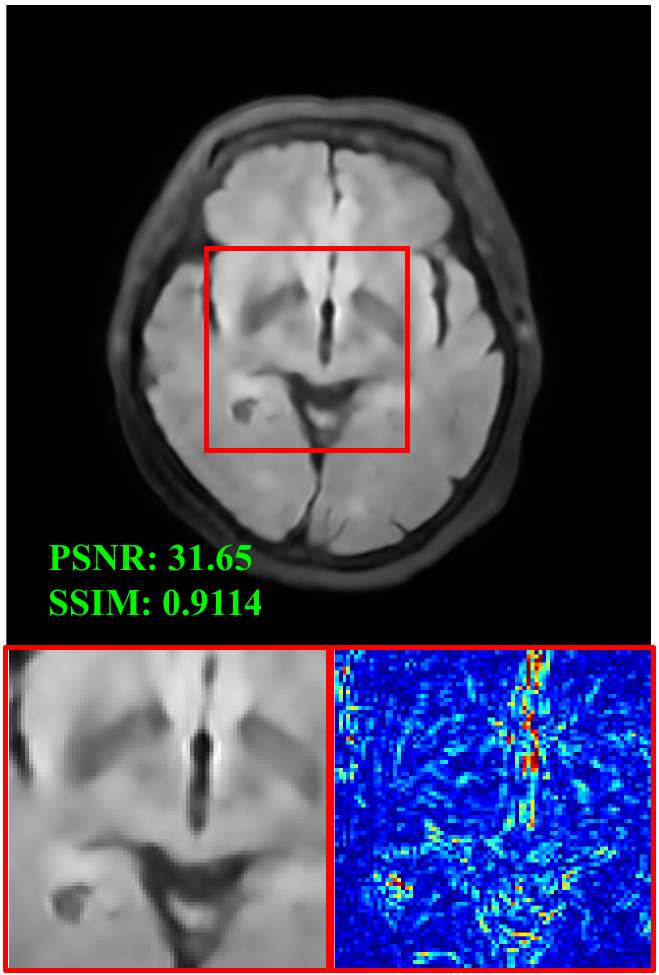}\\
        \includegraphics[scale=0.17]{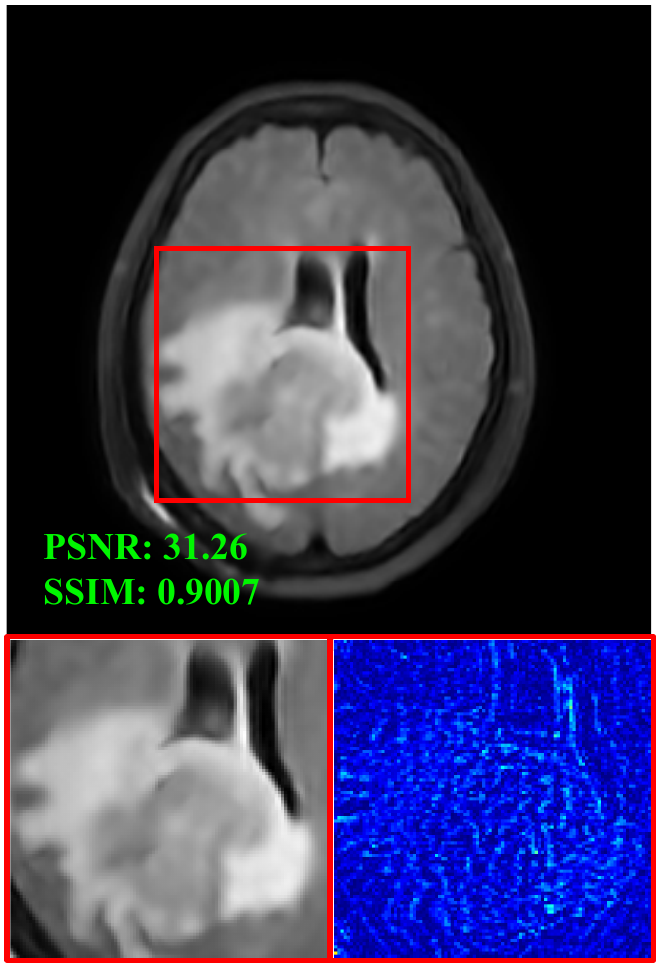}\\
        \includegraphics[scale=0.17]{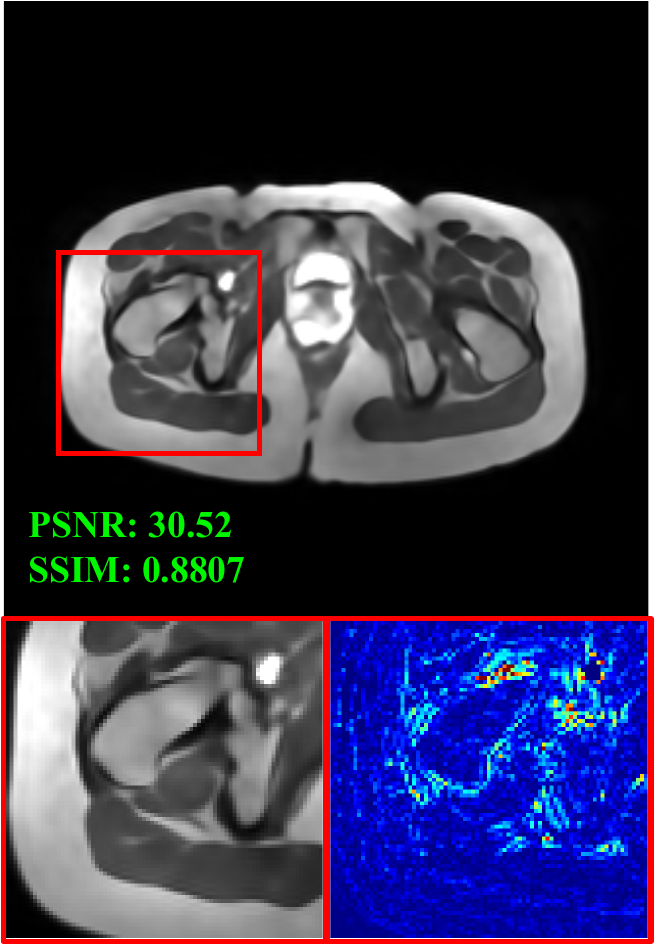}
	\end{minipage}
}
\hfill
    \subfloat[\footnotesize WavTrans \cite{Li2022Wav}]{
	\begin{minipage}[b]{0.1\textwidth}
		\includegraphics[scale=0.17]{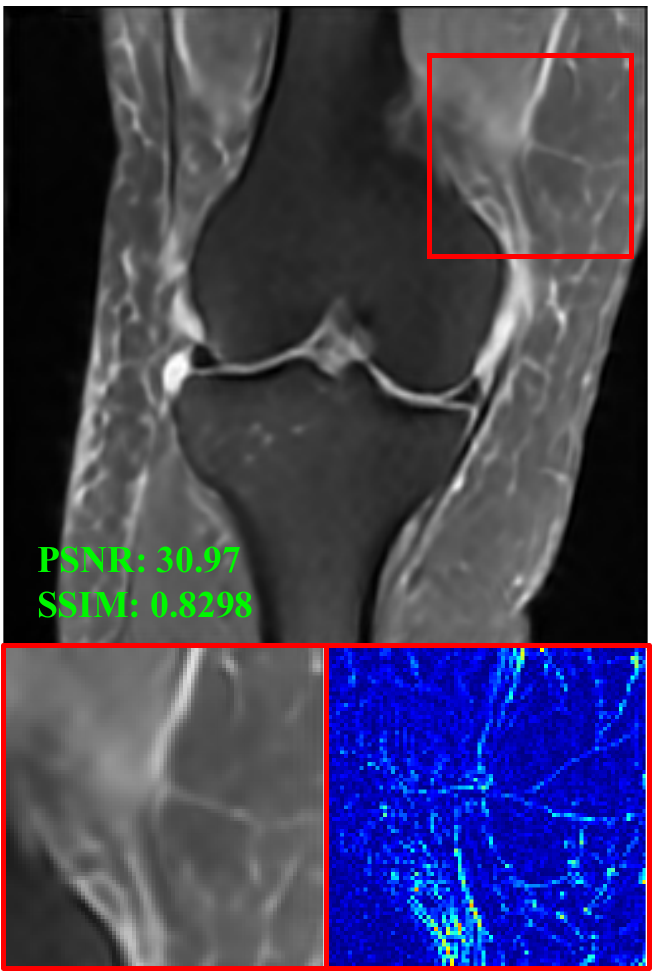} \\
        \includegraphics[scale=0.17]{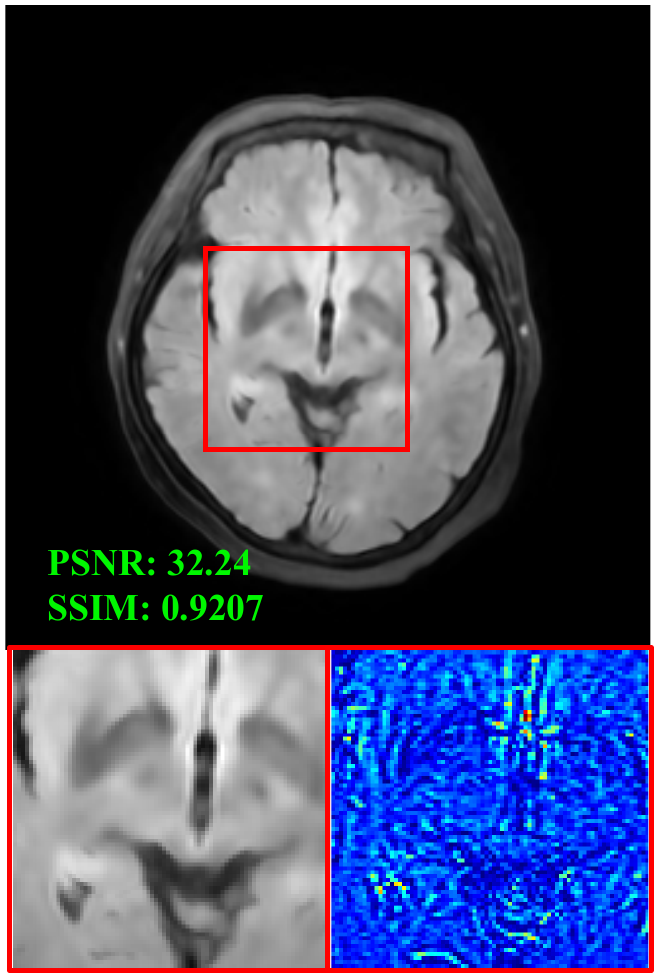}\\
        \includegraphics[scale=0.17]{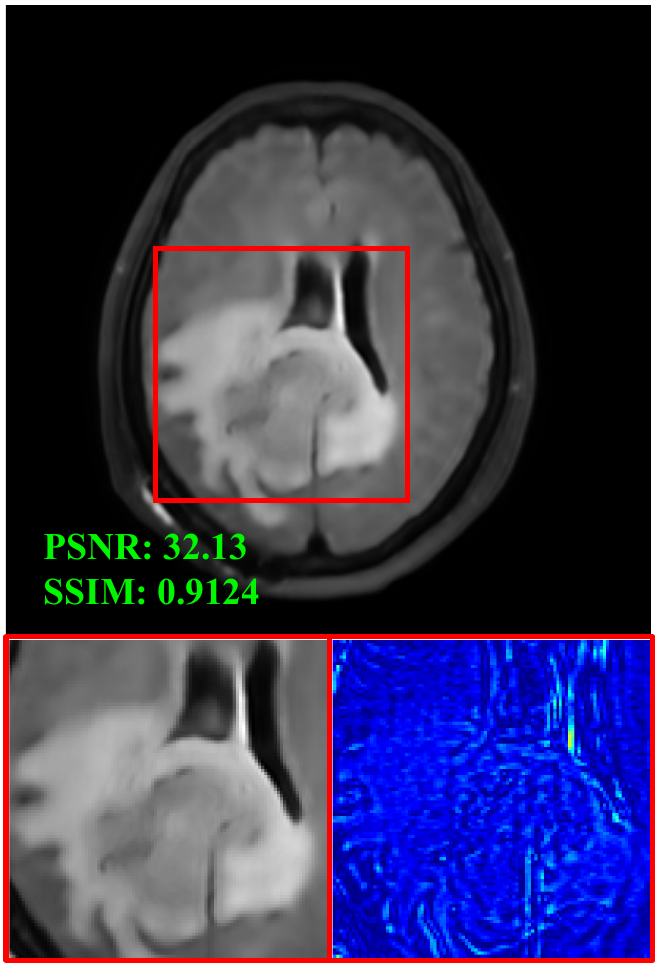}\\
        \includegraphics[scale=0.17]{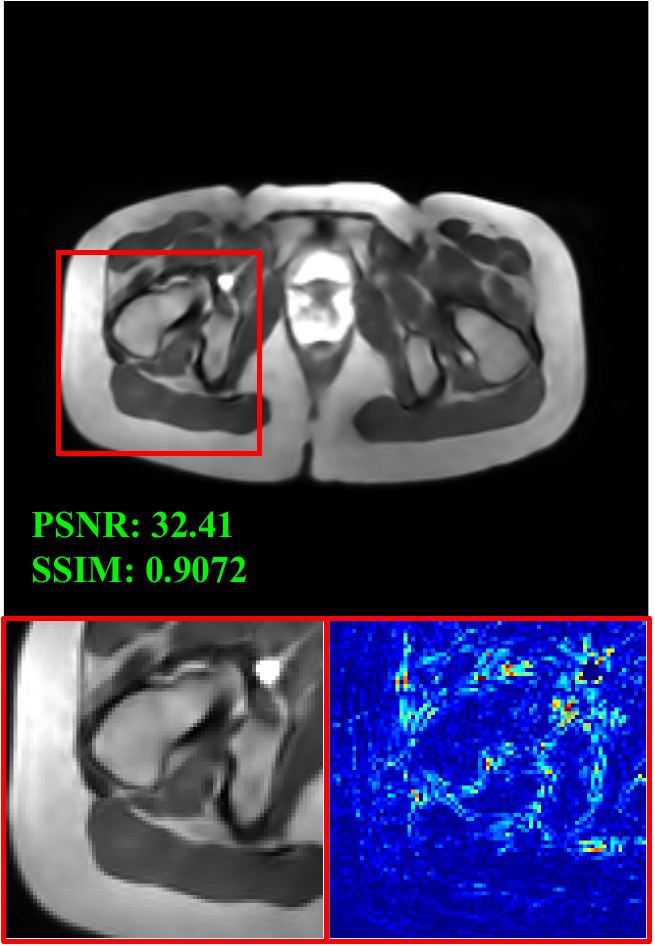}
	\end{minipage}
}
\hfill
    \subfloat[\footnotesize McMRSR \cite{Li2022Trans}]{
	\begin{minipage}[b]{0.1\textwidth}
		\includegraphics[scale=0.17]{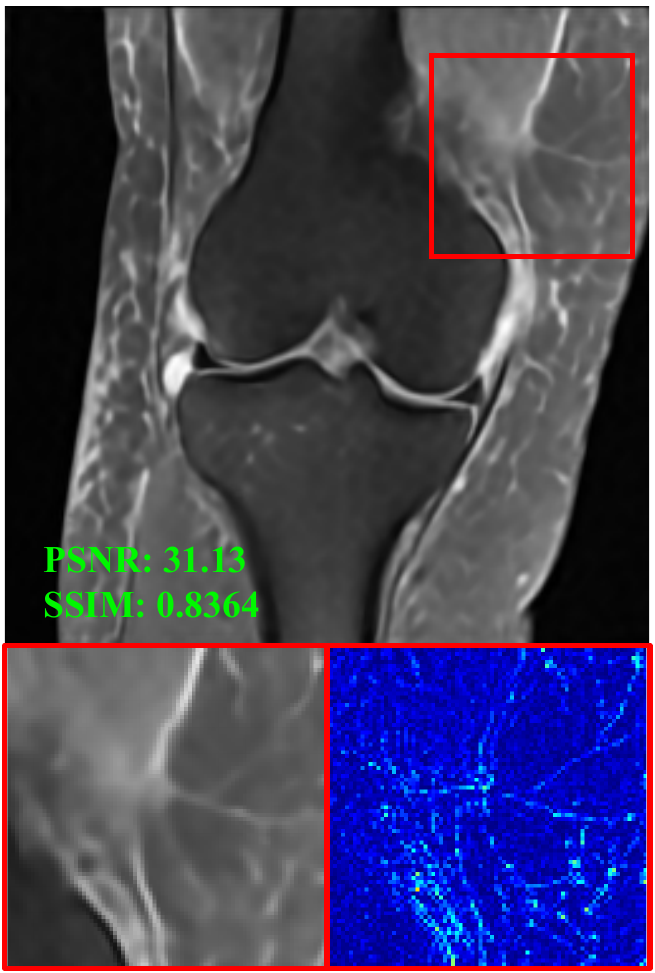} \\
        \includegraphics[scale=0.17]{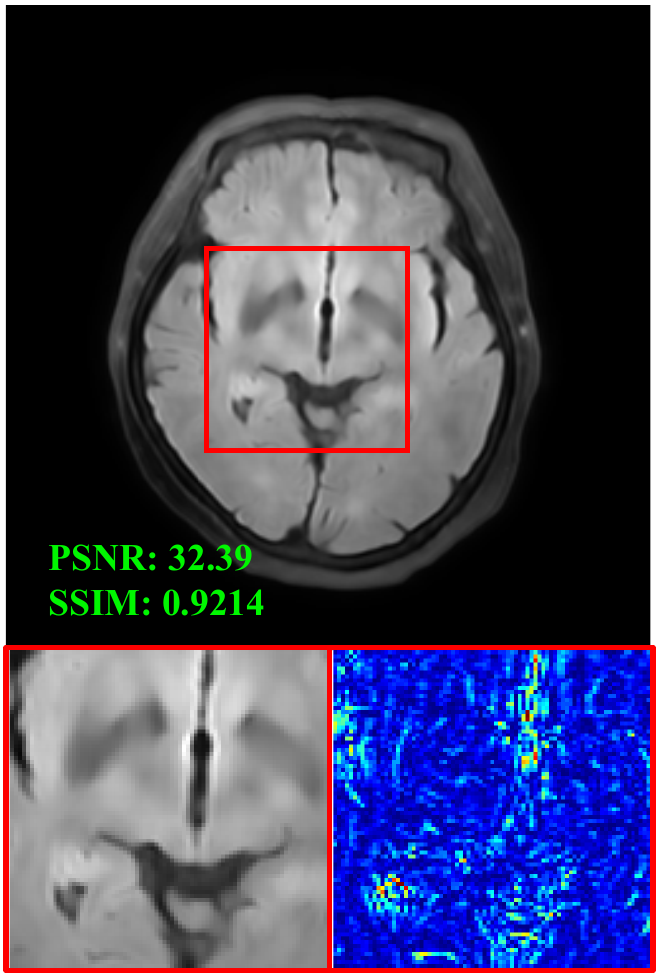}\\
        \includegraphics[scale=0.17]{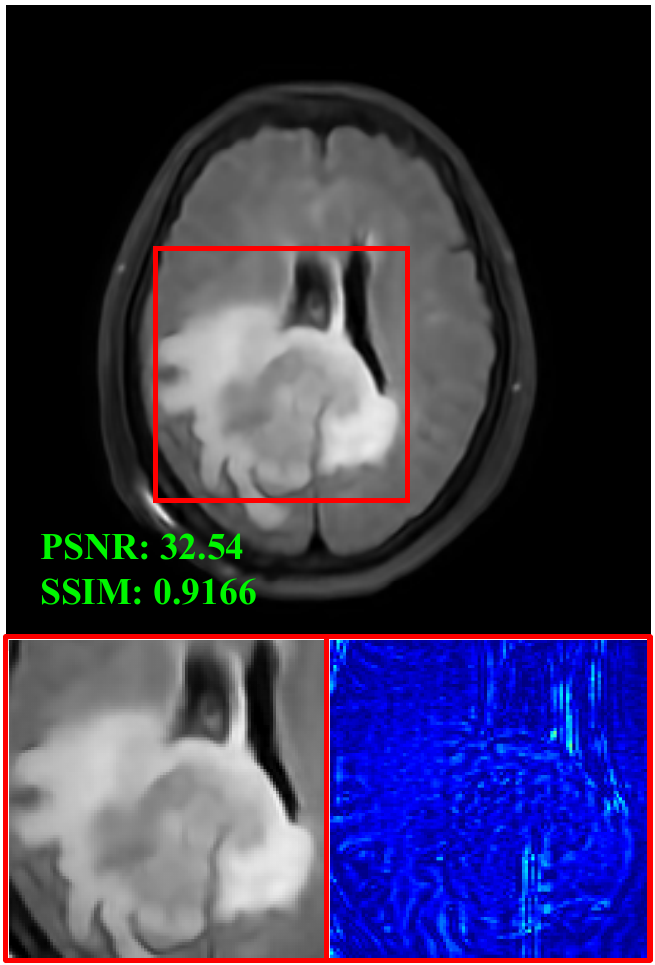}\\
        \includegraphics[scale=0.17]{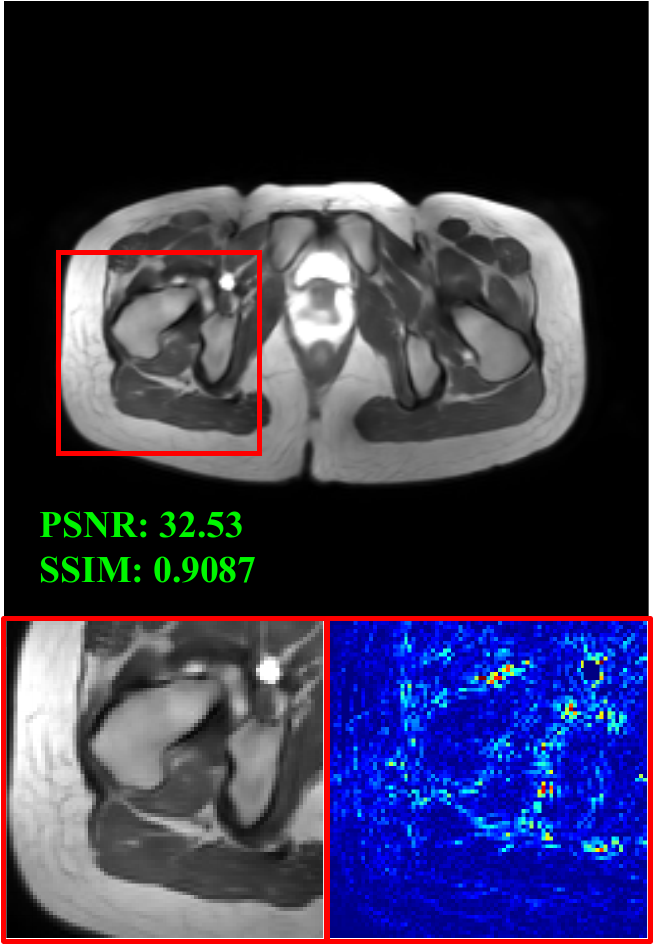}
	\end{minipage}
}
\hfill
    \subfloat[\footnotesize MC-VarNet \cite{lei2023decomposition}]{
	\begin{minipage}[b]{0.1\textwidth}
		\includegraphics[scale=0.17]{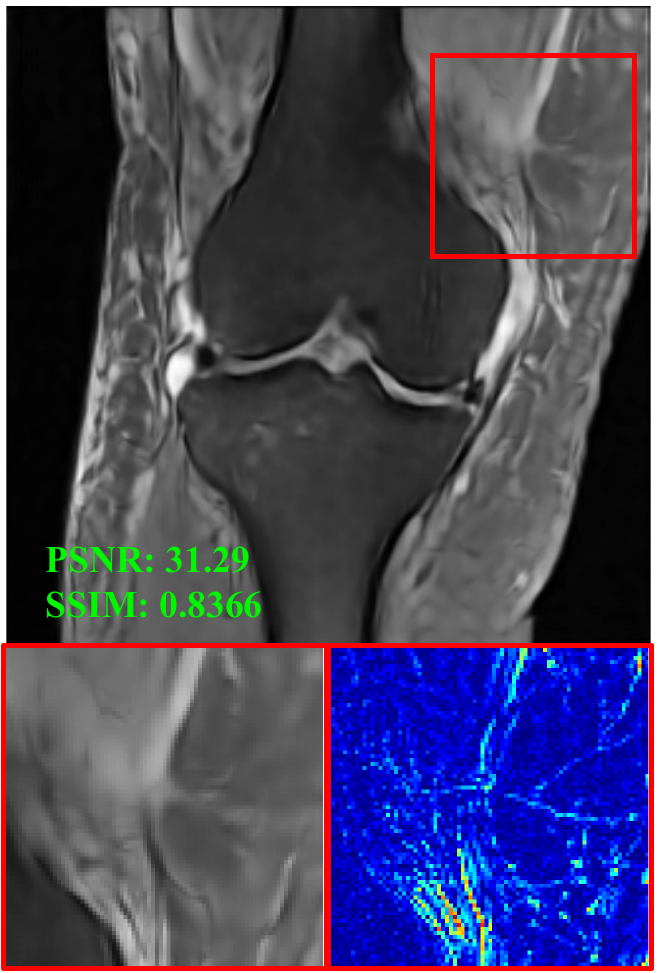}  \\
        \includegraphics[scale=0.17]{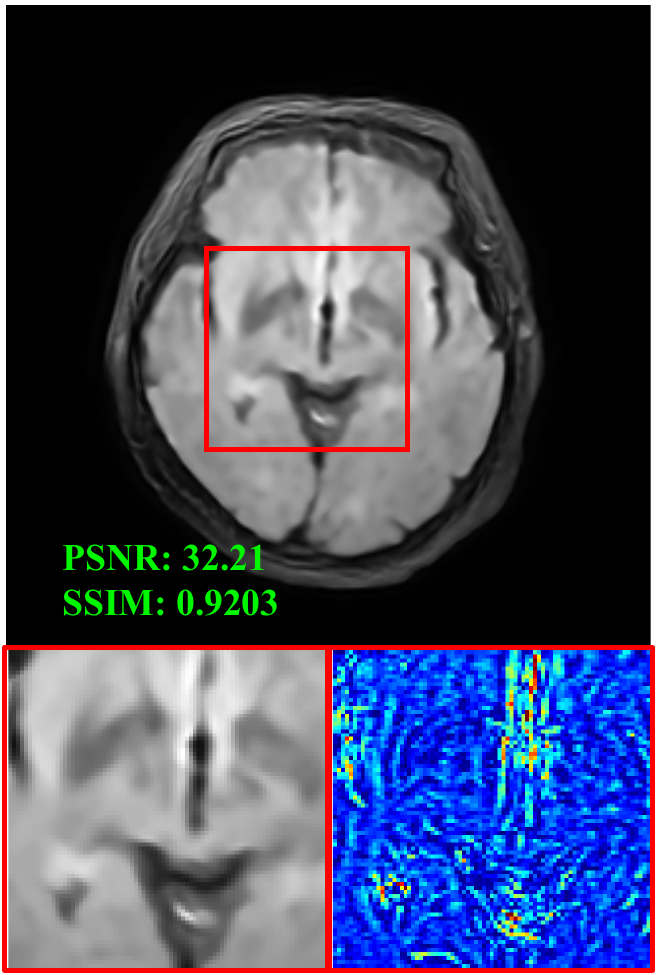} \\
        \includegraphics[scale=0.17]{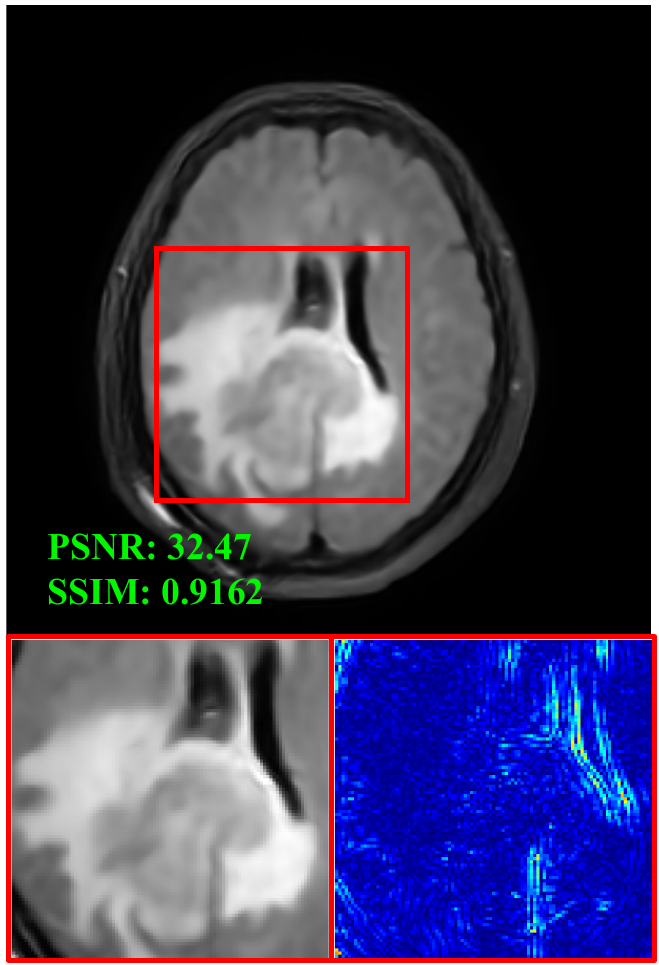}\\
        \includegraphics[scale=0.17]{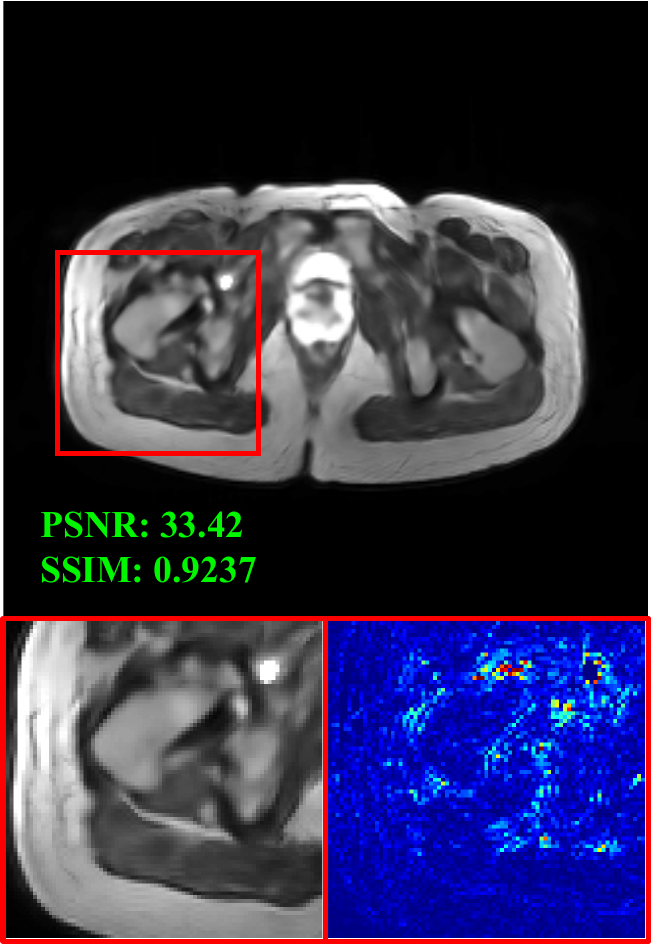} 
	\end{minipage}
}
\hfill
    \subfloat[\footnotesize DisC-Diff \cite{mao2023disc}]{
	\begin{minipage}[b]{0.1\textwidth}
		\includegraphics[scale=0.17]{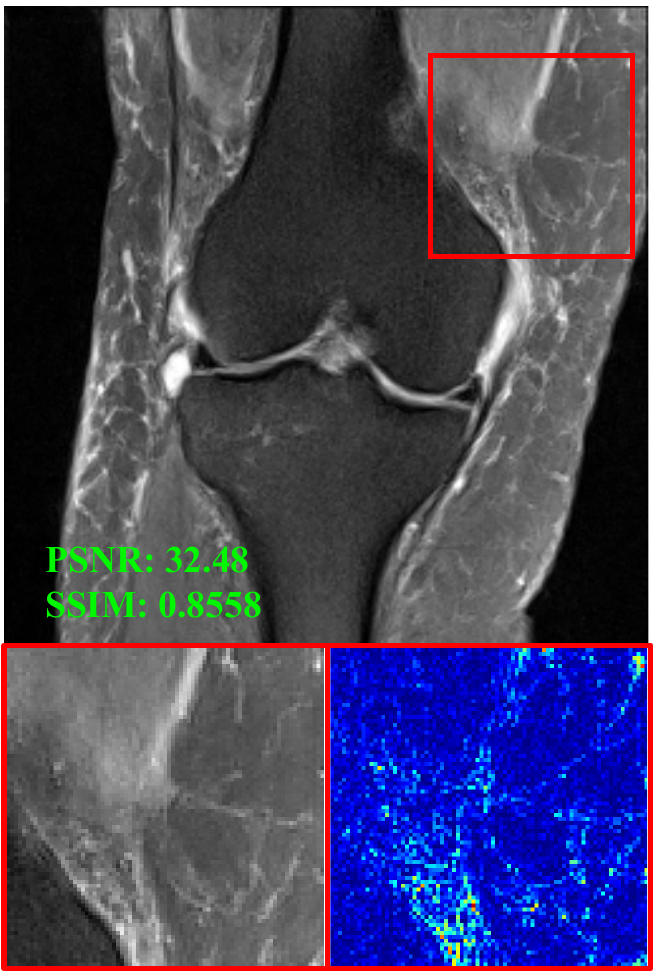} \\
        \includegraphics[scale=0.17]{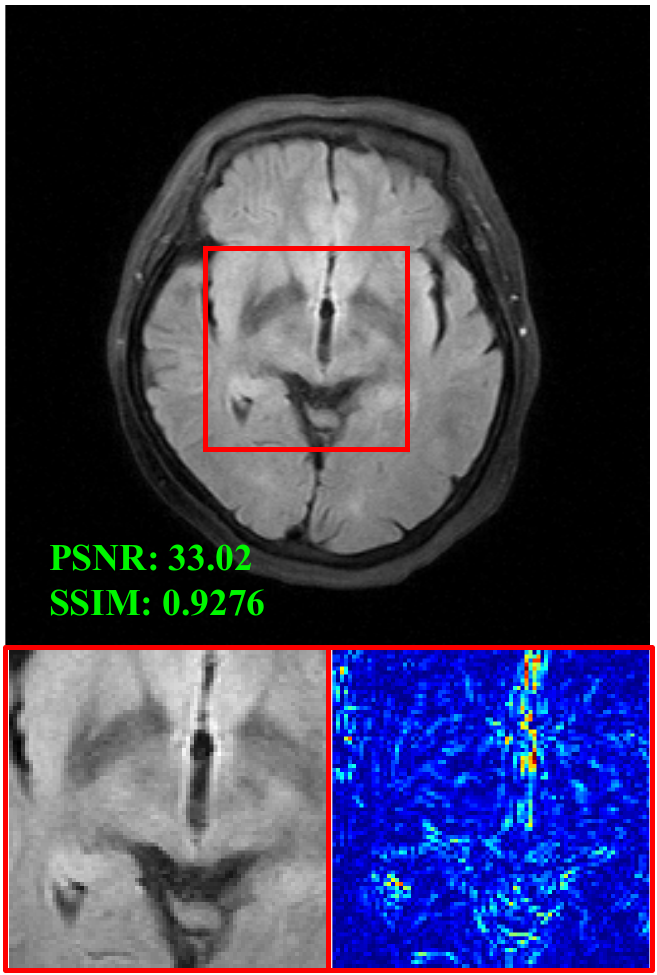}\\
        \includegraphics[scale=0.17]{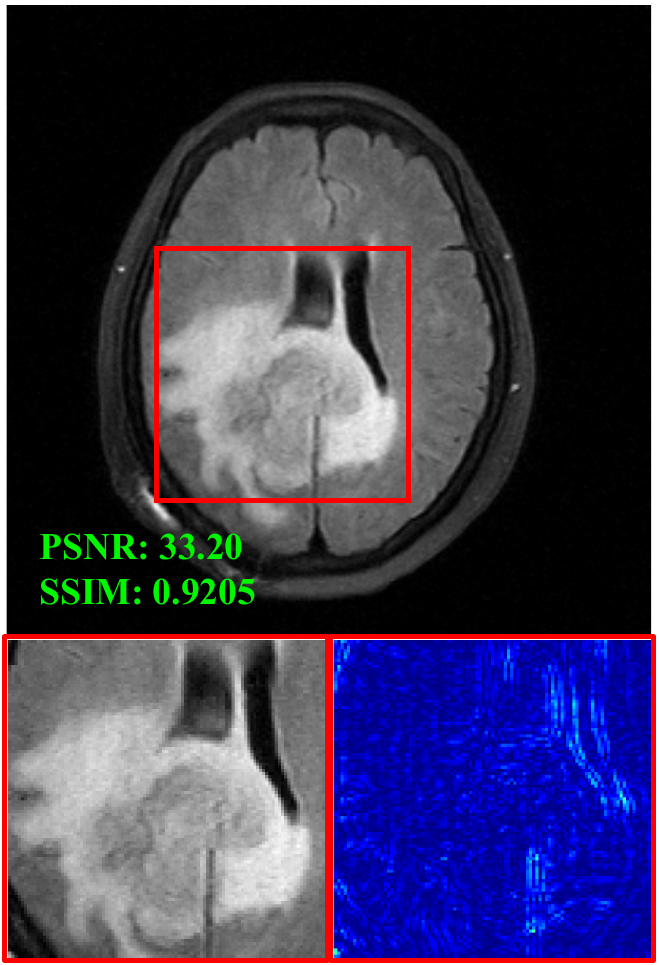}\\
        \includegraphics[scale=0.17]{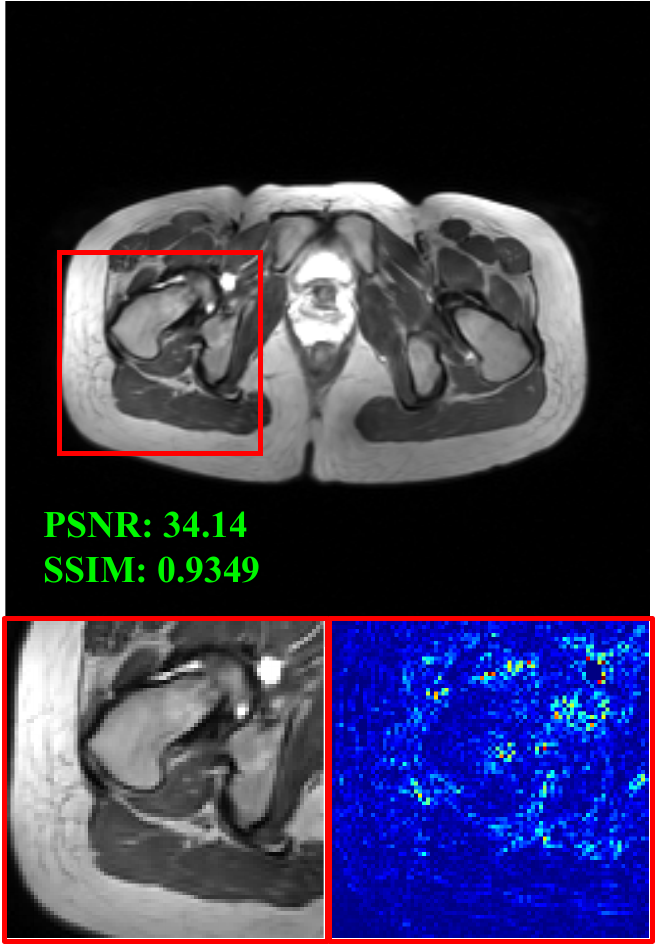}
	\end{minipage}
}
\hfill
    \subfloat[\footnotesize Ours]{
	\begin{minipage}[b]{0.1\textwidth}
		\includegraphics[scale=0.17]{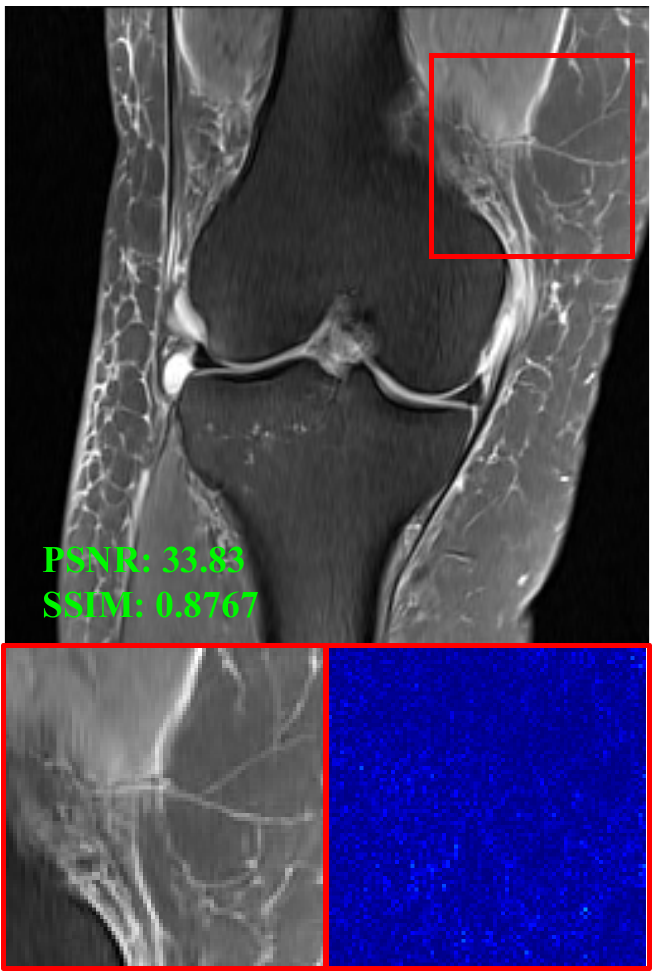} \\
        \includegraphics[scale=0.17]{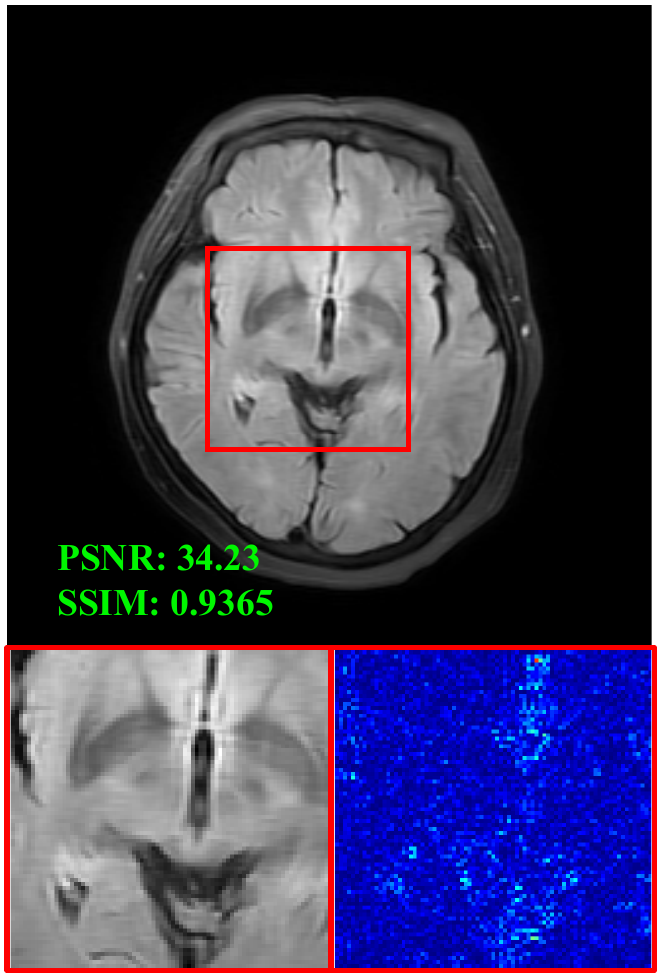}\\
        \includegraphics[scale=0.17]{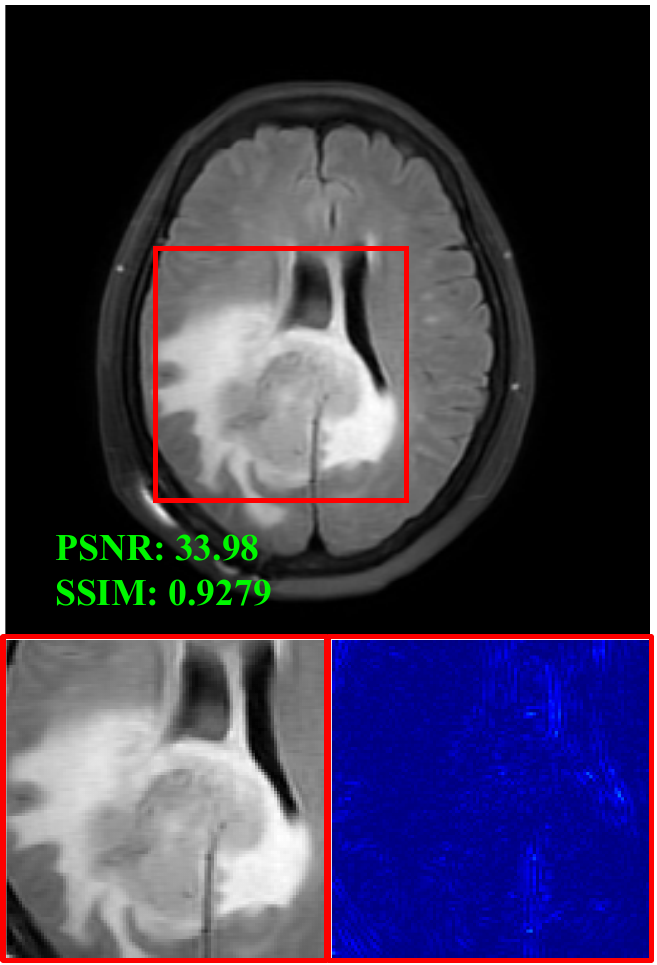}\\
        \includegraphics[scale=0.17]{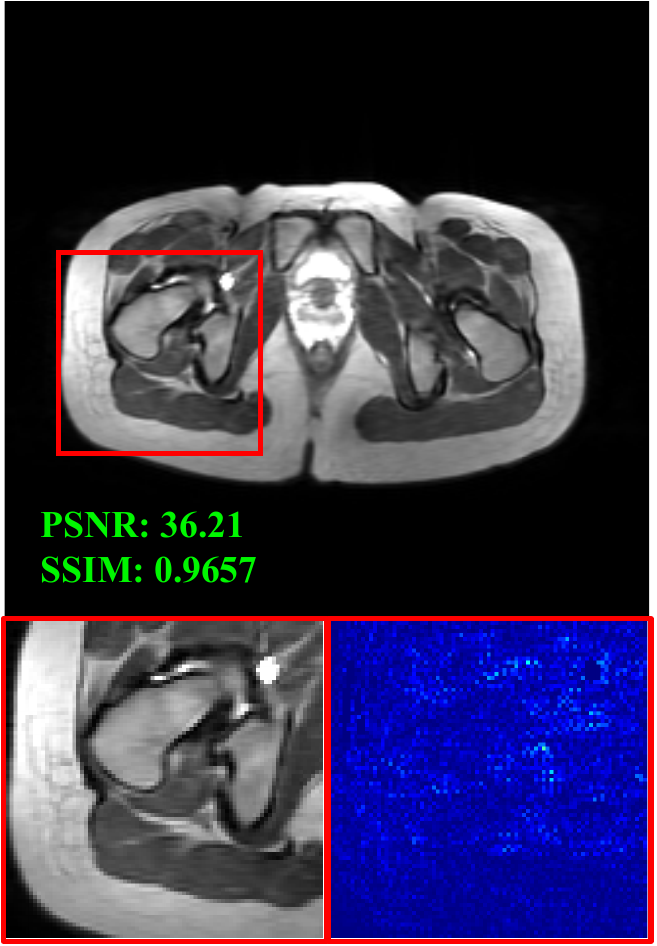}
	\end{minipage}
}
\hfill
    \subfloat[\footnotesize Target HR]{
	\begin{minipage}[b]{0.1\textwidth}
		\includegraphics[scale=0.17]{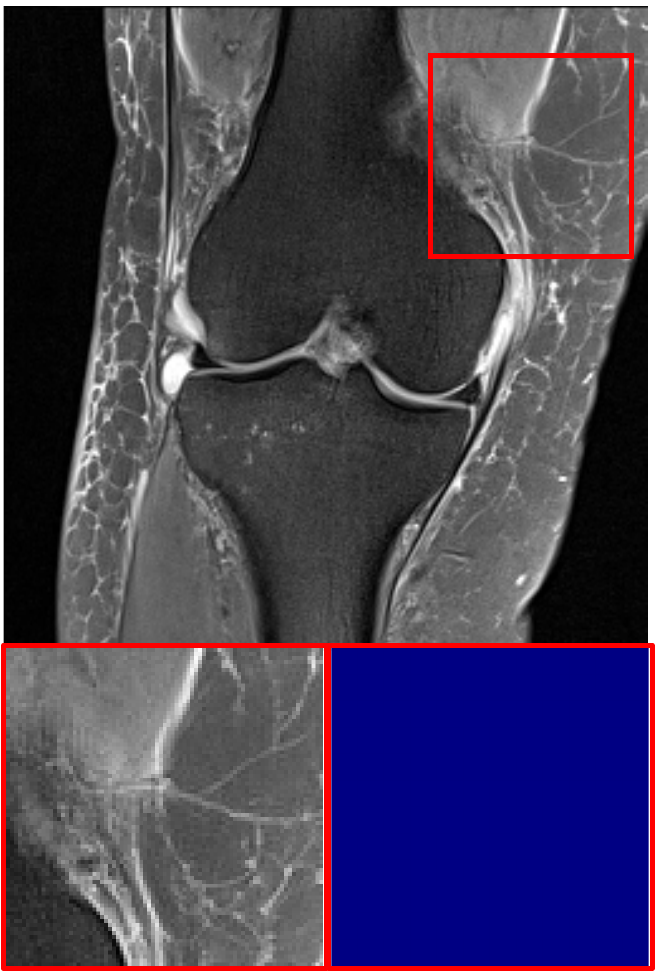} \\
        \includegraphics[scale=0.17]{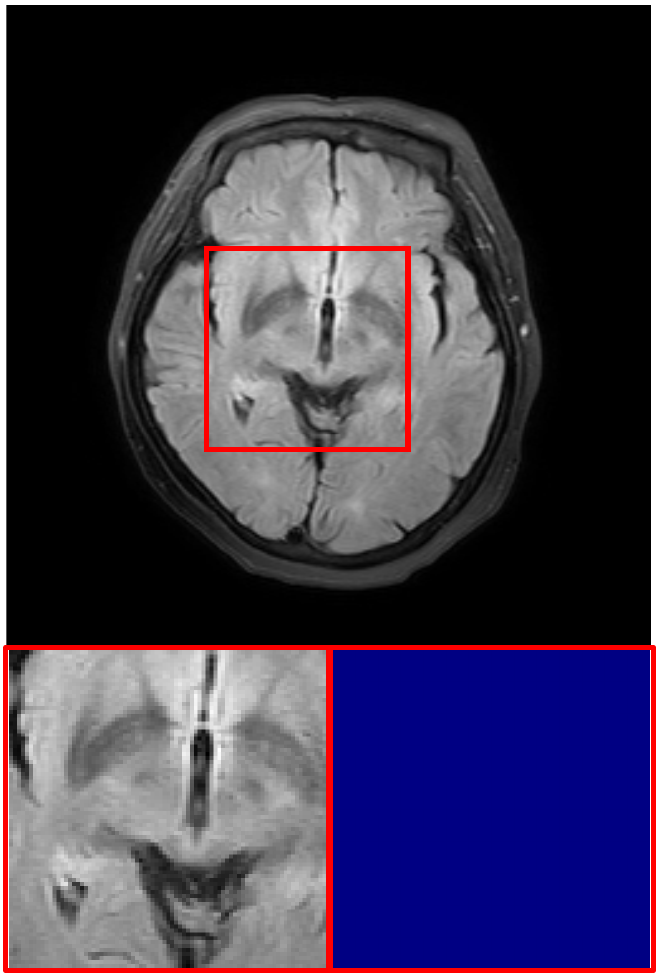}\\
        \includegraphics[scale=0.17]{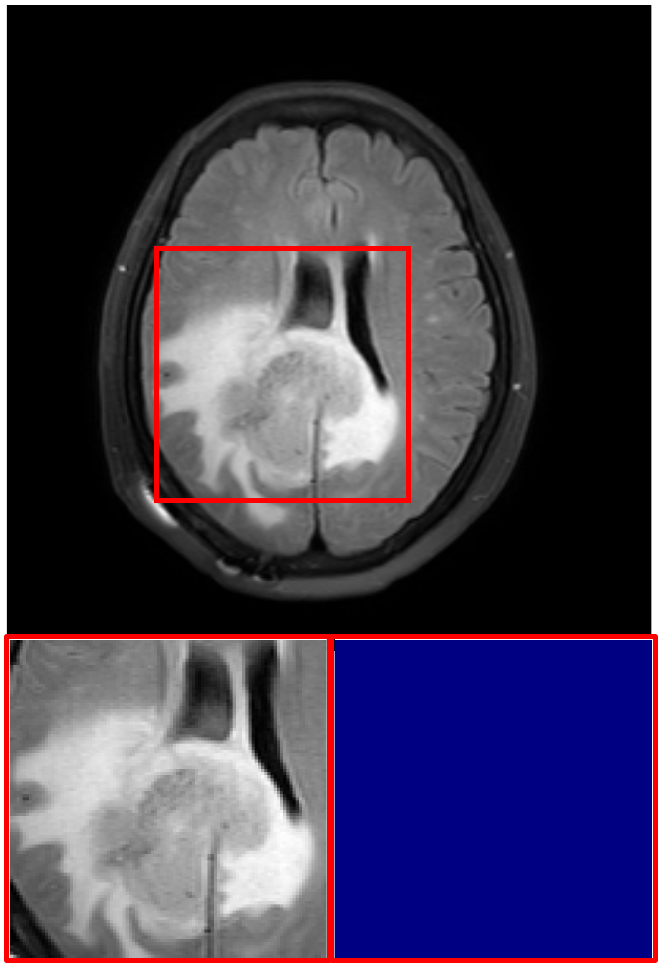}\\
        \includegraphics[scale=0.17]{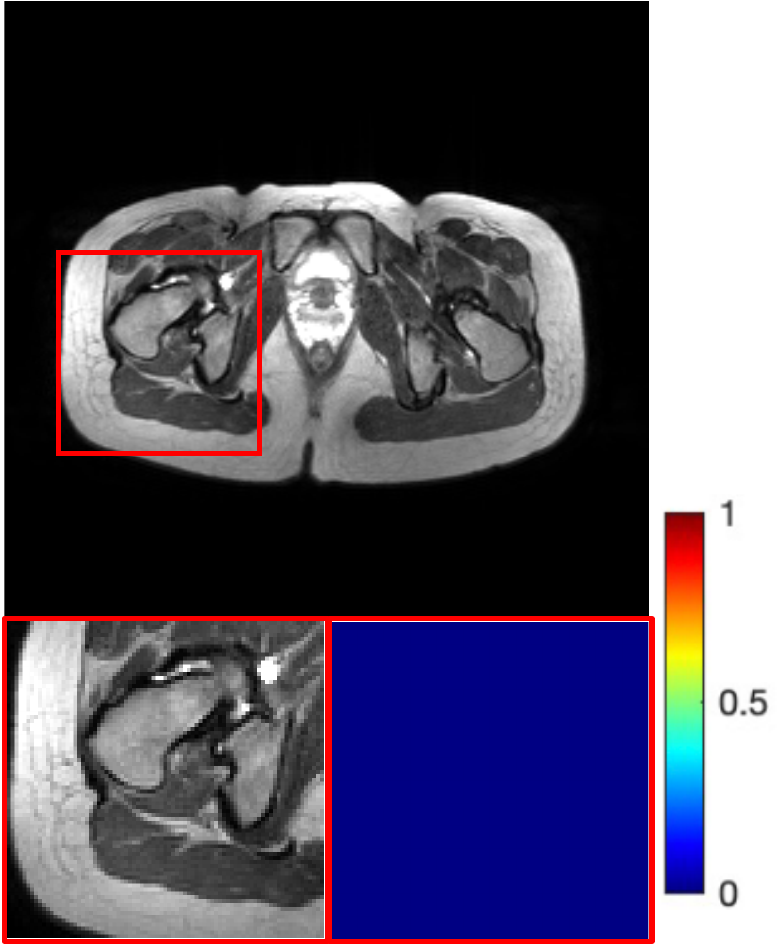}
	\end{minipage}
}
\caption{Qualitative comparison of SOTA MCSR methods on four datasets with an upsampling scale of 4$\times$. The top, second, third, and bottom rows are the SR results under the FastMRI, clinical brain, clinical tumor and clinical pelvic datasets. Please zoom-in for details.}
\label{fig:4fold}
\end{figure*}

\begin{table*}[!h]
\centering
\small
\begin{tabular}{l|c|c|c|cc|cc|cc|cc}  
\hline
\hline
\multirow{2}*{Methods} &
\multirow{2}*{Param} & 
\multirow{2}*{FLOPs} & 
\multirow{2}*{Speed} & 
\multicolumn{2}{c|}{ FastMRI \cite{zbontar2018fastmri}} &
\multicolumn{2}{c|}{ Clinical Brain} & 
\multicolumn{2}{c|}{ Clinical Tumor} & 
\multicolumn{2}{c}{ Clinical Pelvic} 
\\
\cline{5-12}
& & & &   \makecell[c]{PSNR}
&  \makecell[c]{SSIM}  
&  \makecell[c]{PSNR} 
&  \makecell[c]{SSIM} 
&  \makecell[c]{PSNR} 
&  \makecell[c]{SSIM}  
&  \makecell[c]{PSNR} 
&  \makecell[c]{SSIM} 
\\
\hline
\hline
\makecell[l]{MCSR \cite{Lyu2019Multi}} 
& \makecell[c]{3.5\emph{M}}
& \makecell[c]{90.399\emph{G}}
& \makecell[c]{103\emph{ms}}
&  \makecell[c]{28.41} & \makecell[c]{0.8039}   
& \makecell[c]{ 28.55} & \makecell[c]{0.8930} 
& \makecell[c]{28.24} & \makecell[c]{0.8909}  
& \makecell[c]{29.59} & \makecell[c]{0.8713}
\\
\makecell[l]{MINet \cite{Feng2021Multi}} 
& \makecell[c]{11.9\emph{M}}
& \makecell[c]{866.933\emph{G}}
& \makecell[c]{115\emph{ms}}
& \makecell[c]{30.19} & \makecell[c]{0.8101}
& \makecell[c]{30.93} & \makecell[c]{0.9049}
& \makecell[c]{30.57} & \makecell[c]{0.8962} 
& \makecell[c]{30.75} & \makecell[c]{0.8861}
\\
\makecell[l]{MASA \cite{lu2021masa}} 
& \makecell[c]{4.0\emph{M}}
& \makecell[c]{180.134\emph{G}}
& \makecell[c]{110\emph{ms}}
& \makecell[c]{30.51} & \makecell[c]{0.8192}   
& \makecell[c]{31.17} & \makecell[c]{0.9076} 
& \makecell[c]{30.81} & \makecell[c]{0.8981} 
& \makecell[c]{30.94} & \makecell[c]{0.8879} 
\\
\makecell[l]{WavTrans \cite{Li2022Wav}} 
& \makecell[c]{2.1\emph{M}}
& \makecell[c]{162.889\emph{G}}
& \makecell[c]{173\emph{ms}}
& \makecell[c]{30.95} & \makecell[c]{0.8343}
& \makecell[c]{32.42} & \makecell[c]{0.9215} 
& \makecell[c]{31.99} & \makecell[c]{0.9107}
& \makecell[c]{32.15} & \makecell[c]{0.9054} 
\\
\makecell[l]{McMRSR \cite{Li2022Trans}} 
& \makecell[c]{3.5\emph{M}}
& \makecell[c]{269.860\emph{G}}
& \makecell[c]{168\emph{ms}}
& \makecell[c]{30.96} & \makecell[c]{0.8345}  
& \makecell[c]{32.48} & \makecell[c]{0.9217}  
& \makecell[c]{32.07} & \makecell[c]{0.9121}
& \makecell[c]{32.19} & \makecell[c]{0.9063} 
\\
\makecell[l]{MC-VarNet \cite{lei2023decomposition}} 
& \makecell[c]{5.7\emph{M}}
& \makecell[c]{139.862\emph{G}}
& \makecell[c]{97\emph{ms}}
& \makecell[c]{31.65} & \makecell[c]{0.8412}
& \makecell[c]{32.74} & \makecell[c]{0.9231}
& \makecell[c]{32.16} & \makecell[c]{0.9153}
& \makecell[c]{33.51} & \makecell[c]{0.9265}
\\
\rowcolor{green!10} DisC-Diff \cite{mao2023disc}
& 86.1\emph{M}
& 461.002\emph{G}
& 1455\emph{ms}
&  \textcolor{blue}{{32.24}}
&  \textcolor{blue}{{0.8530}}
&  \textcolor{blue}{{32.85}}
&  \textcolor{blue}{{0.9285}}
&  \textcolor{blue}{{32.64}} 
&  \textcolor{blue}{{0.9187}}
&  \textcolor{blue}{{34.26}}
&  \textcolor{blue}{{0.9388}}
\\
\rowcolor{mypink} DiffMSR (Ours)
& 6.8\emph{M}
& 58.617\emph{G}
& 39\emph{ms}
&  \textcolor{red}{33.78}
&  \textcolor{red}{0.8765}
&  \textcolor{red}{34.12}
&  \textcolor{red}{0.9362}
&  \textcolor{red}{33.85} 
&  \textcolor{red}{0.9274}
&  \textcolor{red}{35.47} 
&  \textcolor{red}{0.9607}
\\
\hline
\hline
\end{tabular}
\caption{Quantitative comparison with state-of-the-art methods on four datasets, with performance measured in terms of PSNR (dB) $\uparrow$ and SSIM $\uparrow$. The most outstanding results are indicated in \textcolor{red}{red} (the best) and \textcolor{blue}{blue} (the second-best). Param refers to the model parameters. Speed means the inference speed.}
\label{tab:metrics1}
\end{table*}

\subsection{Qualitative results}
Figure \ref{fig:4fold} provides the qualitative comparison of the various methods on the four datasets at a scale of 4$\times$. 
The top, second, third, and bottom rows are the SR results under the FastMRI, clinical brain, clinical tumor and clinical pelvic datasets, respectively. 
The red boxes indicate the zoom-in region of complicated anatomical structures along with their corresponding error maps. Note that the brighter textures in the error maps, the lower the quality of the reconstructed images.
As can be seen, compared to methods based on Transformers and CNNs, diffusion-based methods like DisC-Diff and DiffMSR (Ours) are capable of reconstructing high-realistic images with promising reconstruction metric scores (PSNR and SSIM).
Nevertheless, while DisC-Diff can reconstruct high-precision MR images, it does not preserve the structure present in the original HR images, introducing some additional information that can affect medical diagnosis. In contrast, our method combines DM and PLWformer, which can preserve the original image's structure while restoring high-frequency information.

\subsection{Quantitative results}
In Table \ref{tab:metrics1}, we provide a comprehensive quantitative analysis, comparing our DiffMSR with other state-of-the-art MCSR methods on four datasets with a 4$\times$ upscaling factor. As we can see, our DiffMSR performs best among all comparison methods in terms of PSNR and SSIM metrics in all MRI datasets.
Specifically, we notice that the performance of CNN-based methods is relatively poor as CNNs struggle to capture long-range dependencies.
Although Transformer-based methods address some of the issues with CNNs, the ability of Transformers to reconstruct high-frequency details is limited, restricting further improvements in metrics.
The DisC-Diff method based on DM achieves high metric values, as DM can generate some high-frequency details, but it also introduces unnecessary information.
Our proposed DiffMSR combines the strengths of Transformer and DM, preserving the original image structure while maximizing the reconstruction of complicated anatomical structures, obtaining the best performance.

In addition, we provide the model parameters, FLOPs, and inference speed of each model in Table \ref{tab:metrics1}. As can be seen, our proposed method has mid-range model parameters, the smallest FLOPs, and the fastest inference speed. 
This is due to the two strategies we introduce to reduce computational overhead and speed up inference. (1) In the PLWformer, we employ permutation operations to decrease the computational cost of self-attention.
(2) We apply the diffusion model within a highly compact latent space, requiring only a simple denoising network and a few iterations to generate prior knowledge.
\begin{table*}[h!]
\small
\centering
\begin{tabular}{l|c|c|c|c|c|c|c|c}   
\hline
\hline
\multirow{2}*{Variants} & \multicolumn{6}{c|}{\cellcolor[gray]{0.9} Components} &\multicolumn{2}{c}{\cellcolor[gray]{0.9} Metrics} \\  
\cline{2-9}
& \cellcolor[gray]{0.9} Reference & \cellcolor[gray]{0.9} Diffusion & \cellcolor[gray]{0.9} Joint-training & \cellcolor[gray]{0.9} DC Loss & \cellcolor[gray]{0.9} Condition & \cellcolor[gray]{0.9} PLWformer & \cellcolor[gray]{0.9} PSNR $\uparrow$ & \cellcolor[gray]{0.9} SSIM $\uparrow$   \\
\hline
\hline
\emph{w/o} reference &  \textcolor{cvprblue}{\usym{2717}} & \usym{2713} & \usym{2713} & \usym{2713} & \usym{2713}  & \usym{2713}  & 32.56 & 0.8562 
\\
\emph{w/o} prior & \usym{2713} &   \textcolor{cvprblue}{\usym{2717}} & \textcolor{cvprblue}{\usym{2717}} & \usym{2713} & \textcolor{cvprblue}{\usym{2717}}  & \usym{2713}  & 31.51 &  0.8428
\\
\emph{w/o} joint & \usym{2713} & \usym{2713}  &   \textcolor{cvprblue}{\usym{2717}} & \usym{2713} & \usym{2713}  & \usym{2713}  & 32.35  &  0.8541
\\
\emph{w/o} DC & \usym{2713} & \usym{2713}  & \usym{2713}  &   \textcolor{cvprblue}{\usym{2717}} & \usym{2713}  & \usym{2713}  & 32.92  &  0.8604
\\
\emph{w/o} CE & \usym{2713} & \usym{2713}  & \usym{2713}  &  \usym{2713} &   \textcolor{cvprblue}{\usym{2717}}  & \usym{2713}  & 33.21  &  0.8638
\\
\emph{w/o} PLWformer & \usym{2713} & \usym{2713}  & \usym{2713}  &  \usym{2713} &  \usym{2713}   &  \textcolor{cvprblue}{\usym{2717}}  & 33.30 &  0.8650
\\
Full model & \usym{2713} &  \usym{2713}  &  \usym{2713}  &  \usym{2713} &  \usym{2713}   &  \usym{2713}  & \textbf{33.78}  &  \textbf{0.8765}
  \\
\hline
\hline
\end{tabular}
\caption{Ablation study on various variants under FastMRI with an upsampling scale of 4$\times$. The best quantitative result is marked in \textbf{bold}. }
\label{tab_ab}
\end{table*}


\begin{table}[t!]
\small
\centering
\begin{tabular}{l|c|c|c}
\hline
\hline
Ratio $k$ & \cellcolor[gray]{0.9} $k=1$ & \cellcolor[gray]{0.9} $k=2$ & \cellcolor[gray]{0.9} $k=4$  \\  
\hline
\hline
 PSNR & \textcolor{red}{33.81}  & \textcolor{blue}{33.78}  & 33.70   
 \\
\hline
 FLOPs & 68.691\emph{G}  & \textcolor{blue}{58.617\emph{G}}  & \textcolor{red}{56.224\emph{G}} 
\\
\hline
\hline
\end{tabular}
\caption{Ablation study on various ratios of $k$ in FastMRI with an upsampling scale of 4$\times$. The best and second-best results are indicated in \textcolor{red}{red} and \textcolor{blue}{blue}, respectively.}
\label{tab:k}
\end{table}

\subsection{Ablation Study}
In this section, we explore the effectiveness of each key component of our proposed DiffMSR. All variants are retrained the same way as before and tested on the FastMRI dataset with an upsampling scale of 4$\times$.

\textbf{Effect of Reference Image.}
To investigate the contribution of the reference image, we design a variant \emph{w/o} reference, which only utilizes the target LR image for reconstruction without the reference image, as shown in Table \ref{tab_ab}.
As can be seen, without employing the reference image, the reconstruction performance is significantly decreased, which demonstrates that the reference image can provide some valuable supplementary information.

\textbf{Effect of Prior.}
To validate the role of prior knowledge, we design a variant that only utilizes the first stage and excludes the prior extraction and guidance module, denoted as \emph{w/o} prior, as shown in Table \ref{tab_ab}. 
As can be seen, without the high-frequency detail information provided by prior knowledge, the reconstruction performance has significantly declined, indicating the limited ability of the Transformer to reconstruct high-frequency details.

\textbf{Effect of Joint-Training.}
To investigate the effect of the joint training strategy in the second stage, we only optimize the diffusion model and CE in stage two, denoted as \emph{w/o} joint. Specifically, we employ $\mathcal{L}_{diff}$ in stage two to solely train the diffusion model. After training, the diffusion model is directly combined with the PLWformer trained in the first stage for evaluation. As shown in Table \ref{tab_ab}, the results indicate that the performance of this separated training is lower than that of joint training, demonstrating the effectiveness of employing joint training in stage two.

\textbf{Effect of Data Consistency Loss.}
To evaluate the contribution of data consistency (DC) loss, we conduct an ablation study by removing $\mathcal{L}_{dc}$ in the optimization function, named as \emph{w/o} DC, as shown in Table \ref{tab_ab}.
As can be seen, the reconstruction performance of \emph{w/o} DC has significantly decreased, with a reduction of 0.86 in PSNR, indicating that the DC loss can effectively supplement frequency domain information for MR images, thereby improving the reconstruction results in the image domain.

\begin{figure}[t]
  \centering
   \includegraphics[width=\linewidth]{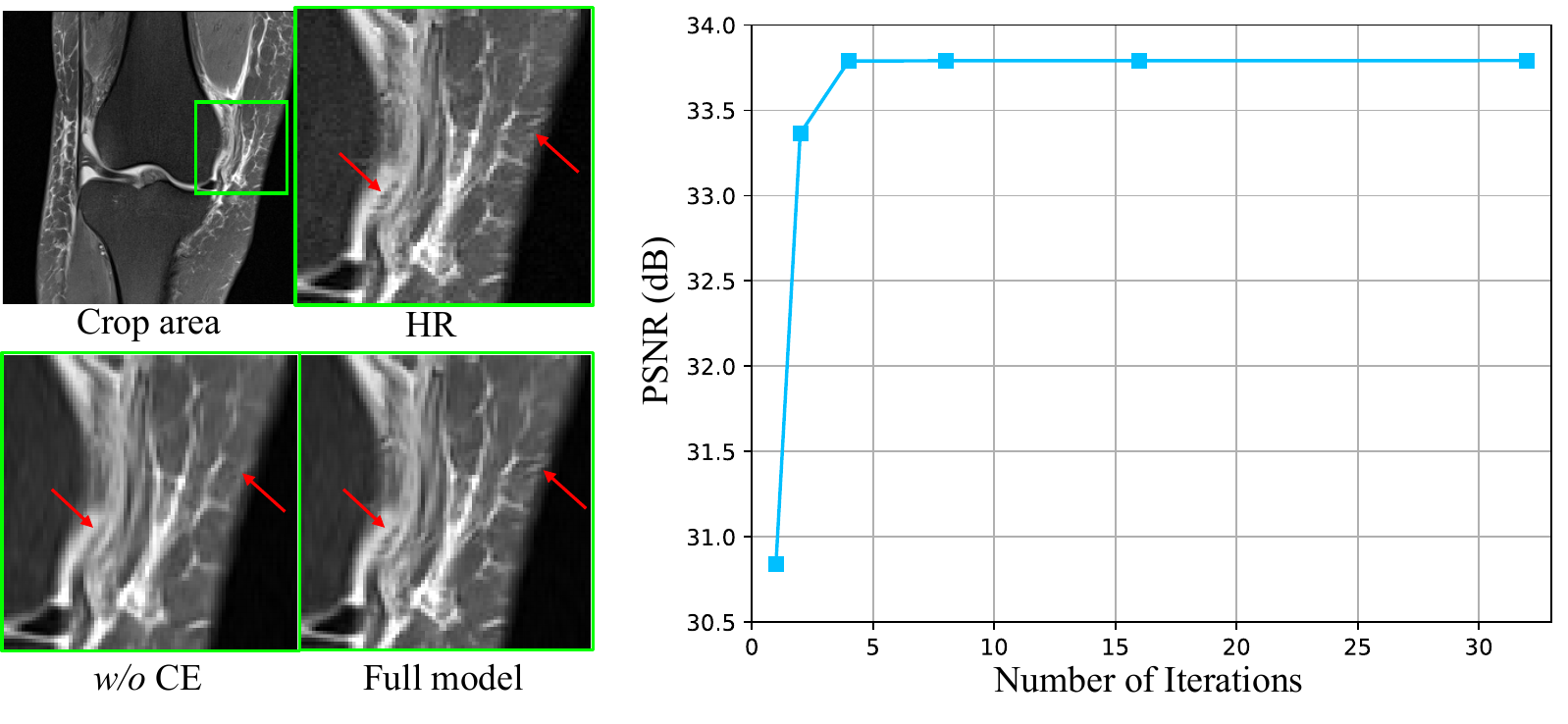}
   \caption{Left: Visualization comparison between \emph{w/o} CE and the full model. Right: Ablation study of the number of iteration steps in the diffusion model.}
   \label{fig_ab}
\end{figure}

\textbf{Effect of Condition.}
To evaluate the effect of the condition extraction module, we design a variant by removing CE (named as \emph{w/o} CE), which means that condition $C$ is not employed in the denoising process, as shown in Table \ref{tab_ab}.
As can be seen, without condition $C$, the reconstruction performance decreases, which demonstrates that condition $C$ extracted by the CE can provide supplementary information for the denoising network.
Besides, Figure \ref{fig_ab} provides a qualitative comparison \emph{w/o} CE. As can be seen, without condition $C$, the reconstructed image will lose some complicated anatomical structures.

\textbf{Effect of Iterations.}
To explore the impact of the number of iterations on the diffusion model, we set up six variants utilizing different numbers of iteration steps $T$=[1, 2, 4, 8, 16, 32]. We plot the PSNR for different iterations in Figure \ref{fig_ab}. We notice that when $T$=1, the DM fails to generate reasonable prior knowledge, thus limiting the reconstruction performance. When $T$=4, the growth rate of PSNR becomes very flat, indicating that generating valuable prior knowledge only requires a small number of iterations as the simple distribution of the highly compact latent space.


\textbf{Effect of Ratio \emph{k}.}
We conduct an ablation study to investigate the impact of $k$. Specifically, we set $k$ to 1, 2, and 4, respectively and the results are shown in Table \ref{tab:k}. As can be seen, compared to $k$=1, the model with $k$=2 shows a significant reduction in FLOPs by about 10\emph{G}, while the PSNR only slightly decreases by 0.03dB. 
However, when $k$=4, the model's performance is deceased as the spatial dimensions of $K$ and $V$ are greatly reduced, resulting in partial information loss.
Therefore, to balance the FLOPs and performance, we set $k$ to 2.

\section{Conclusion}
We propose an efficient diffusion model for multi-contrast MRI SR, which combines DM and Transformer, requiring only four iterations to reconstruct high-quality images. 
Besides, we introduce the PLWformer, which can expand the window size of attention without increasing the computational burden and can utilize the prior knowledge generated by DM to reconstruct MR images with high-frequency information.
Extensive experiments demonstrate that our DiffMSR outperforms existing SOTA methods.


\noindent \textbf{Acknowledgements.} This work was supported in part by Zhejiang Province Program (2022C01222, 2023C03199, 2023C03201), the National Program of China (62172365, 2021YFF0900604, 19ZDA197), Ningbo Science and Technology Plan Project (022Z167, 2023Z137), and MOE Frontier Science Center for Brain Science $\&$ Brain-Machine Integration (Zhejiang University).

{
    \small
    \bibliographystyle{ieeenat_fullname}
    \bibliography{main}
}

\clearpage
\setcounter{page}{1}
\maketitlesupplementary

\section{Prior Extraction}
\begin{figure}[h]
  \centering
   \includegraphics[width=\linewidth]{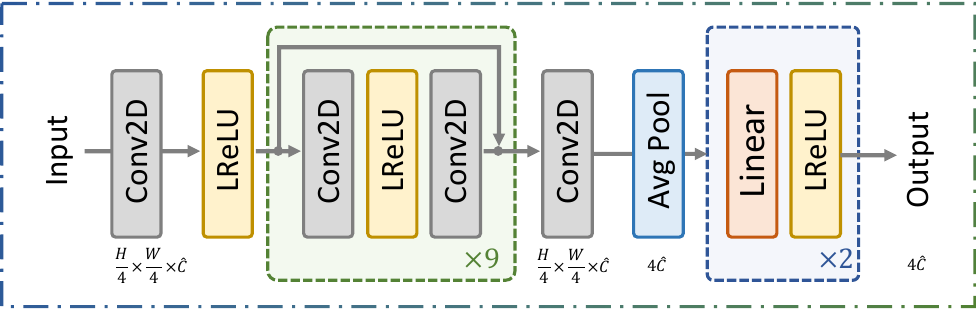}
   \caption{The architecture of the prior extraction module.}
   \label{fig_pe}
\end{figure}

The architecture of the PE is shown in Figure \ref{fig_pe}. As can be seen, PE mainly consists of 9 residual blocks and 2 linear layers. 
Specifically, we first concatenate the target LR image $I_{LR} \in \mathbb{R}^{H \times W \times 2}$ and the target HR image $I_{HR}^P \in \mathbb{R}^{H \times W \times 2}$ after the PixelUnshuffle operation along the channel dimension to obtain $X \in \mathbb{R}^{H \times W \times 4}$. 
Then, input X into the PE to generate the prior knowledge $Z \in \mathbb{R}^{4\hat{C}}$. Note that when employed as the condition extraction (CE) module, only the target LR image $I_{LR}$ is utilized as input.

\section{Frequency-Domain Data Consistency Loss}
Unlike natural images, MR images undergo data acquisition in the frequency domain. Therefore, in MR image SR reconstruction, frequency domain loss is crucial for maintaining data consistency by using sampled values to replace specific \emph{k}-space positions. Specifically, Fourier transforms are performed on both $I_{SR}$ and $I_{HR}$ to obtain \emph{k}-space data $K_{SR}$ and $K_{HR}$, respectively. Then, a sampling mask $M$ is employed to evaluate \emph{k}-space sampling. If the coefficients in $K_{SR}$ have been sampled, they are replaced with the corresponding coefficients in $K_{HR}$; otherwise, they stay unchanged, as follows:
\begin{eqnarray}
K_{D C}[a, b] = \begin{cases} K_{S R}[a, b] & \mbox{if }(a, b) \notin M\\
\frac{K_{S R}[a, b]+n K_{H R}[a, b]}{1+n} & \mbox{if }(a, b) \in M \end{cases},
\end{eqnarray}
where $n \geq 0$ is noise ($n$ is set to infinity), [a, b] is a matrix indexing operation, and $K_{D C}$ is the \emph{k}-space data after fidelity. The sampling mask $M$ employs a commonly used \emph{k}-space downsampling mask \cite{Li2022Trans,lyu2023multicontrast}.
Then, the mean squared error is used to constrain $K_{D C}$ and $K_{H R}$:
\begin{equation}
\mathcal{L}_{dc} = \left\|K_{D C}-K_{H R}\right\|_2.
\end{equation}

\begin{figure*} [h]
	\centering
	\captionsetup[subfloat]{labelformat=empty}
	\subfloat[\footnotesize MCSR \cite{Lyu2019Multi}]{
	\begin{minipage}[b]{0.1\textwidth}
    	\includegraphics[scale=0.17]{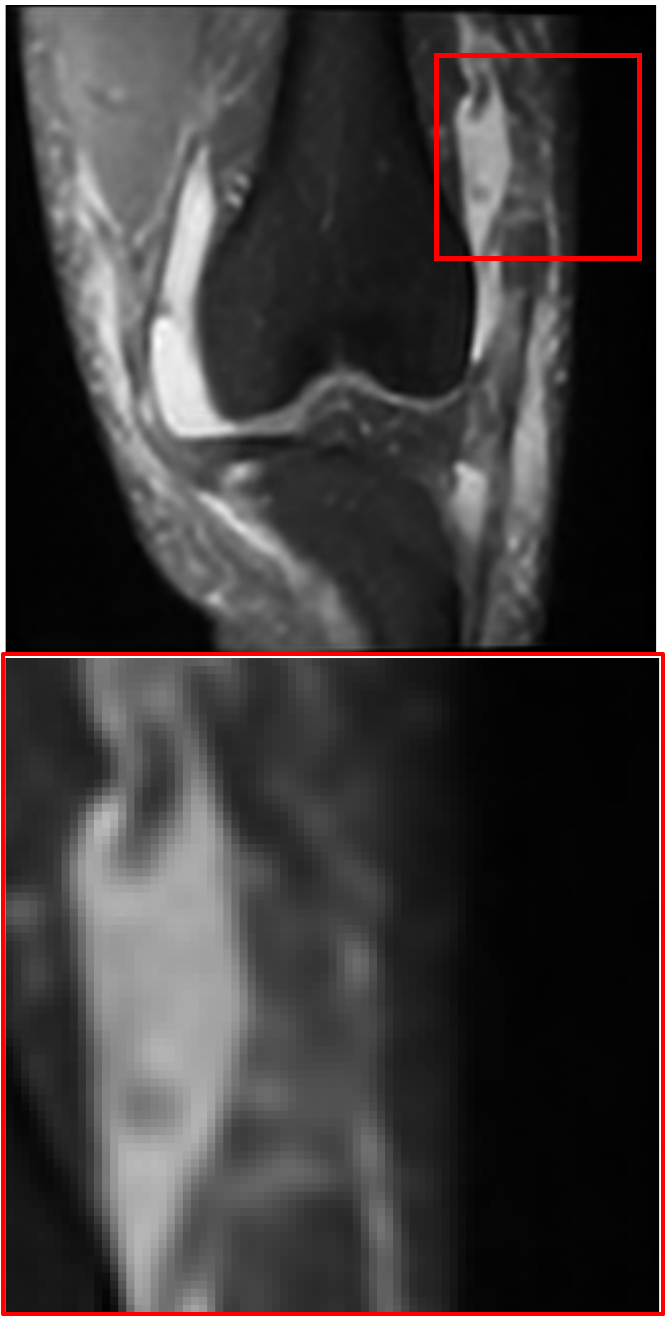} \\
        \includegraphics[scale=0.17]{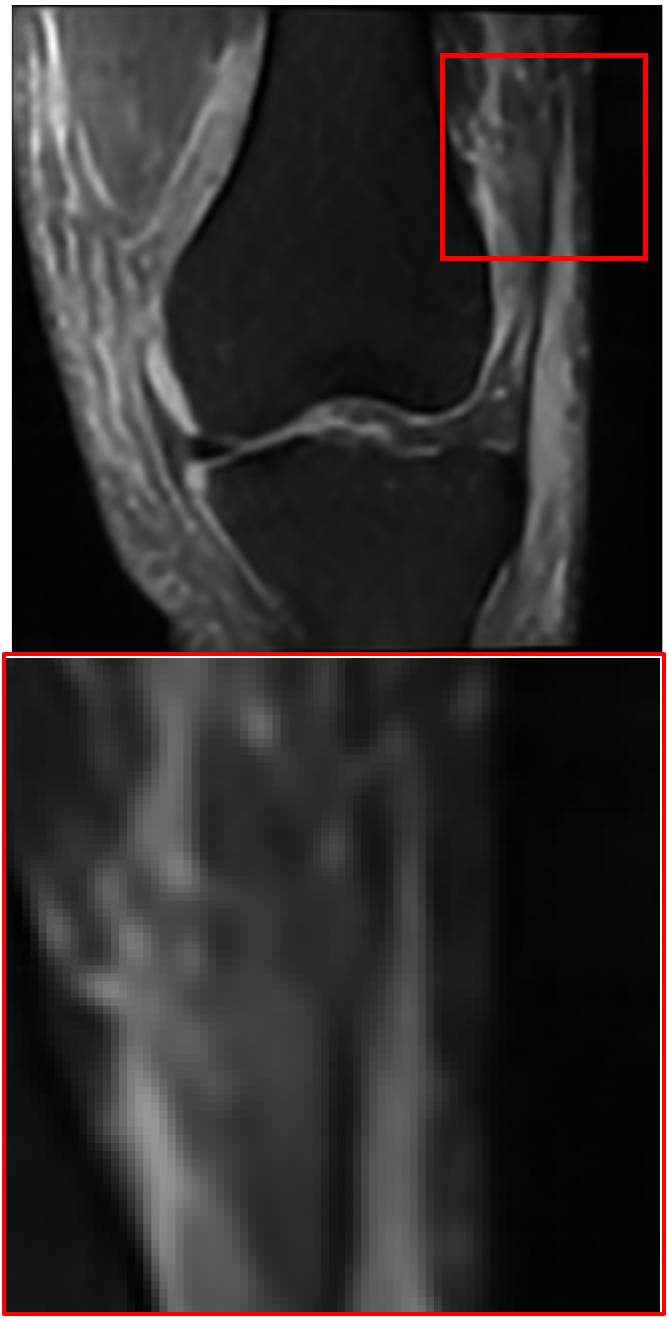}
	\end{minipage}
}
\hfill
	\subfloat[\footnotesize MINet \cite{Feng2021Multi}]{
	\begin{minipage}[b]{0.1\textwidth}
		\includegraphics[scale=0.17]{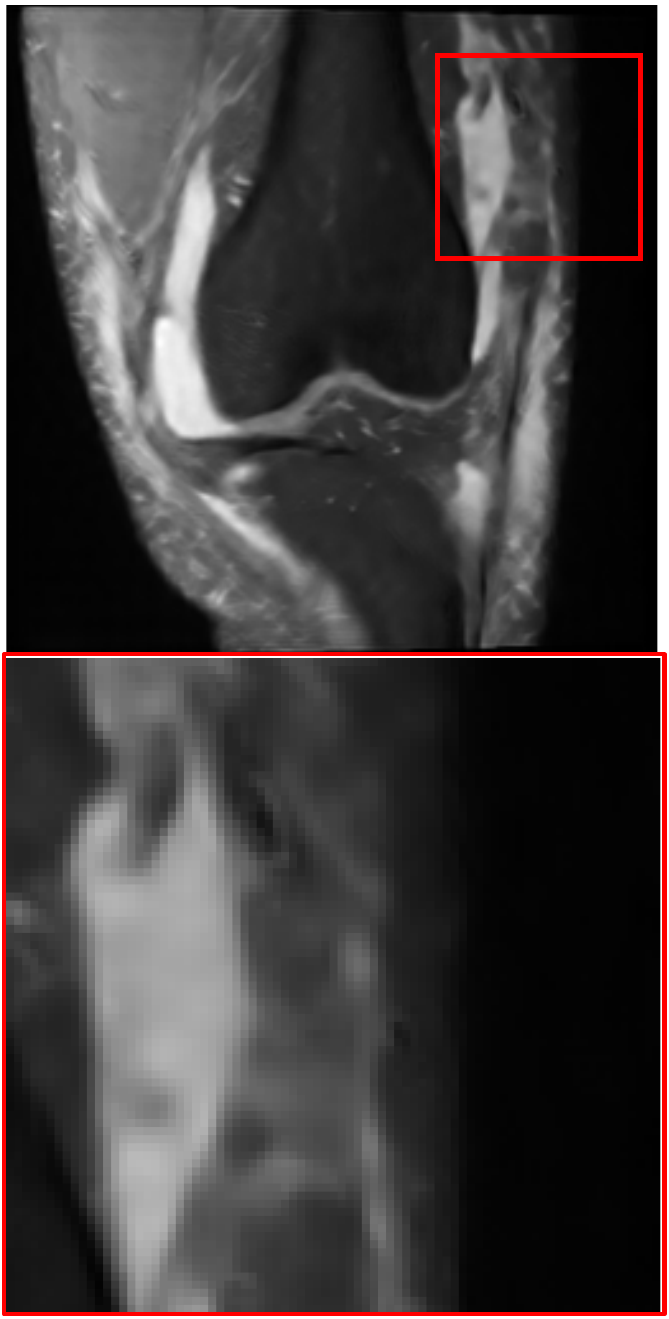} \\
        \includegraphics[scale=0.17]{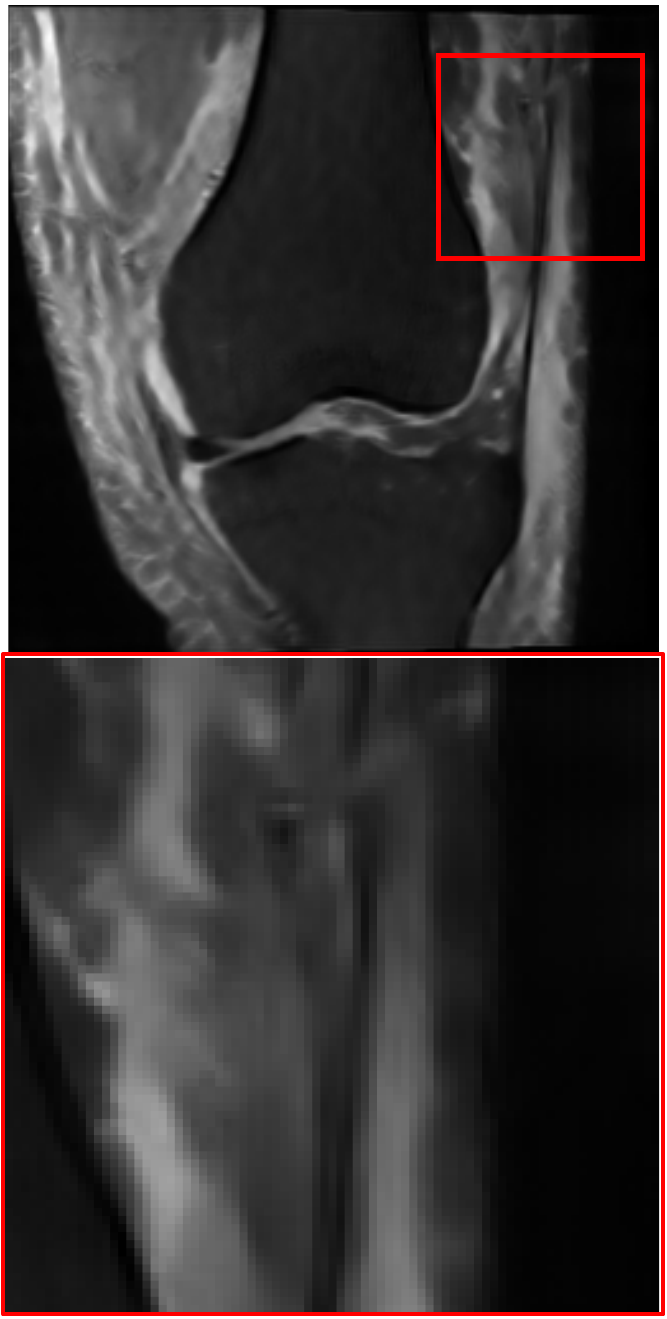}
	\end{minipage}
}
\hfill
	\subfloat[\footnotesize MASA \cite{lu2021masa}]{
	\begin{minipage}[b]{0.1\textwidth}
		\includegraphics[scale=0.17]{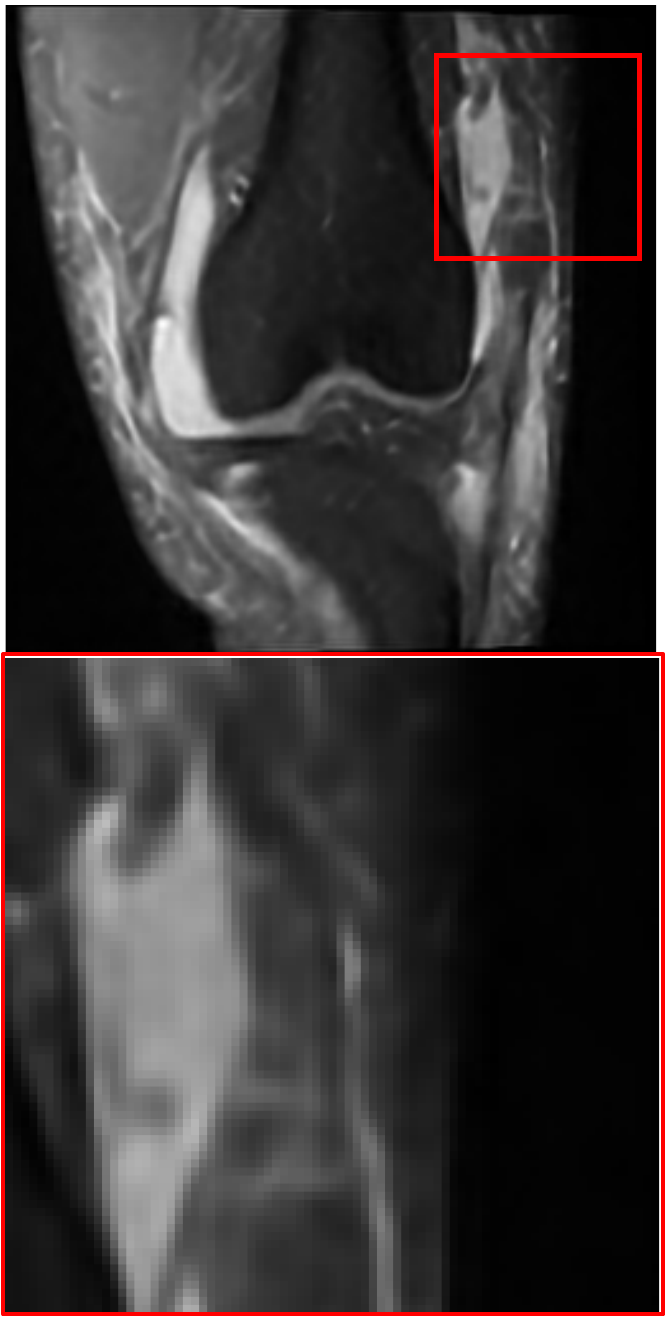} \\
        \includegraphics[scale=0.17]{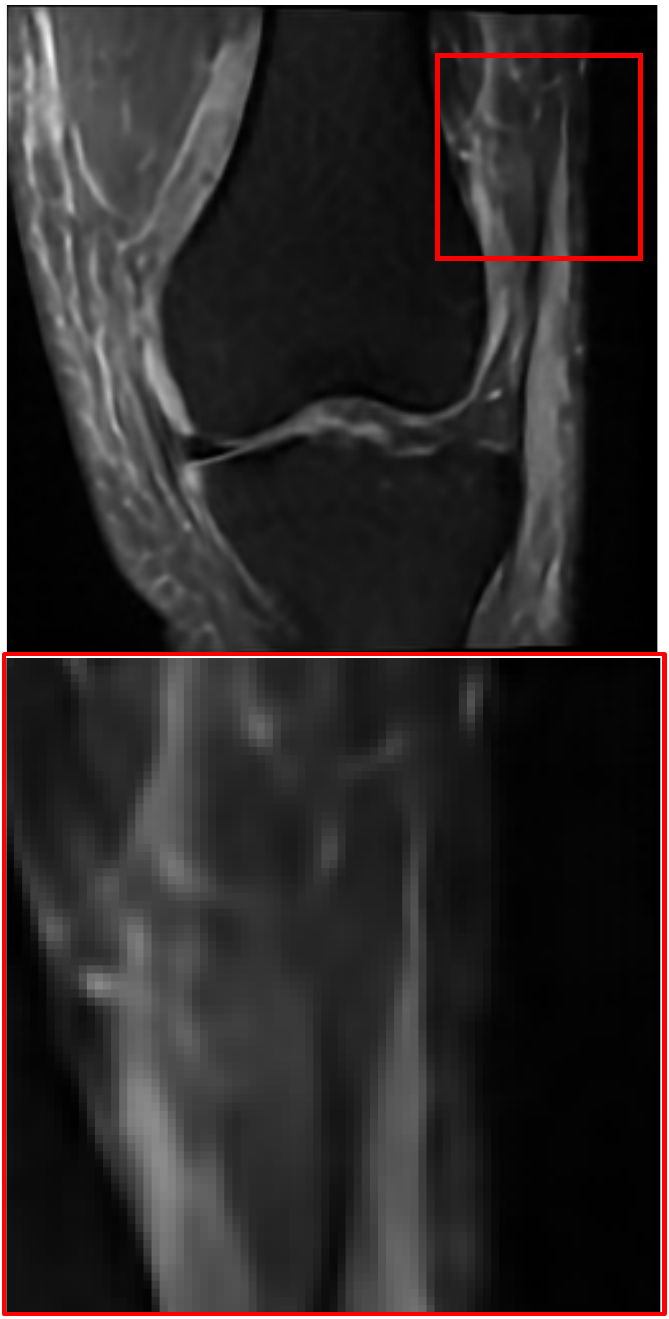}
	\end{minipage}
}
\hfill
    \subfloat[\footnotesize WavTrans \cite{Li2022Wav}]{
	\begin{minipage}[b]{0.1\textwidth}
		\includegraphics[scale=0.17]{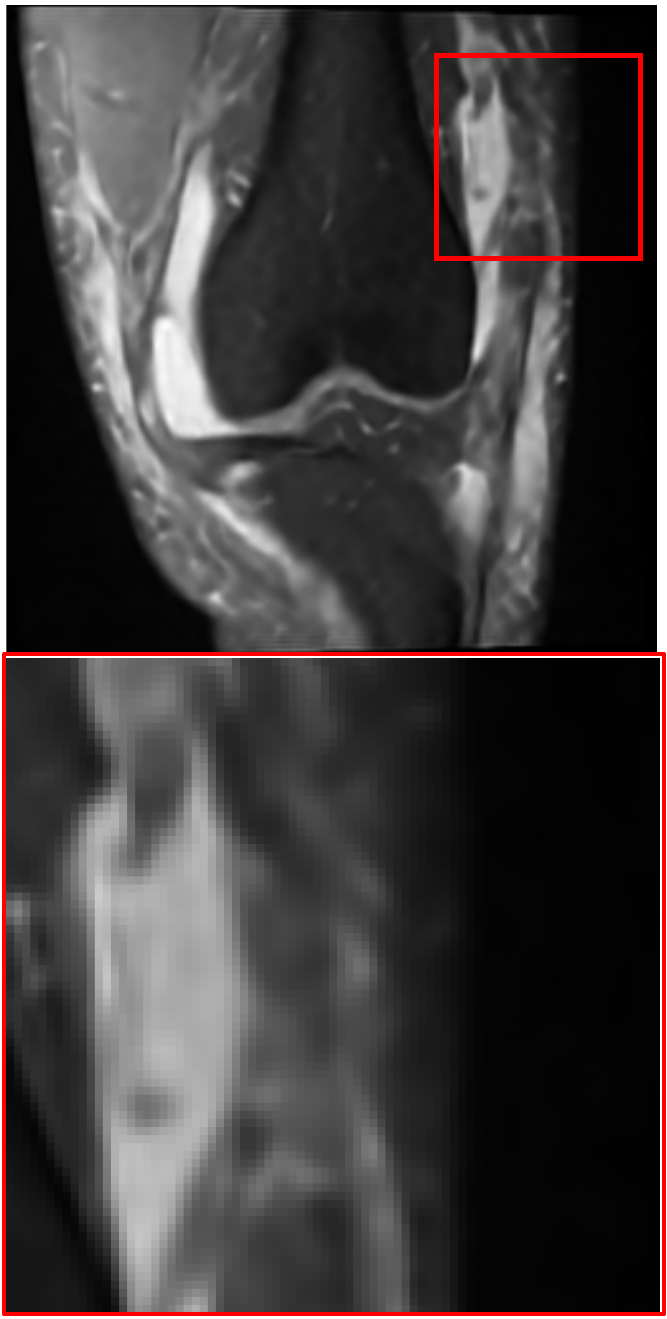} \\
        \includegraphics[scale=0.17]{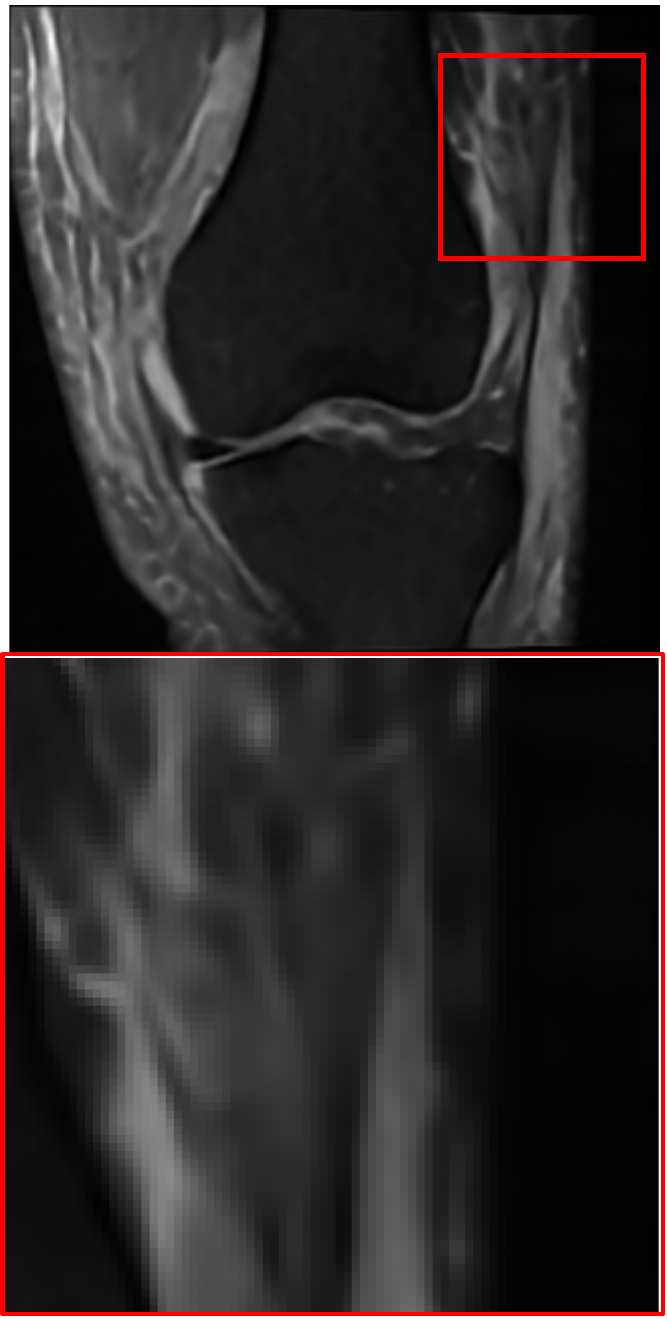}
	\end{minipage}
}
\hfill
    \subfloat[\footnotesize McMRSR \cite{Li2022Trans}]{
	\begin{minipage}[b]{0.1\textwidth}
		\includegraphics[scale=0.17]{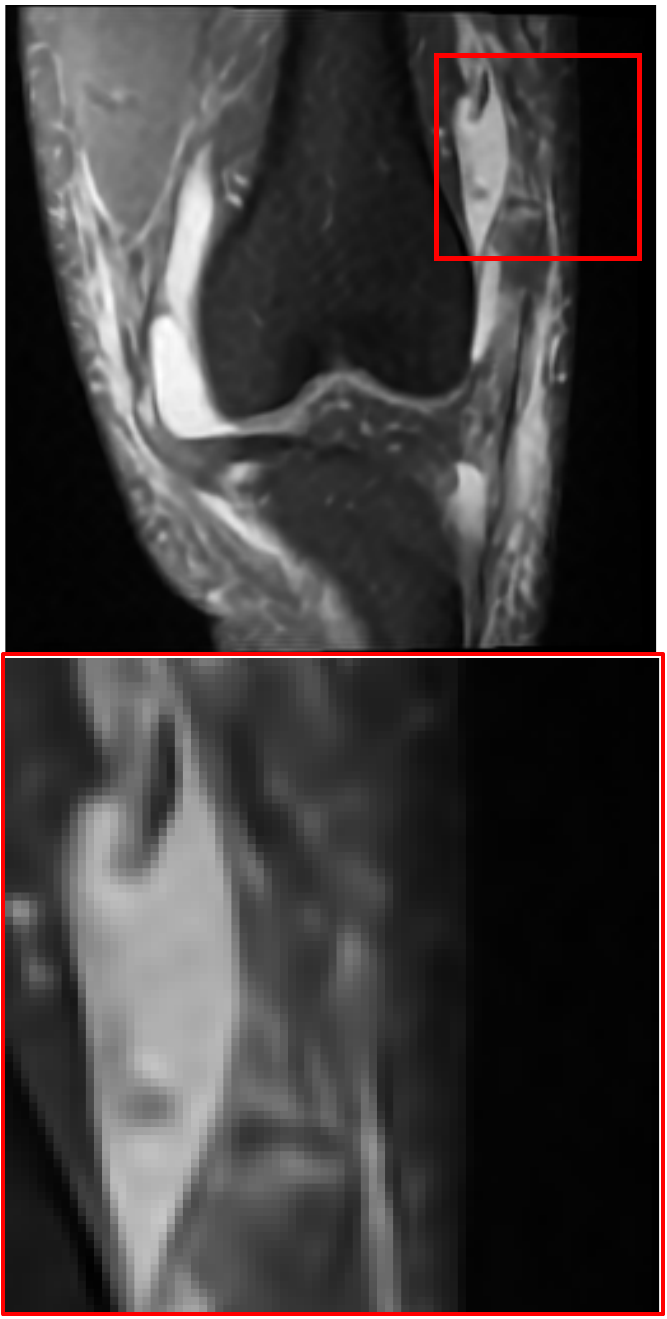} \\
        \includegraphics[scale=0.17]{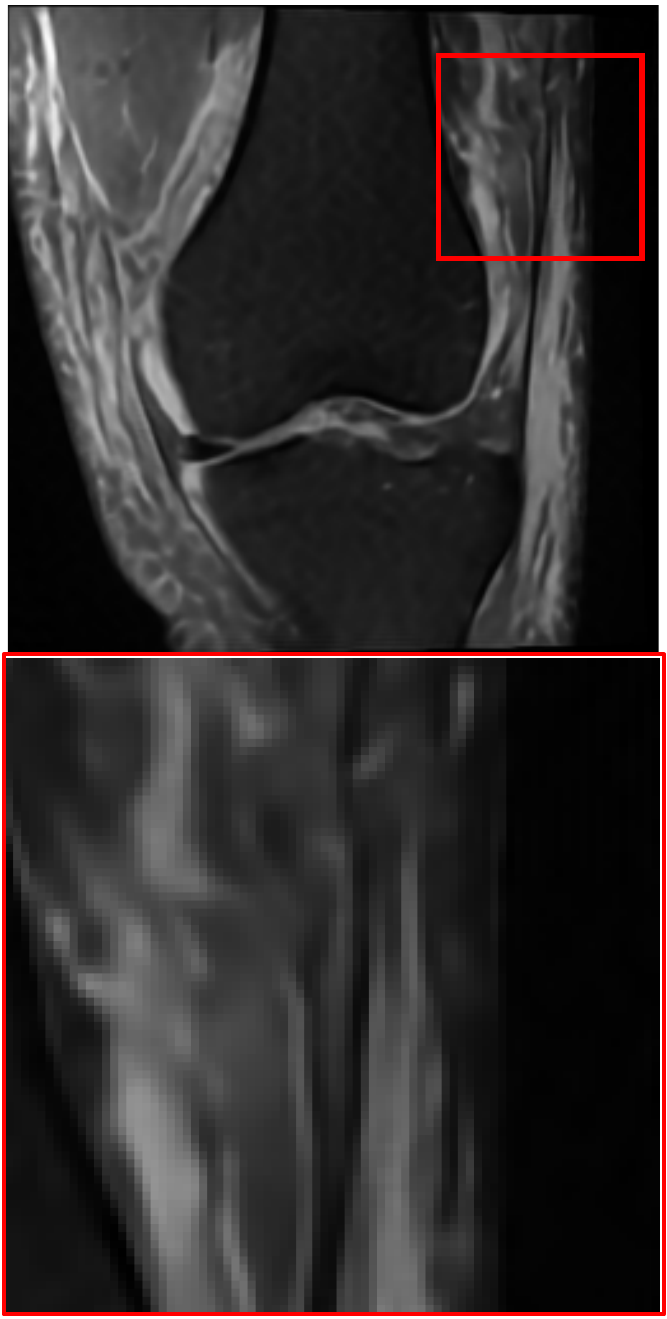}
	\end{minipage}
}
\hfill
    \subfloat[\footnotesize MC-VarNet \cite{lei2023decomposition}]{
	\begin{minipage}[b]{0.1\textwidth}
		\includegraphics[scale=0.17]{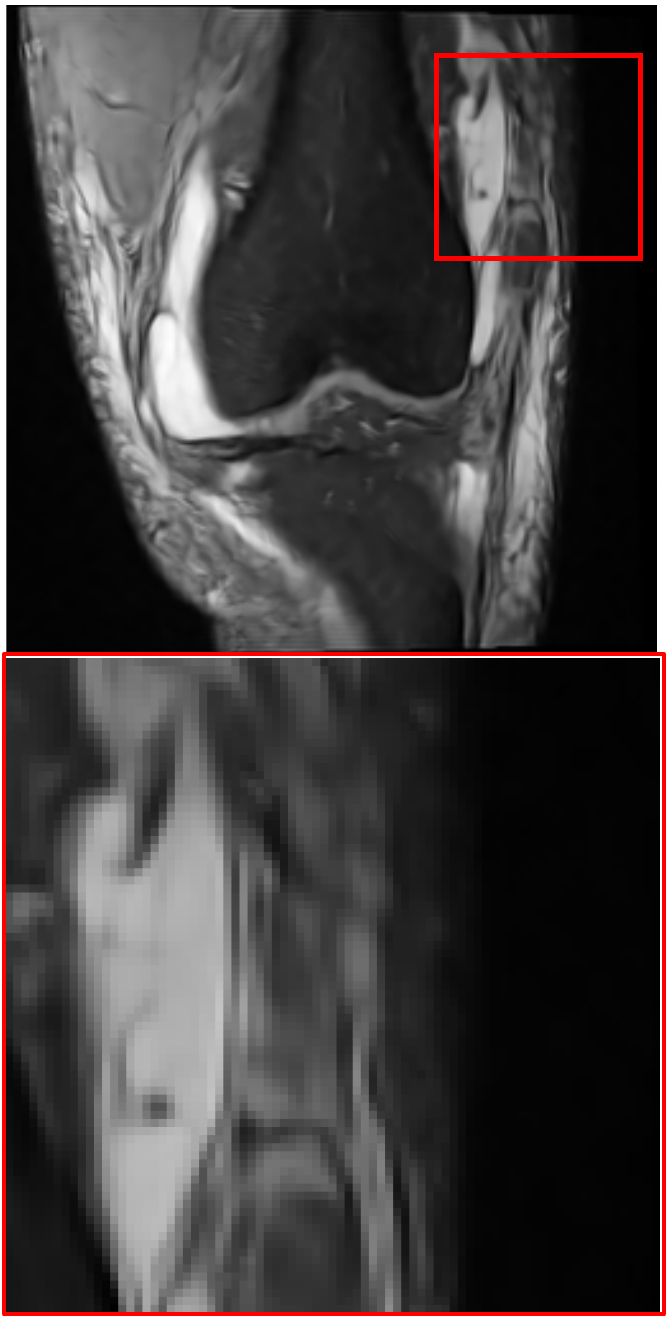}  \\
        \includegraphics[scale=0.17]{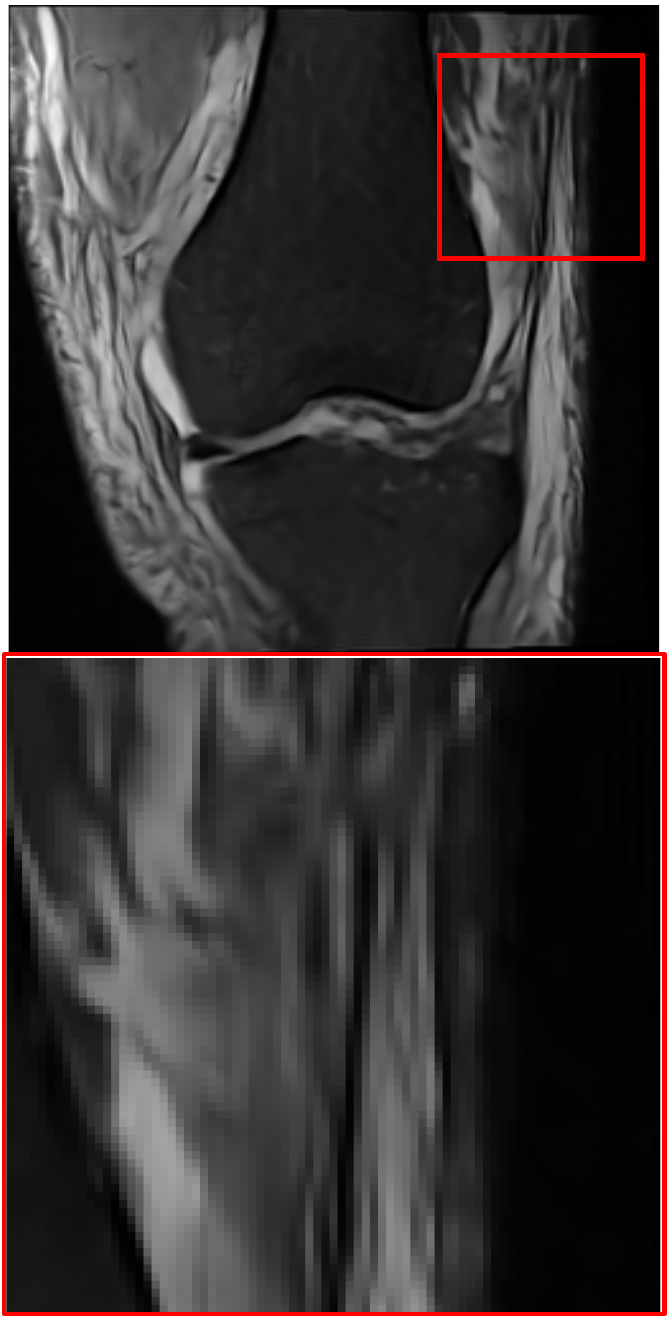}  
	\end{minipage}
}
\hfill
    \subfloat[\footnotesize DisC-Diff \cite{mao2023disc}]{
	\begin{minipage}[b]{0.1\textwidth}
		\includegraphics[scale=0.17]{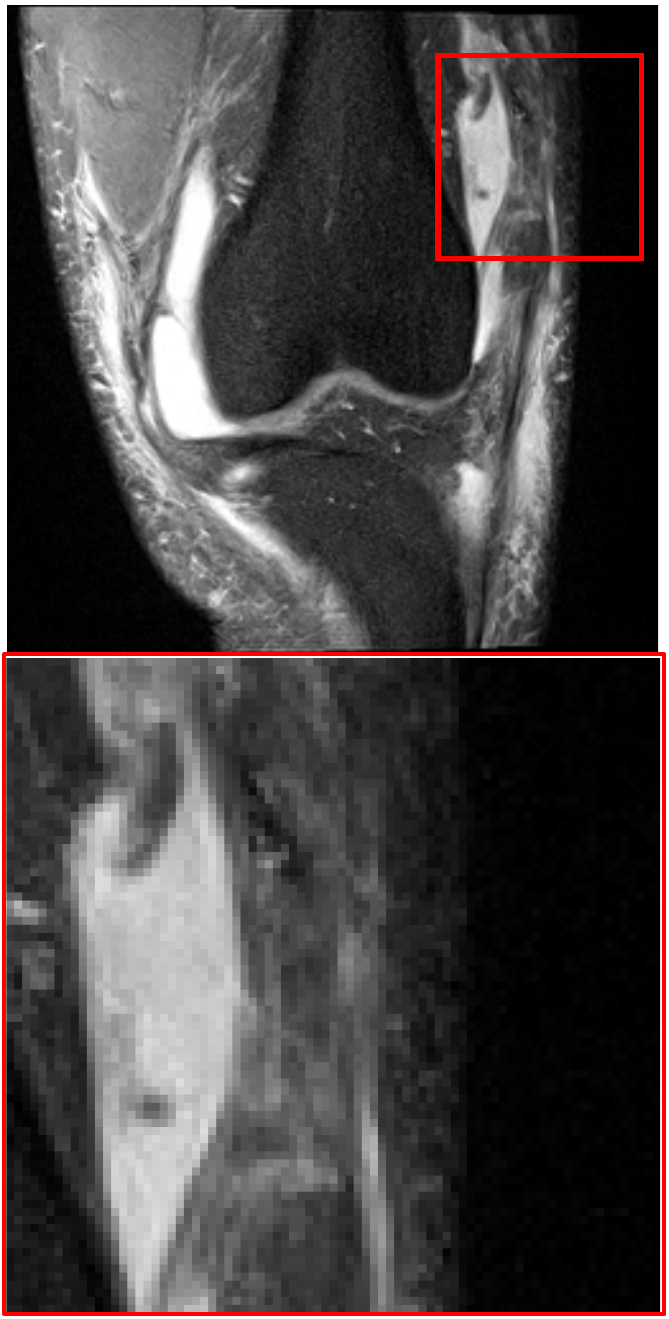} \\
        \includegraphics[scale=0.17]{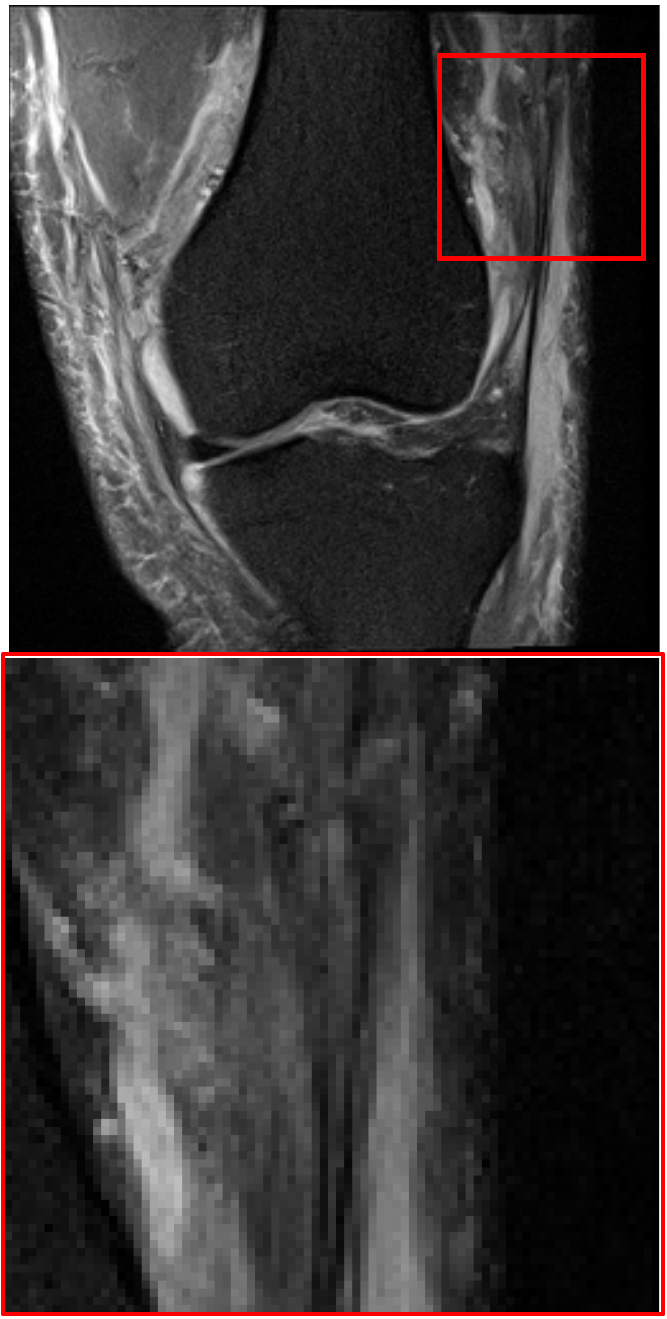}
	\end{minipage}
}
\hfill
    \subfloat[\footnotesize Ours]{
	\begin{minipage}[b]{0.1\textwidth}
		\includegraphics[scale=0.17]{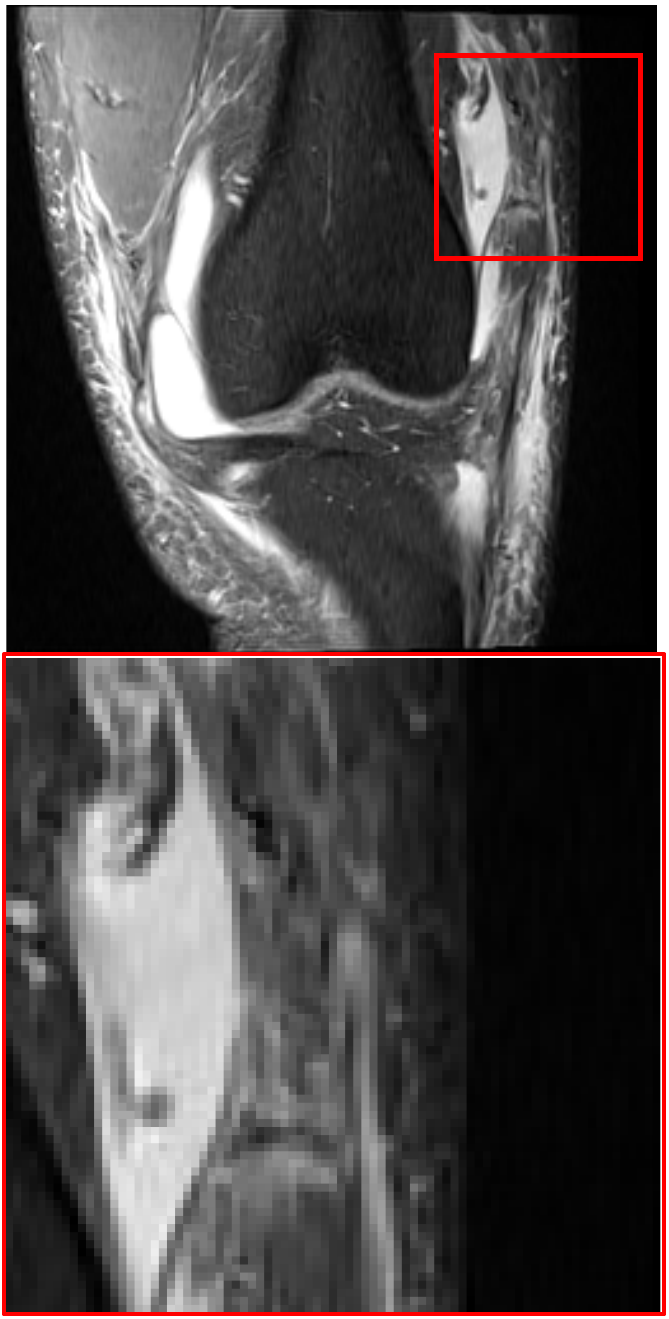} \\
        \includegraphics[scale=0.17]{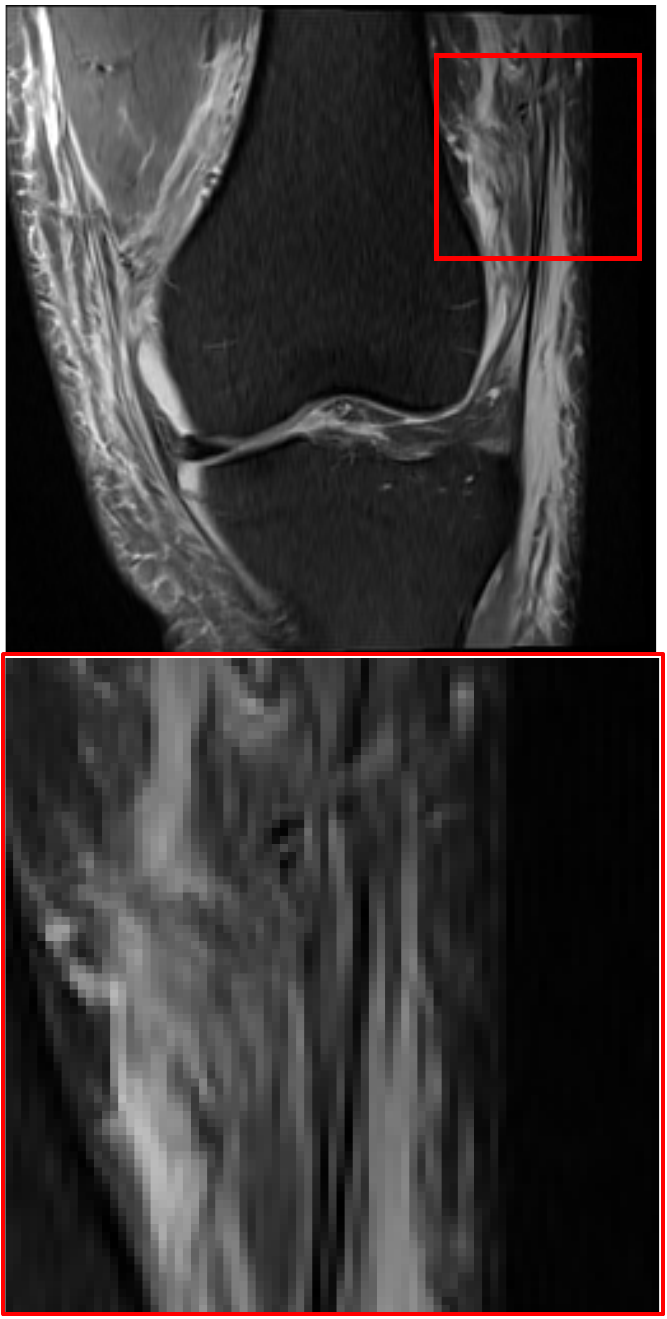}
	\end{minipage}
}
\hfill
    \subfloat[\footnotesize Target HR]{
	\begin{minipage}[b]{0.1\textwidth}
		\includegraphics[scale=0.17]{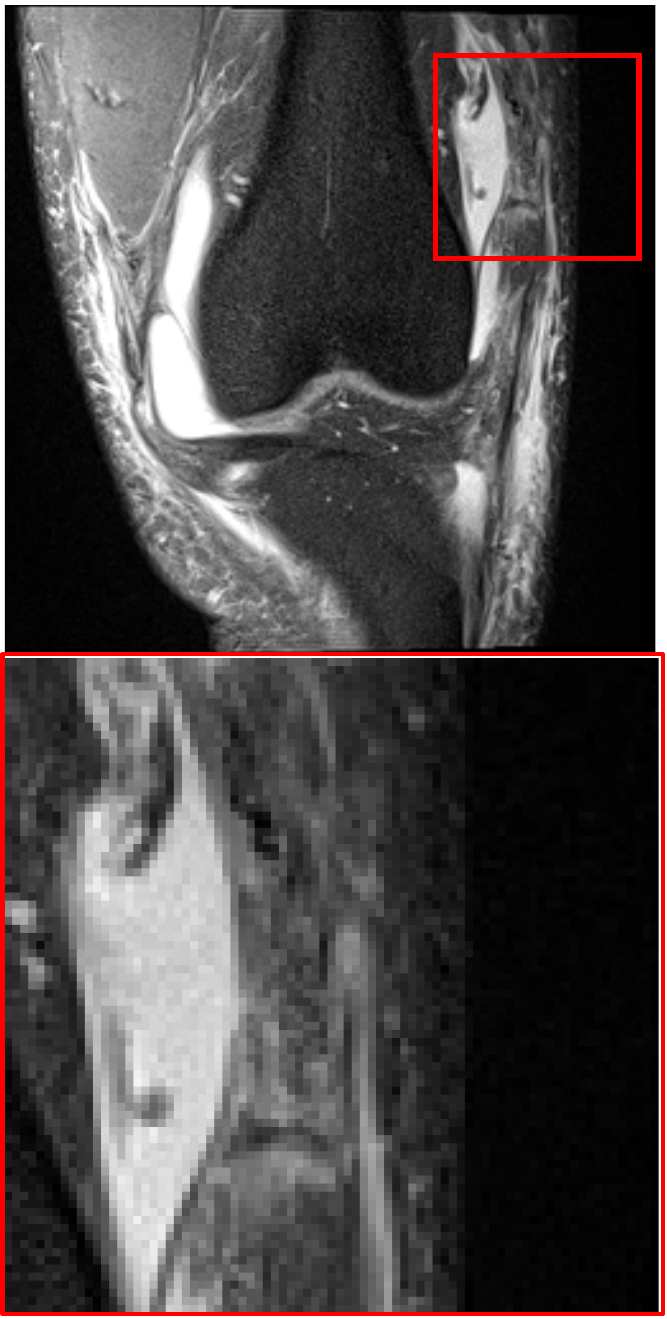} \\
        \includegraphics[scale=0.17]{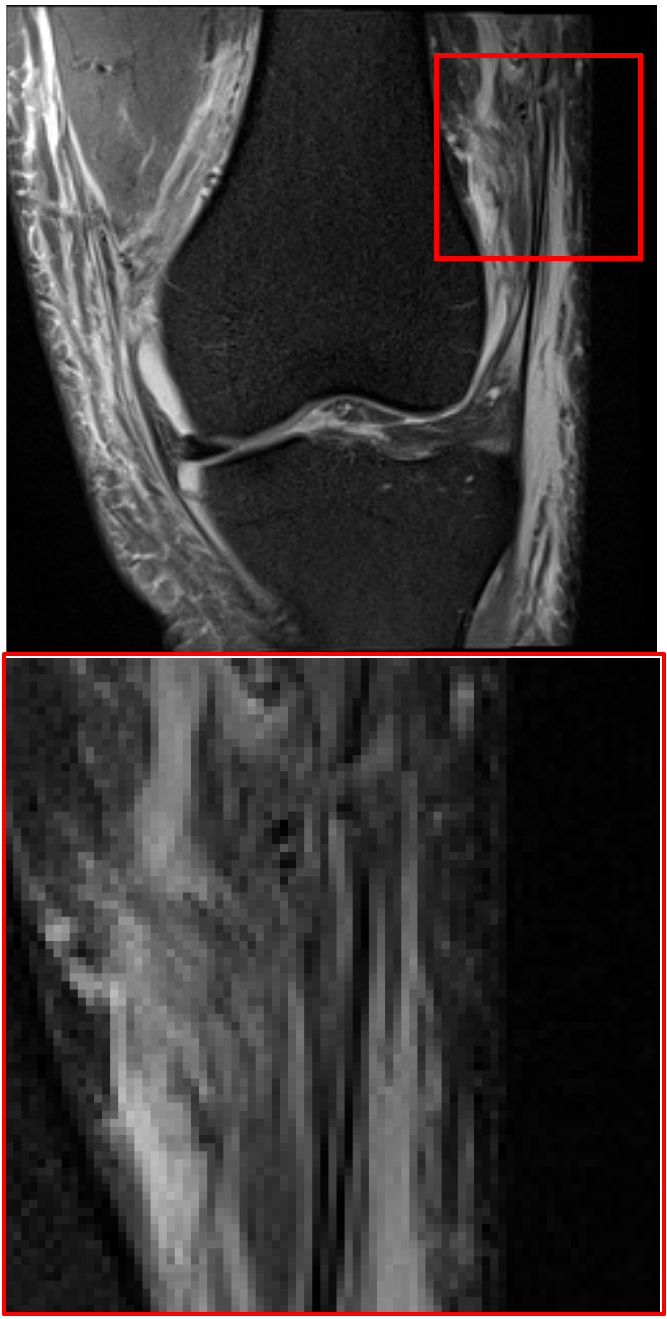}
	\end{minipage}
}
\caption{Qualitative visual comparison of various methods on the FastMRI dataset ($4\times$).}
\label{fig_knee}
\end{figure*}

\begin{figure*} [h]
	\centering
	\captionsetup[subfloat]{labelformat=empty}
	\subfloat[\footnotesize MCSR \cite{Lyu2019Multi}]{
	\begin{minipage}[b]{0.1\textwidth}
    	\includegraphics[scale=0.17]{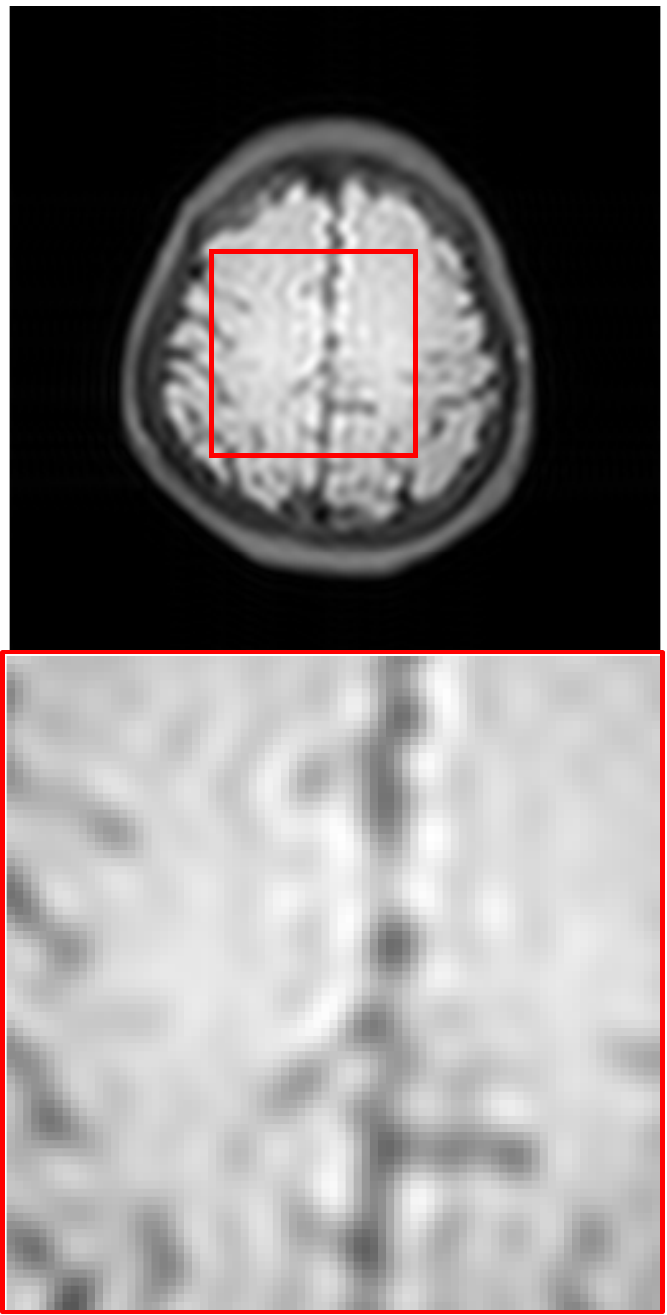} \\
        \includegraphics[scale=0.17]{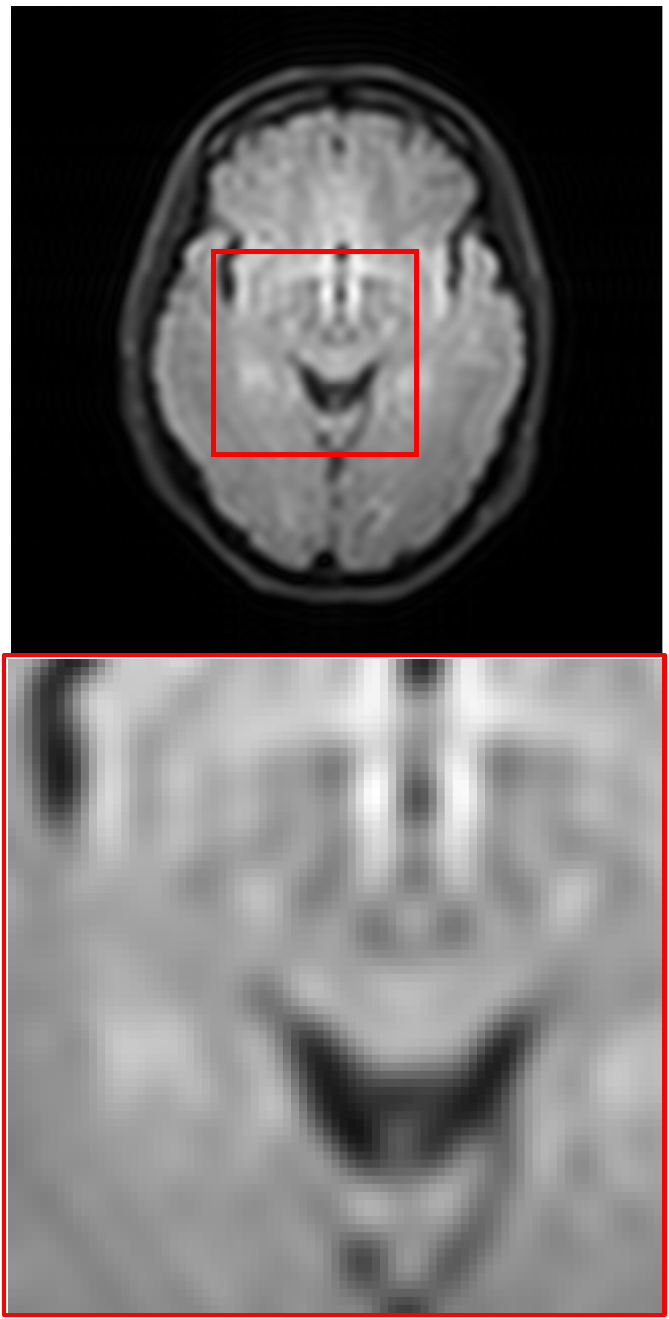}
	\end{minipage}
}
\hfill
	\subfloat[\footnotesize MINet \cite{Feng2021Multi}]{
	\begin{minipage}[b]{0.1\textwidth}
		\includegraphics[scale=0.17]{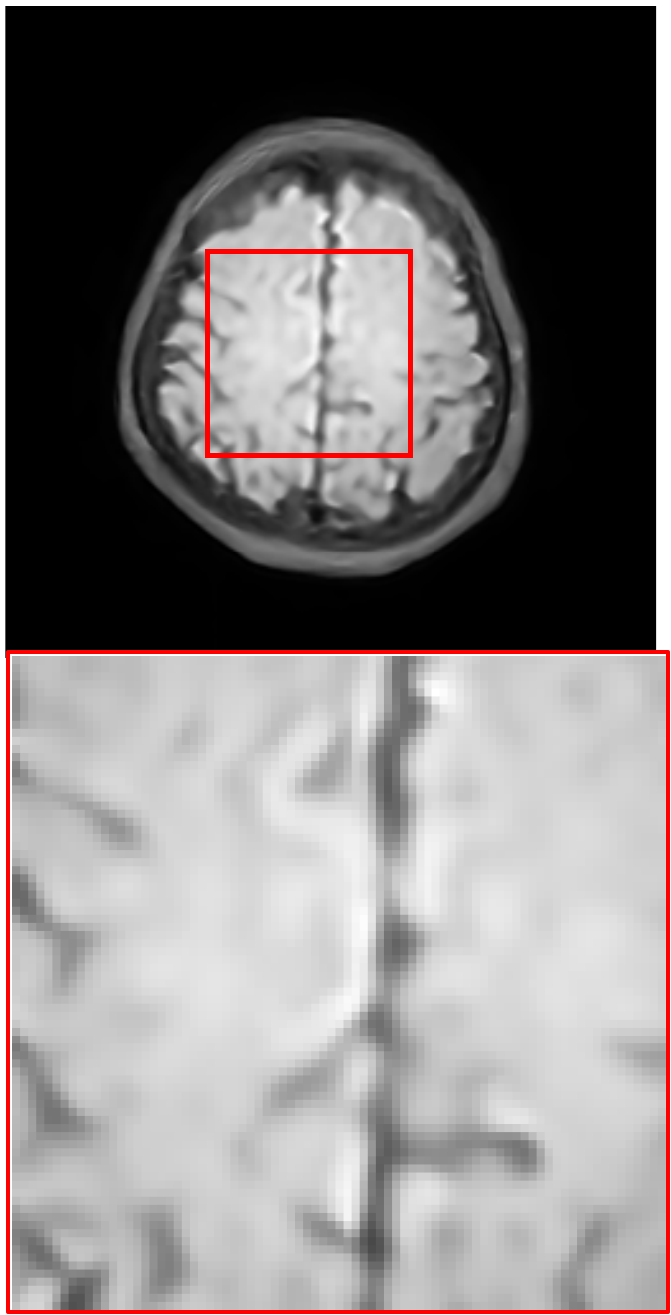} \\
        \includegraphics[scale=0.17]{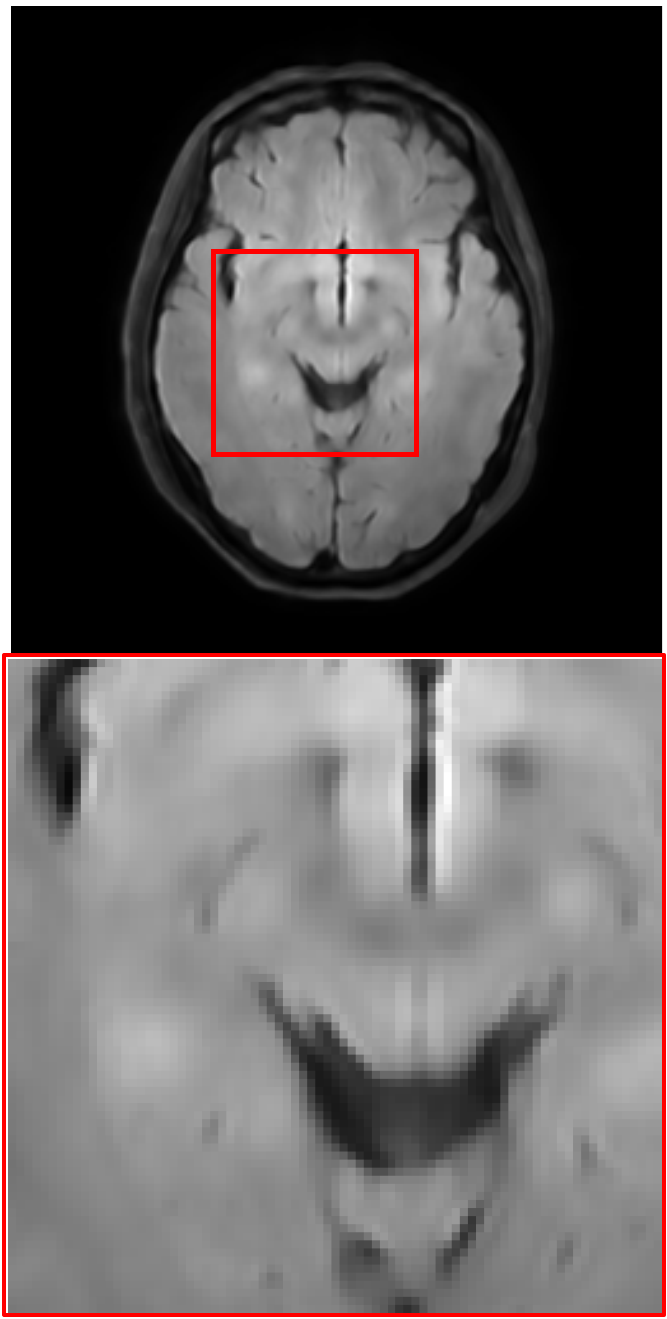}
	\end{minipage}
}
\hfill
	\subfloat[\footnotesize MASA \cite{lu2021masa}]{
	\begin{minipage}[b]{0.1\textwidth}
		\includegraphics[scale=0.17]{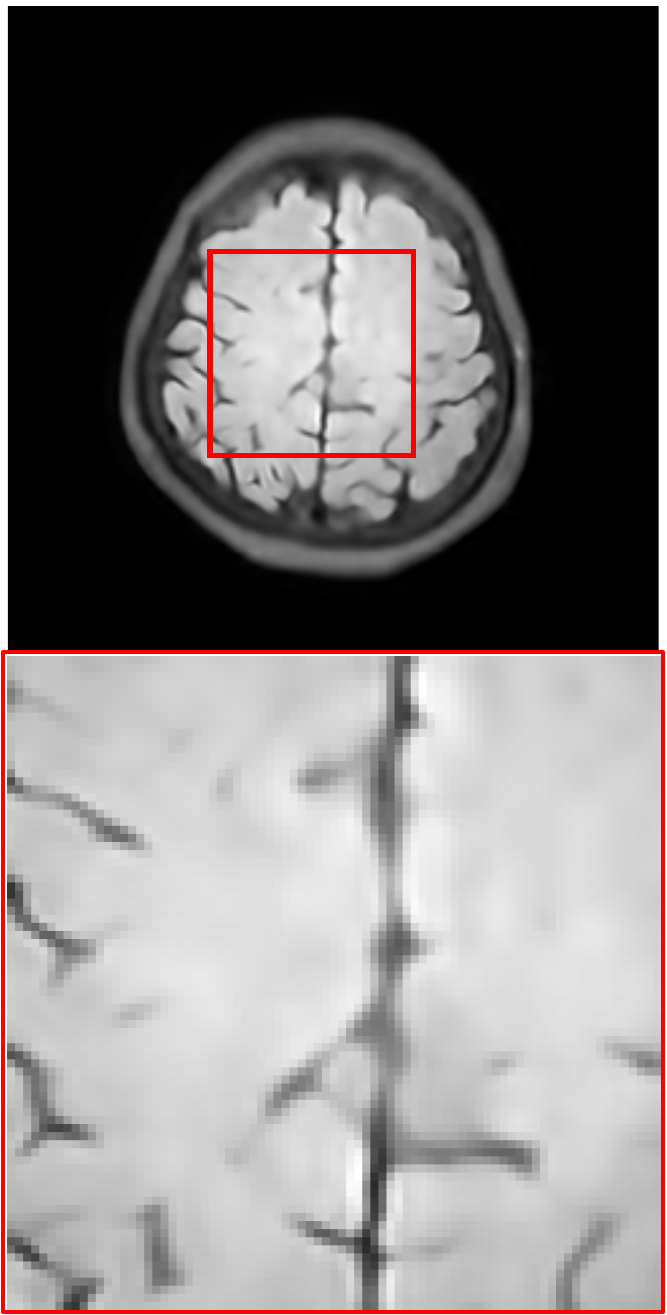} \\
        \includegraphics[scale=0.17]{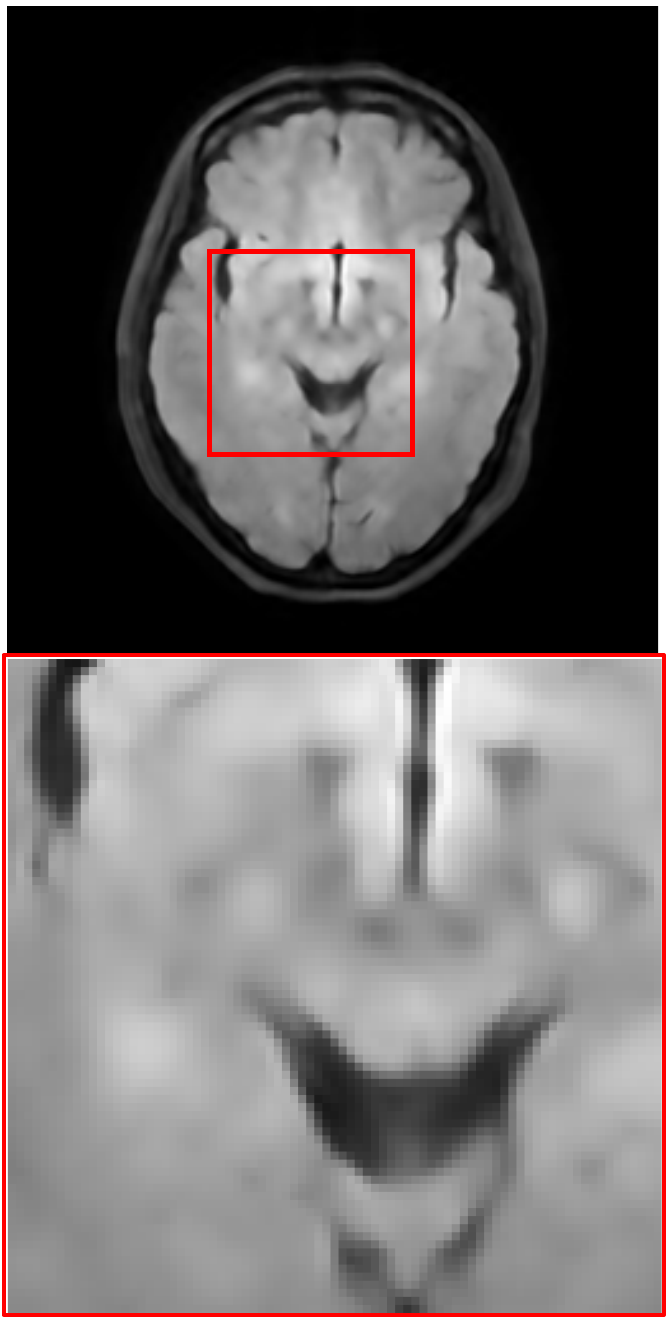}
	\end{minipage}
}
\hfill
    \subfloat[\footnotesize WavTrans \cite{Li2022Wav}]{
	\begin{minipage}[b]{0.1\textwidth}
		\includegraphics[scale=0.17]{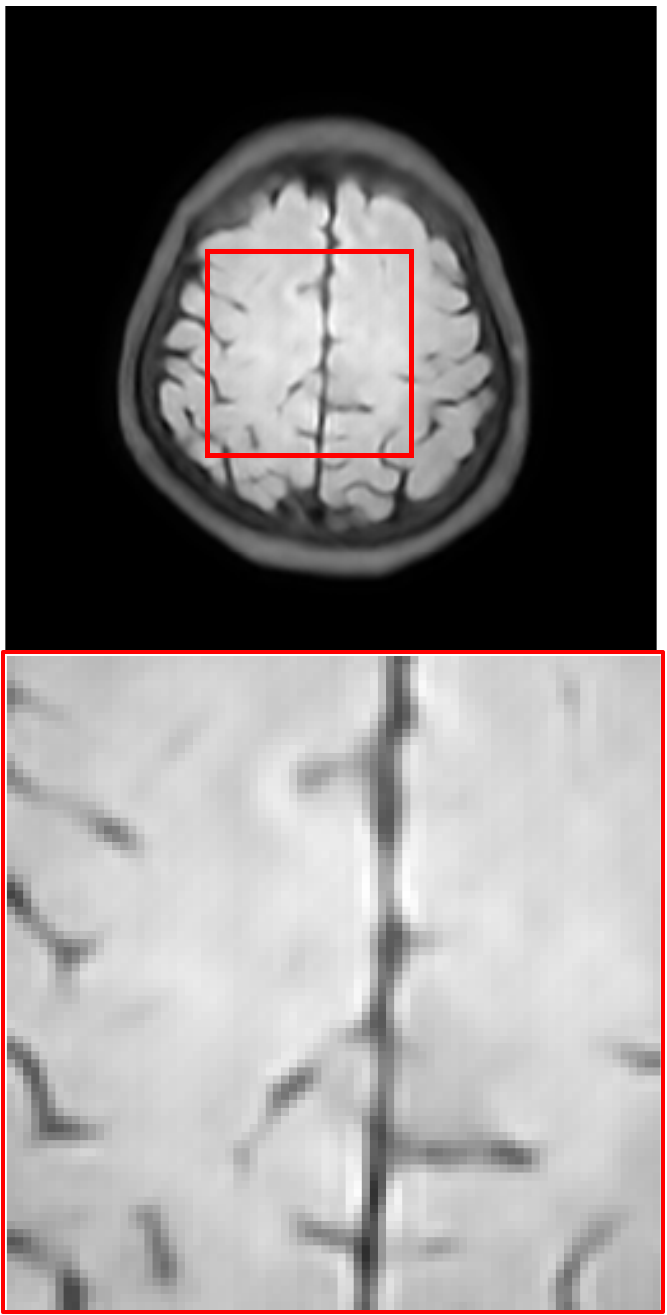} \\
        \includegraphics[scale=0.17]{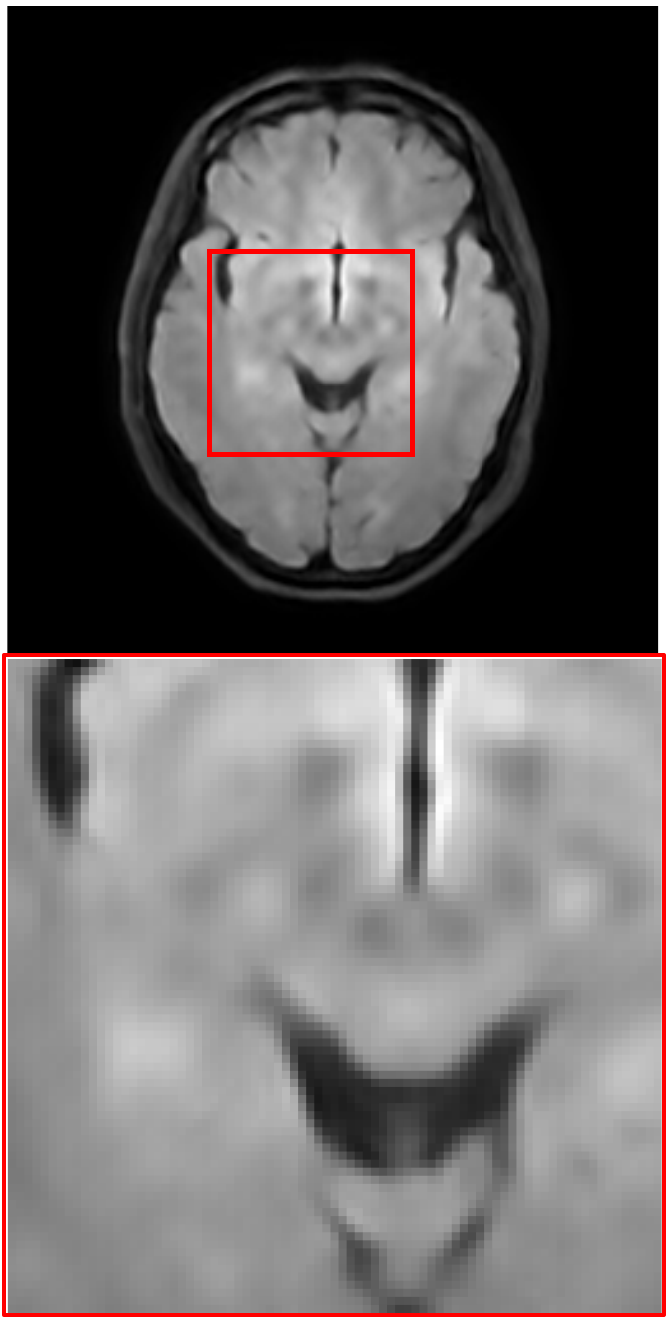}
	\end{minipage}
}
\hfill
    \subfloat[\footnotesize McMRSR \cite{Li2022Trans}]{
	\begin{minipage}[b]{0.1\textwidth}
		\includegraphics[scale=0.17]{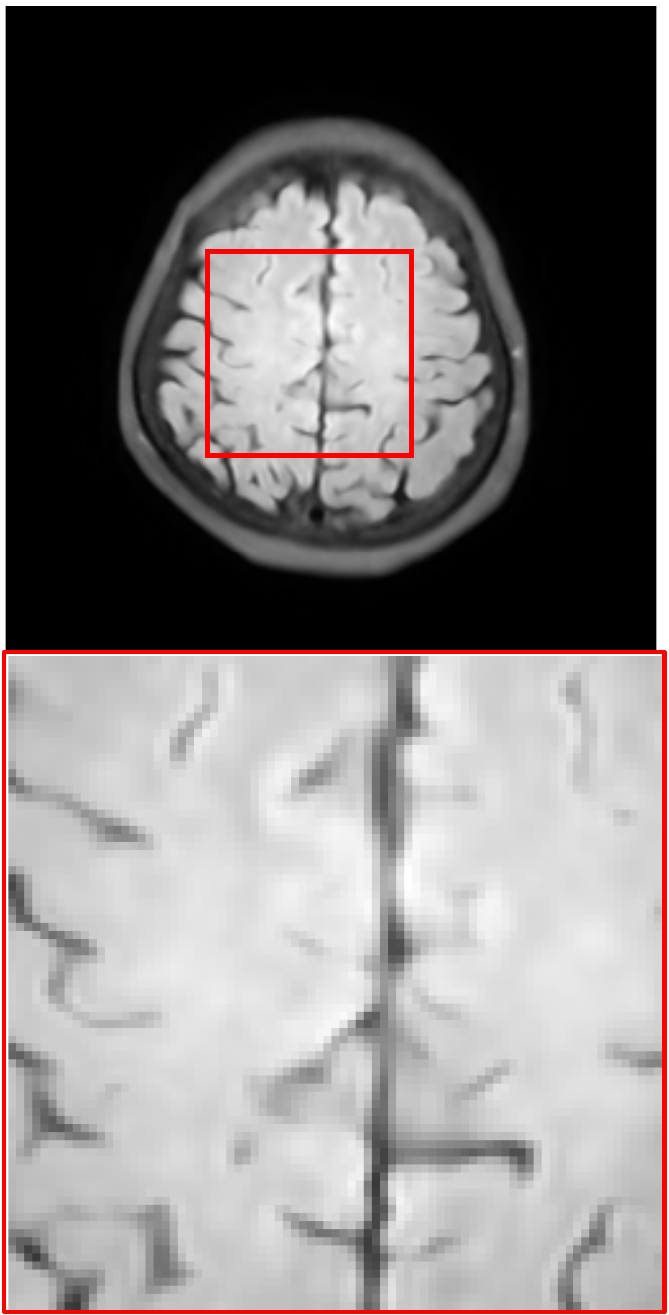} \\
        \includegraphics[scale=0.17]{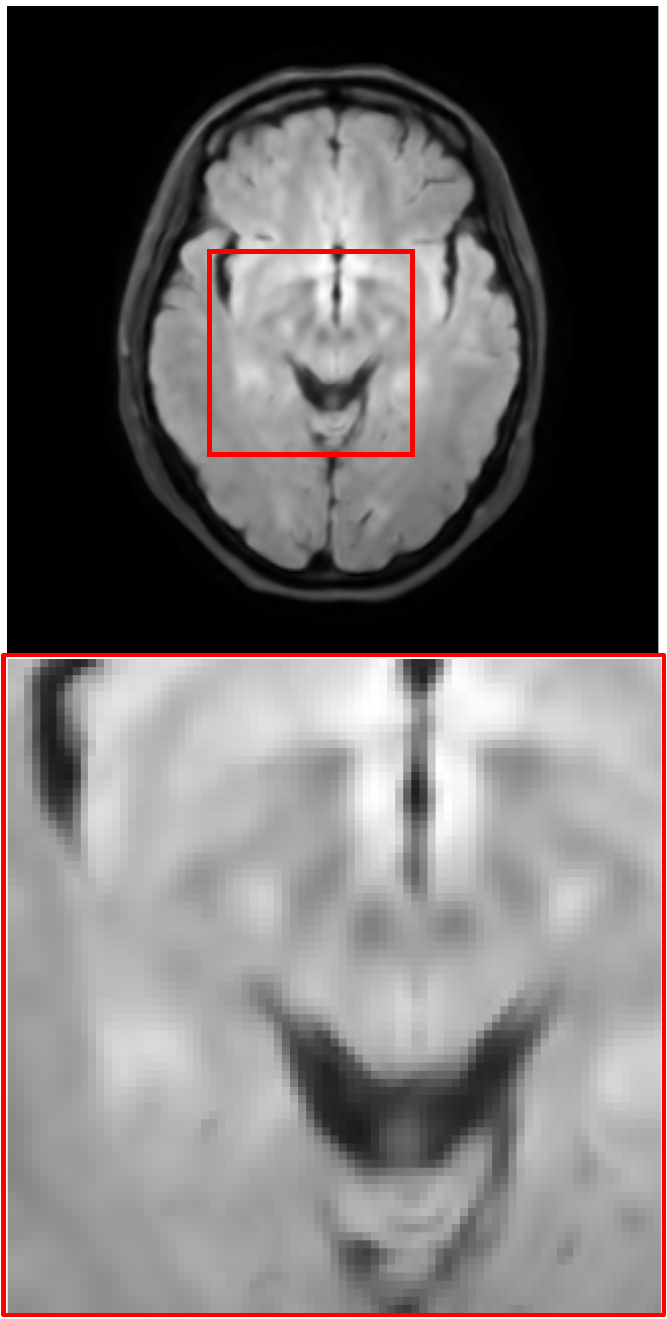}
	\end{minipage}
}
\hfill
    \subfloat[\footnotesize MC-VarNet \cite{lei2023decomposition}]{
	\begin{minipage}[b]{0.1\textwidth}
		\includegraphics[scale=0.17]{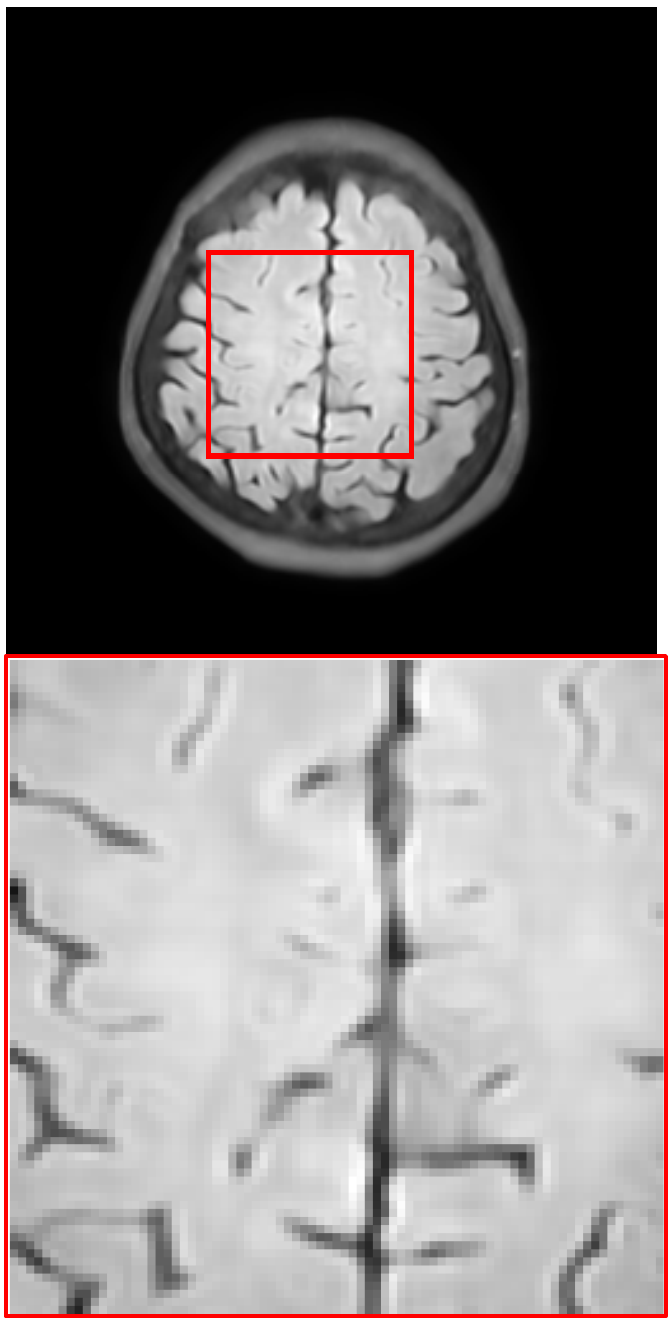}  \\
        \includegraphics[scale=0.17]{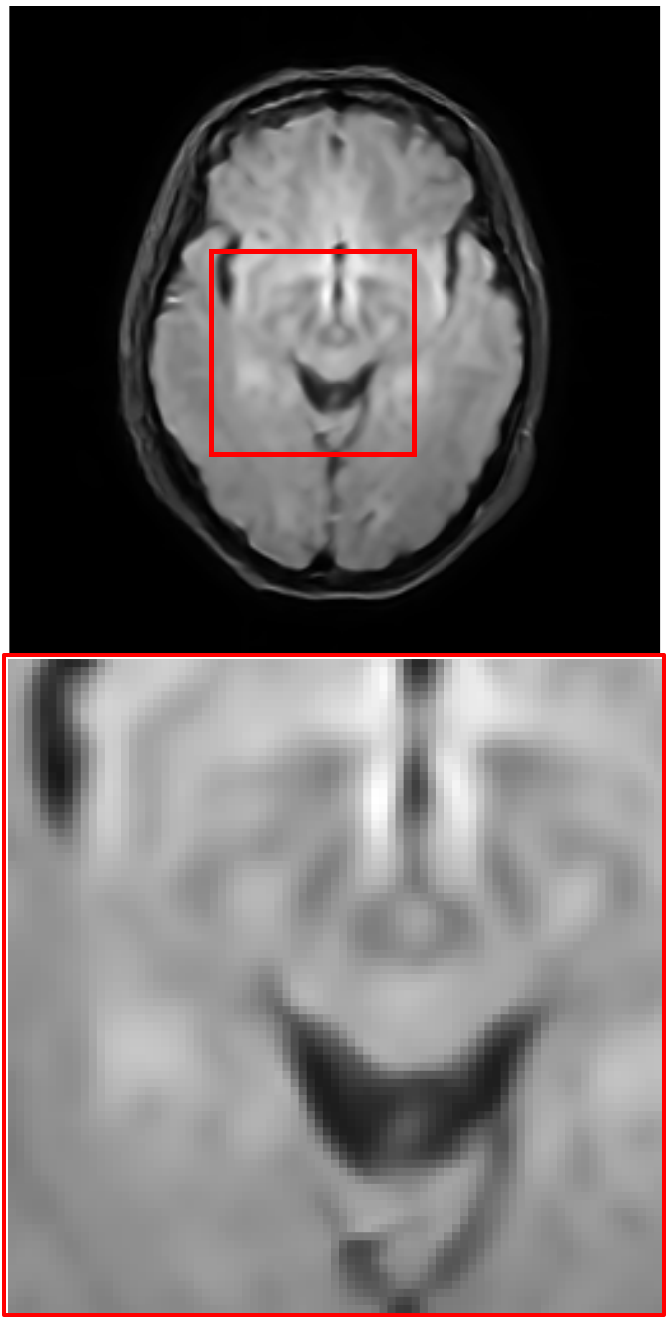}  
	\end{minipage}
}
\hfill
    \subfloat[\footnotesize DisC-Diff \cite{mao2023disc}]{
	\begin{minipage}[b]{0.1\textwidth}
		\includegraphics[scale=0.17]{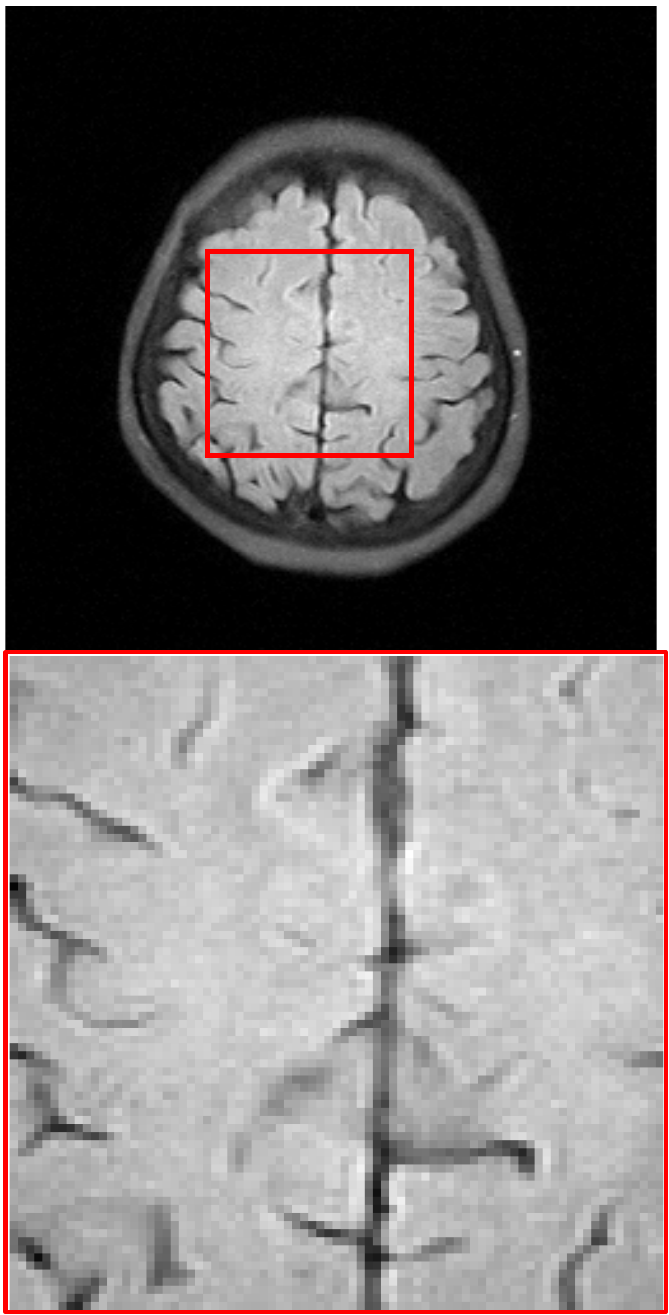} \\
        \includegraphics[scale=0.17]{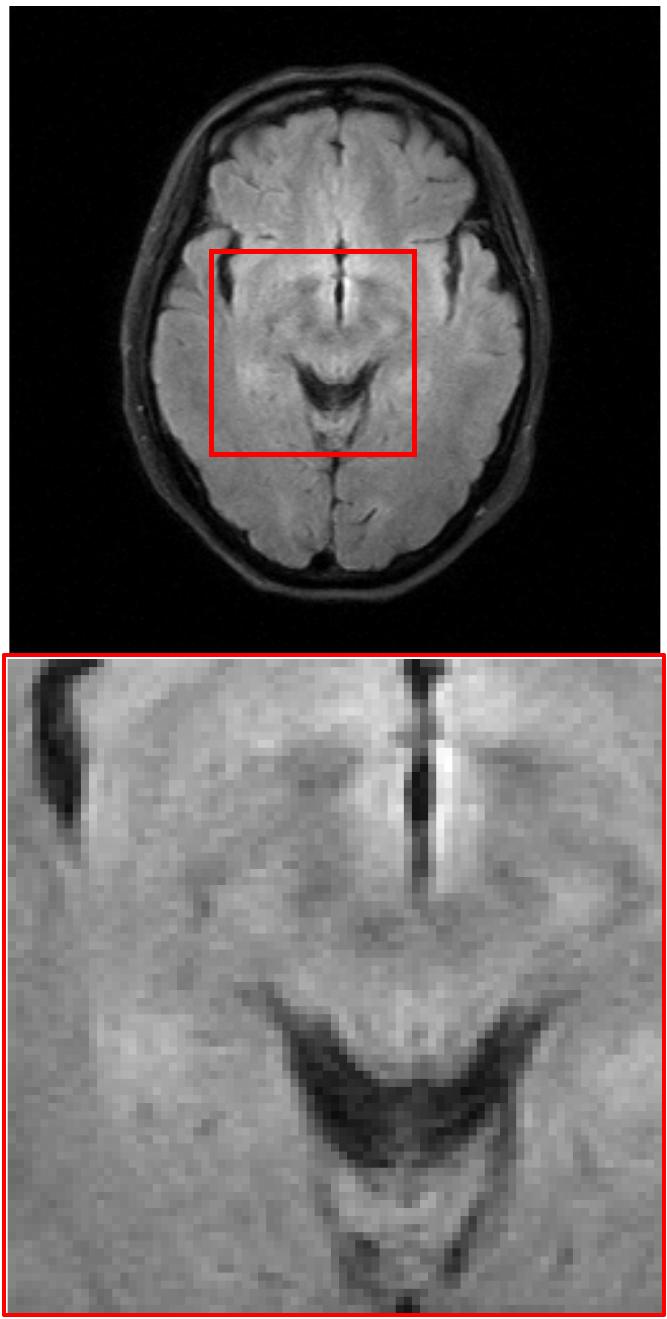}
	\end{minipage}
}
\hfill
    \subfloat[\footnotesize Ours]{
	\begin{minipage}[b]{0.1\textwidth}
		\includegraphics[scale=0.17]{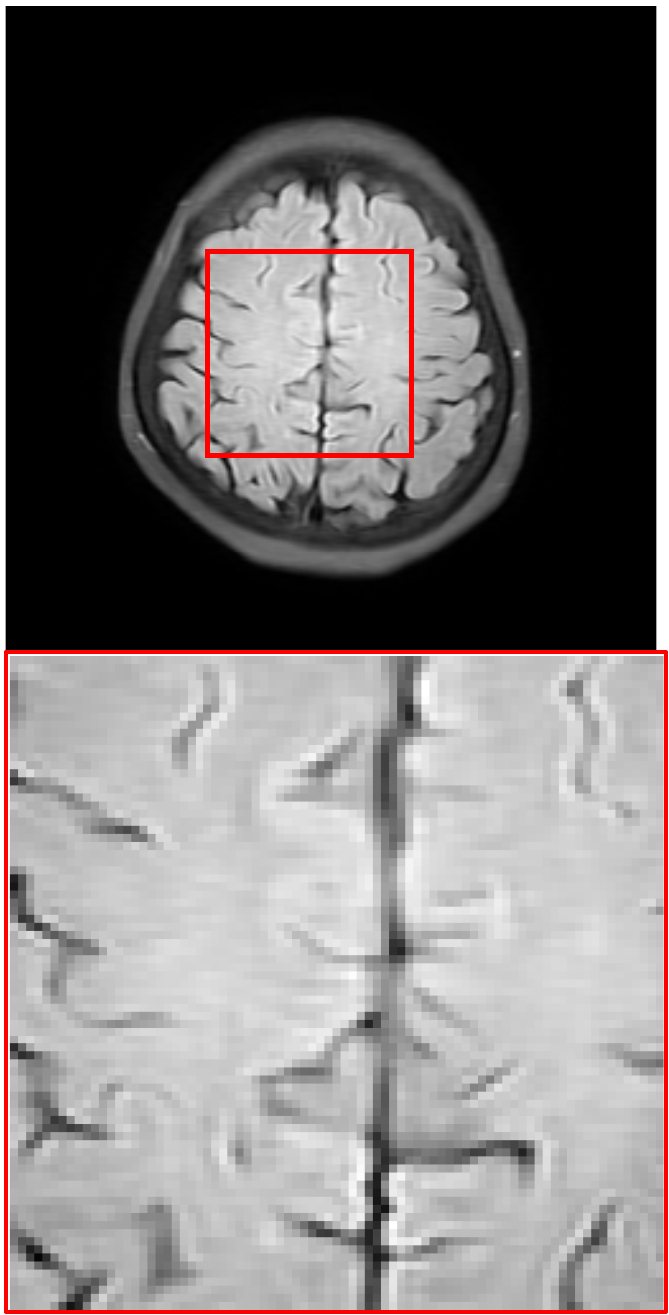} \\
        \includegraphics[scale=0.17]{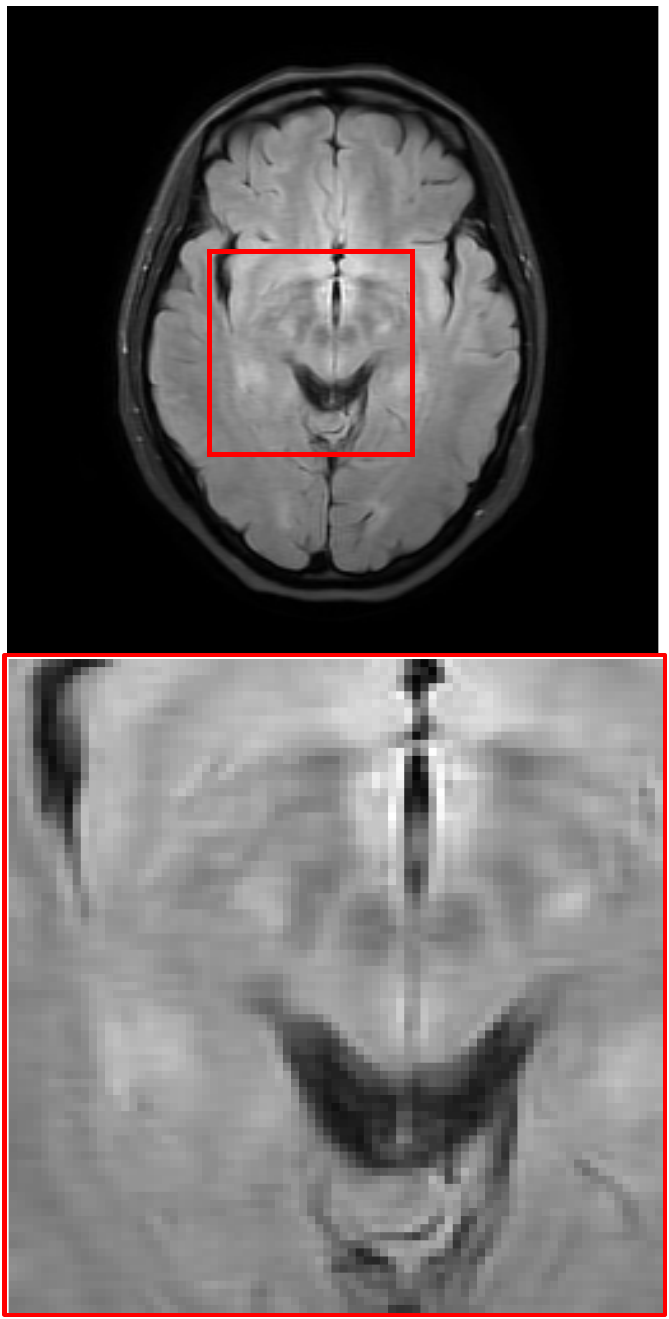}
	\end{minipage}
}
\hfill
    \subfloat[\footnotesize Target HR]{
	\begin{minipage}[b]{0.1\textwidth}
		\includegraphics[scale=0.17]{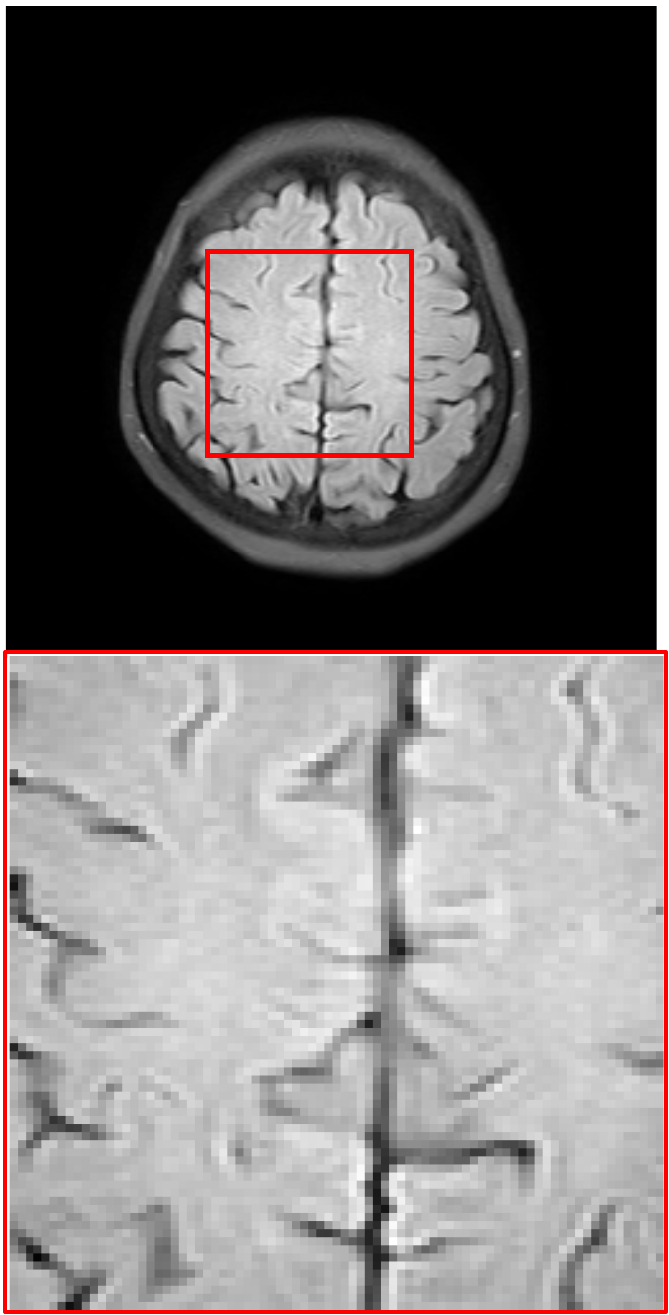} \\
        \includegraphics[scale=0.17]{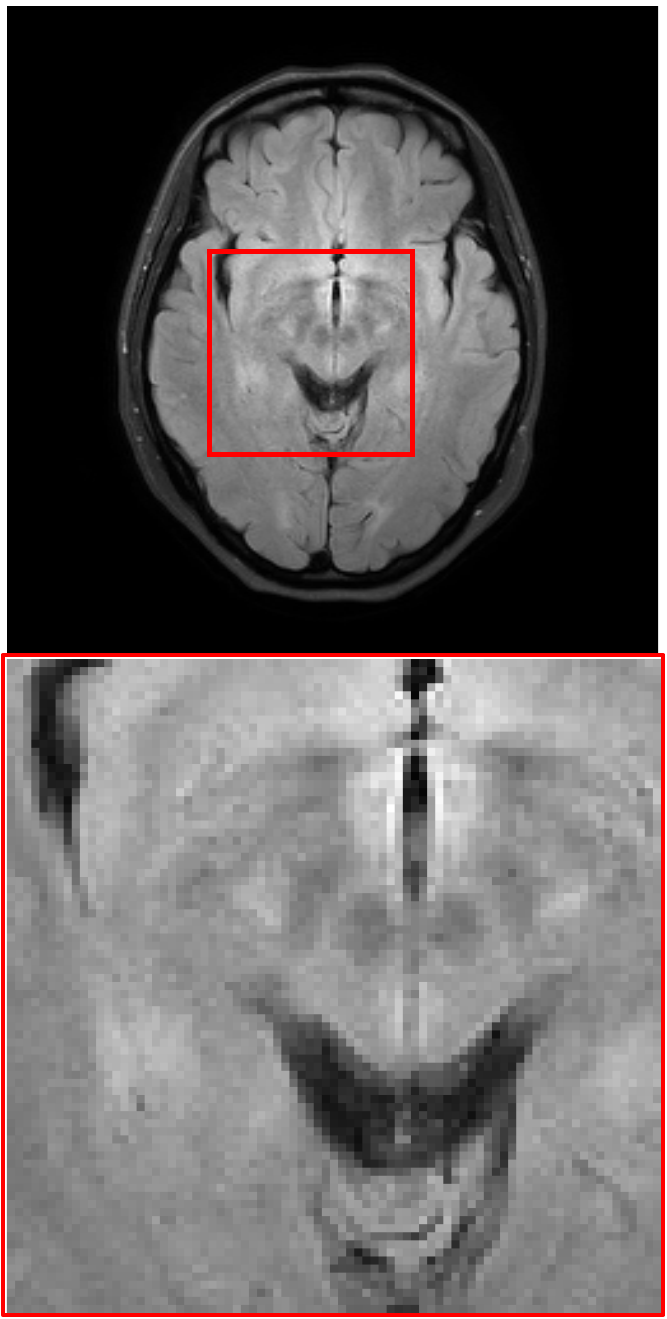}
	\end{minipage}
}
\caption{Qualitative visual comparison of various methods on the clinical brain dataset ($4\times$).}
\label{fig_brain}
\end{figure*}

\begin{figure*} [h]
	\centering
	\captionsetup[subfloat]{labelformat=empty}
	\subfloat[\footnotesize MCSR \cite{Lyu2019Multi}]{
	\begin{minipage}[b]{0.1\textwidth}
    	\includegraphics[scale=0.17]{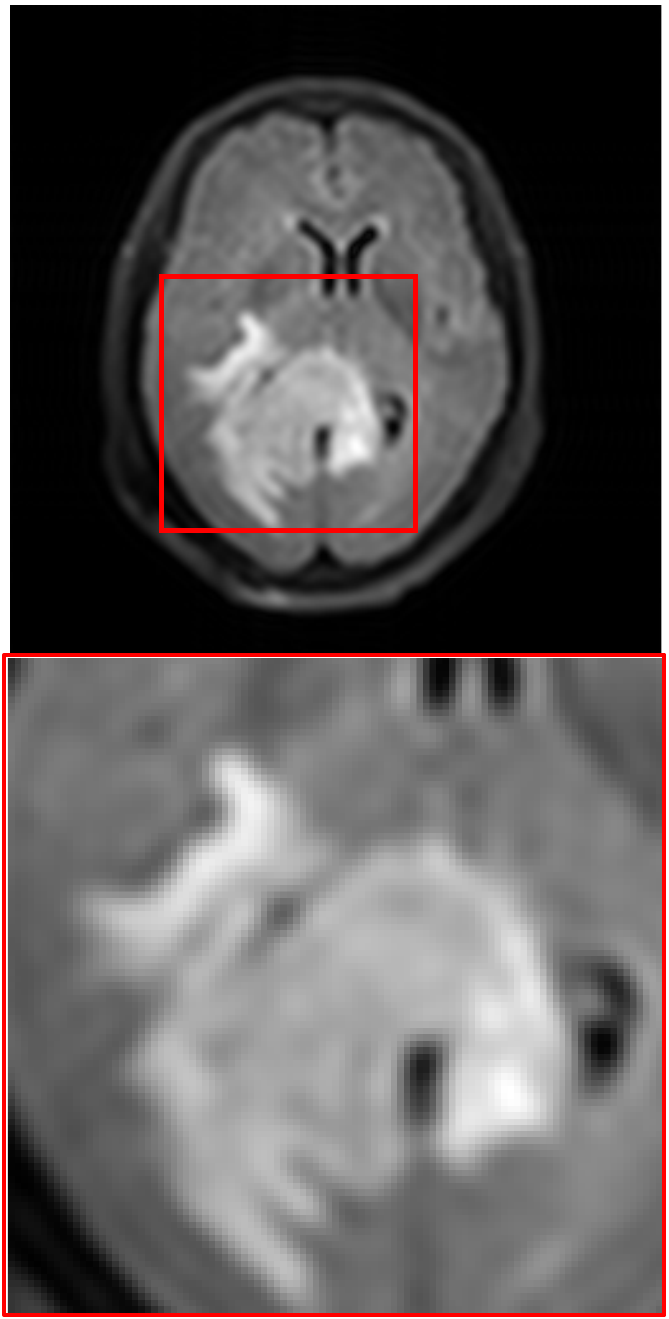} \\
        \includegraphics[scale=0.17]{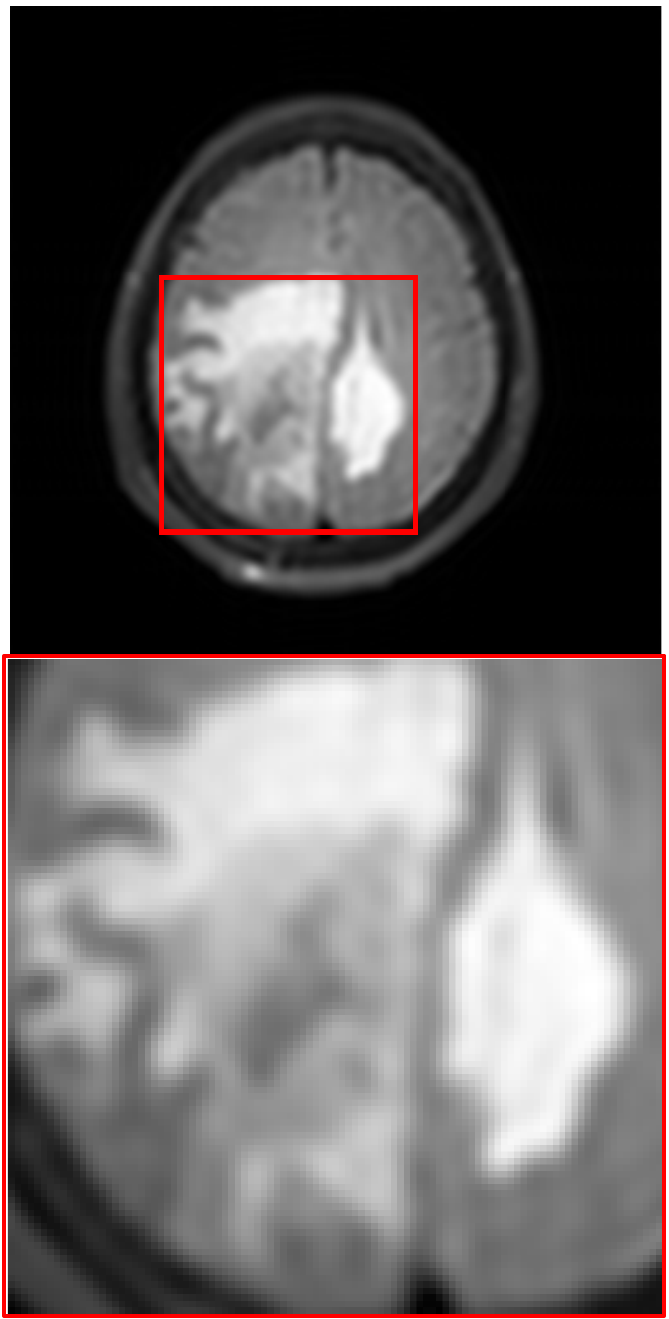}
	\end{minipage}
}
\hfill
	\subfloat[\footnotesize MINet \cite{Feng2021Multi}]{
	\begin{minipage}[b]{0.1\textwidth}
		\includegraphics[scale=0.17]{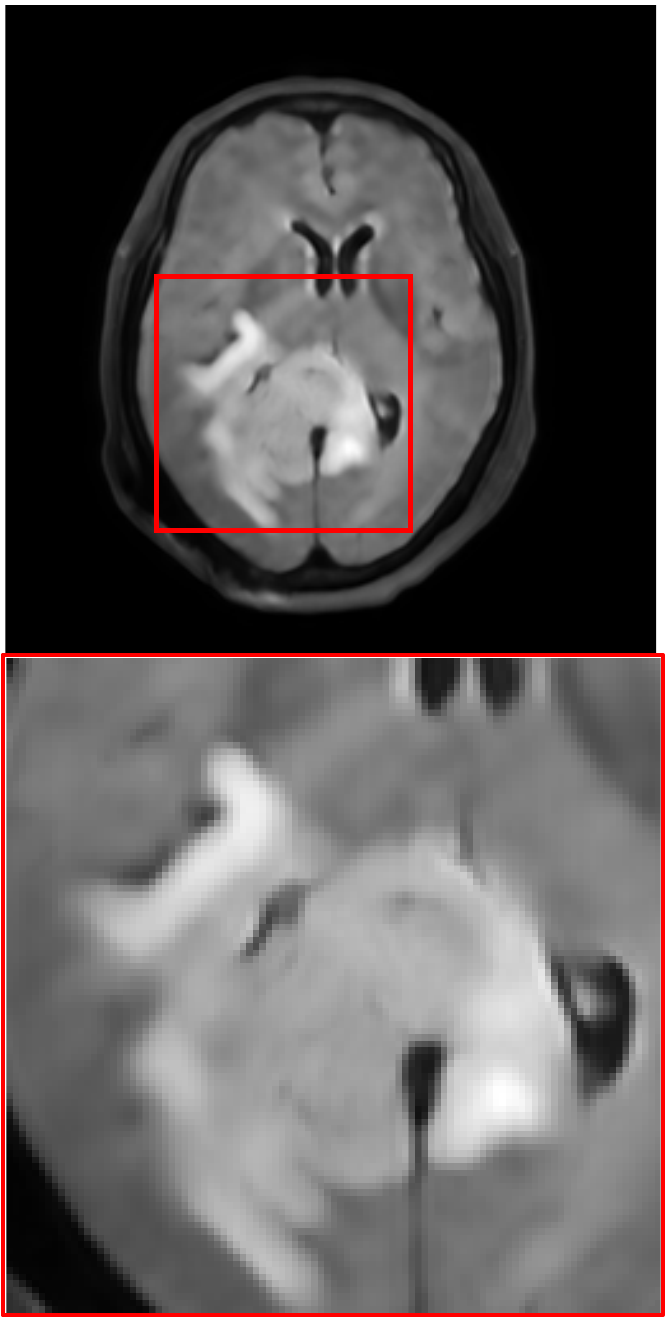} \\
        \includegraphics[scale=0.17]{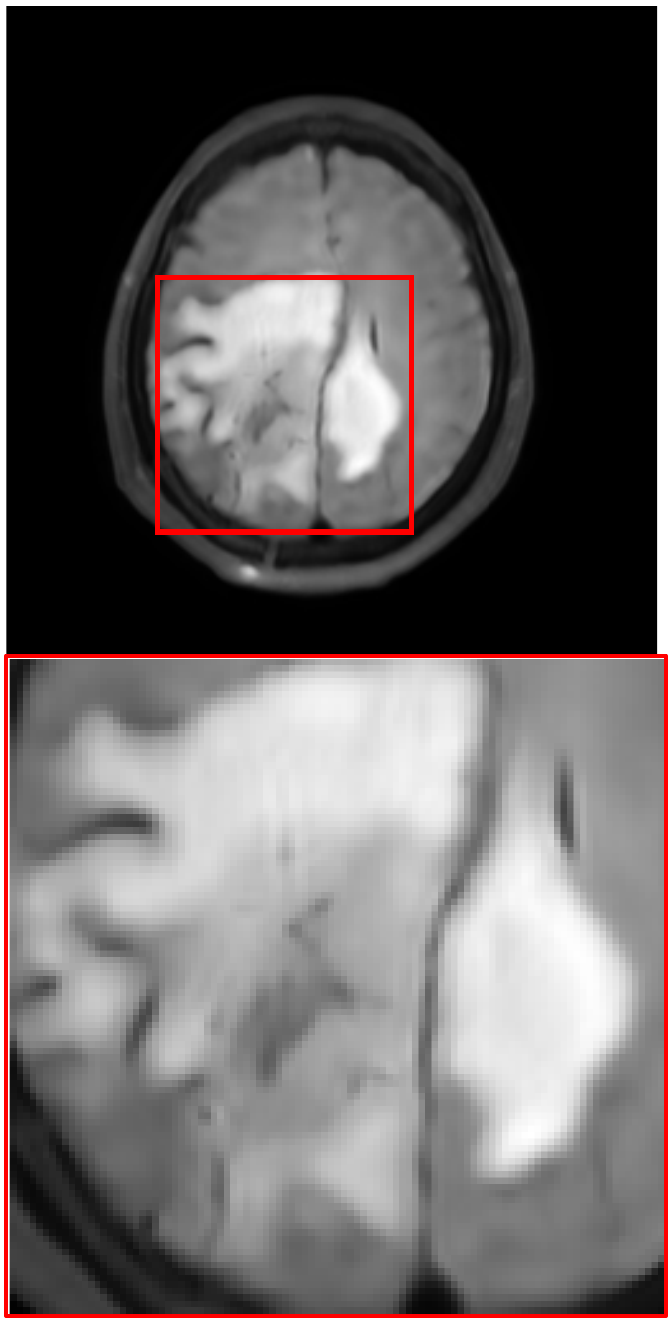}
	\end{minipage}
}
\hfill
	\subfloat[\footnotesize MASA \cite{lu2021masa}]{
	\begin{minipage}[b]{0.1\textwidth}
		\includegraphics[scale=0.17]{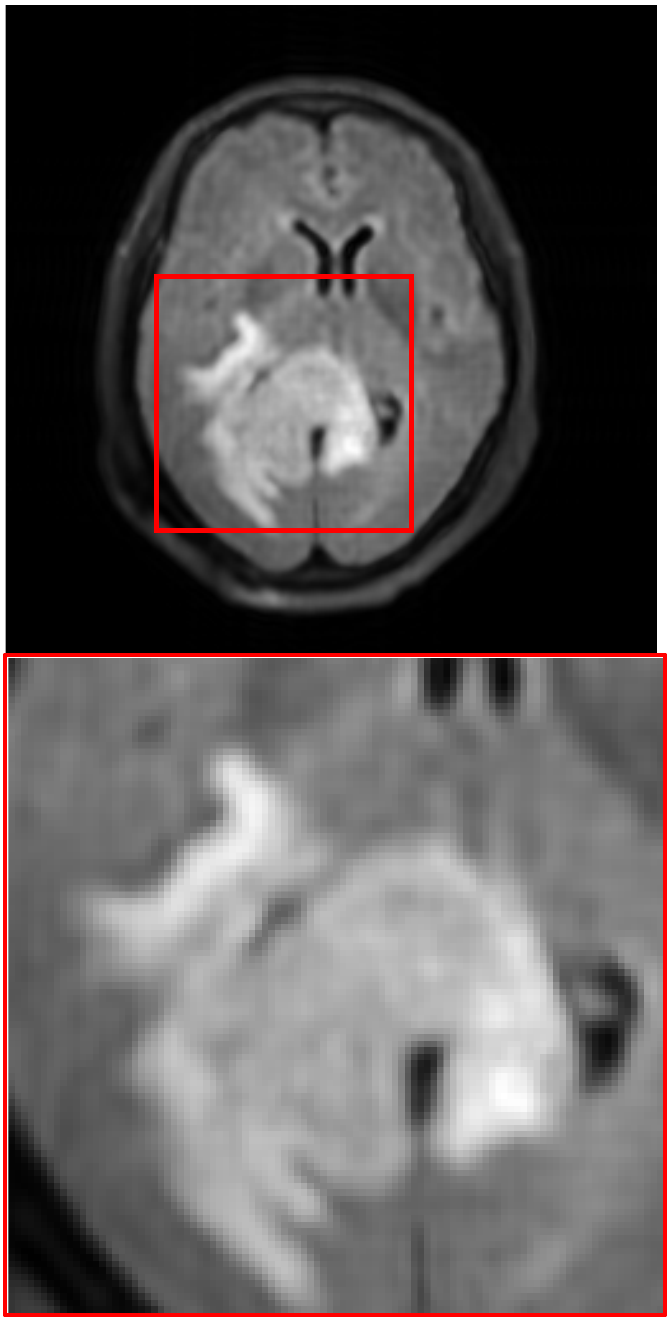} \\
        \includegraphics[scale=0.17]{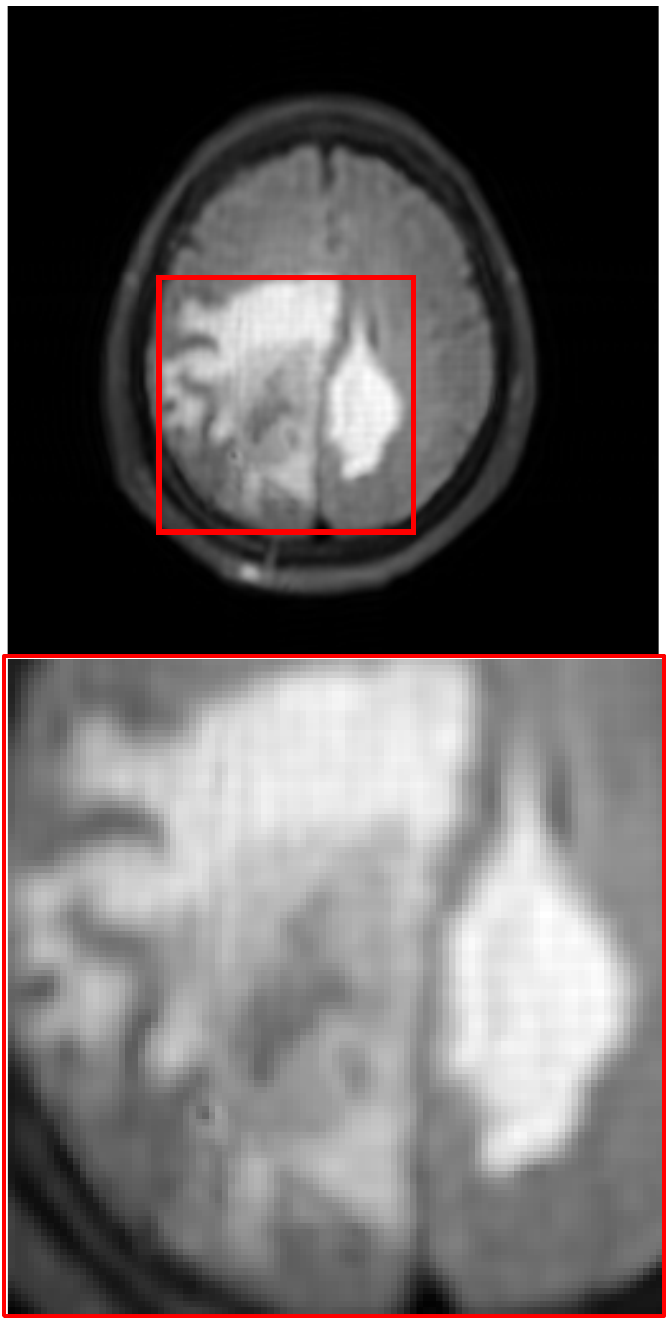}
	\end{minipage}
}
\hfill
    \subfloat[\footnotesize WavTrans \cite{Li2022Wav}]{
	\begin{minipage}[b]{0.1\textwidth}
		\includegraphics[scale=0.17]{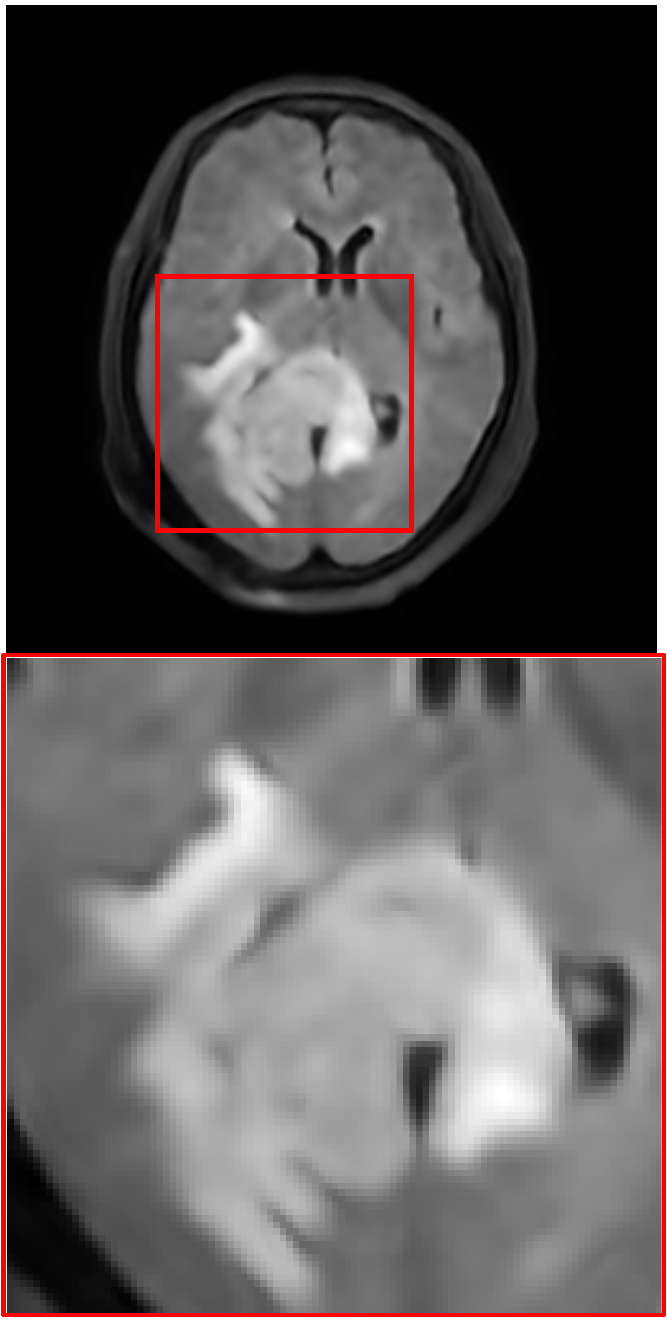} \\
        \includegraphics[scale=0.17]{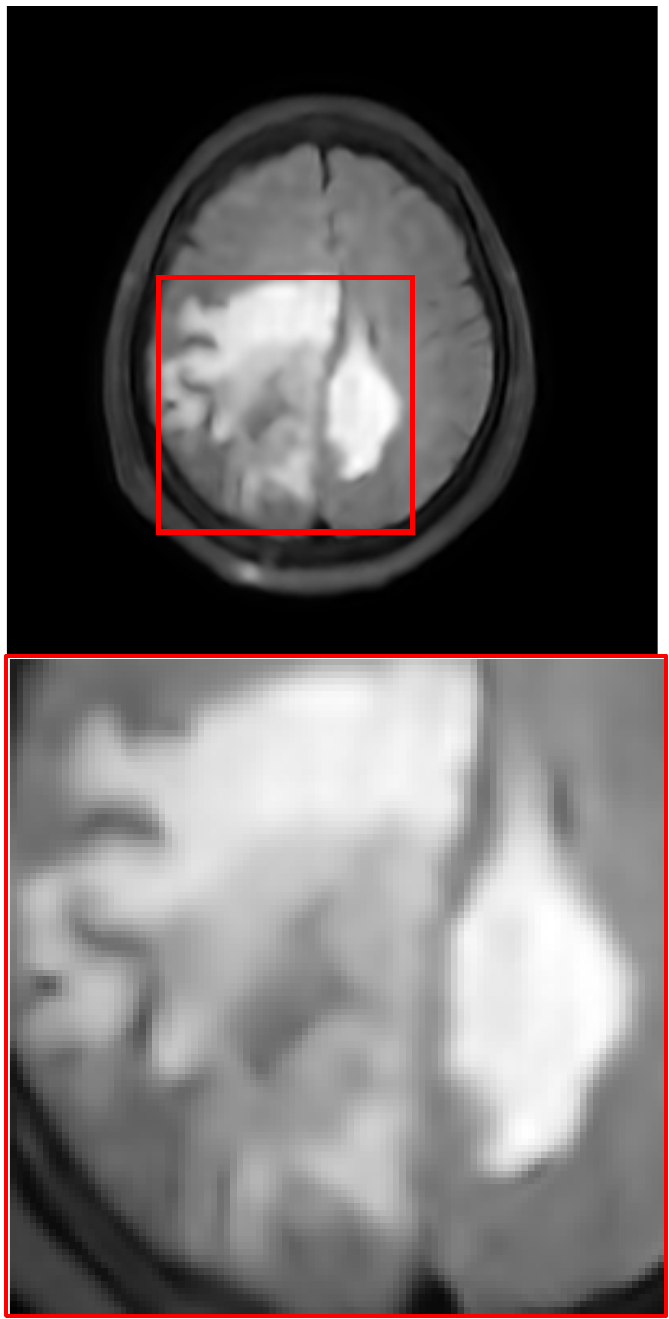}
	\end{minipage}
}
\hfill
    \subfloat[\footnotesize McMRSR \cite{Li2022Trans}]{
	\begin{minipage}[b]{0.1\textwidth}
		\includegraphics[scale=0.17]{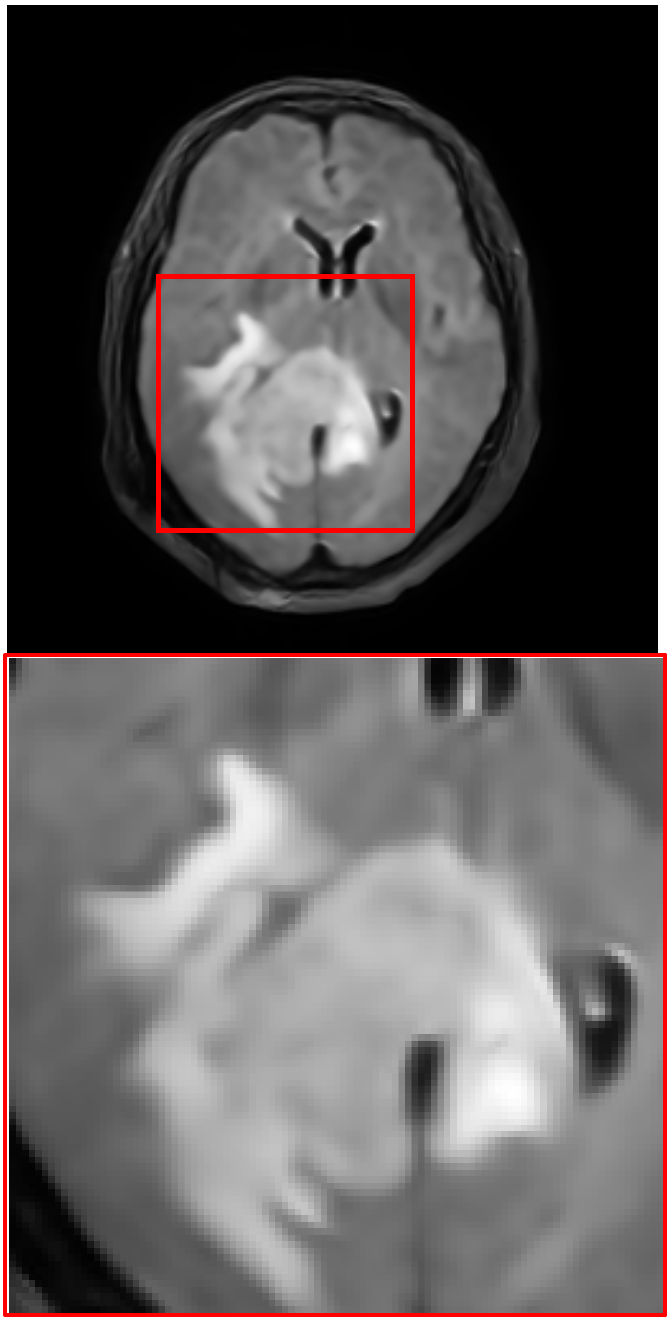} \\
        \includegraphics[scale=0.17]{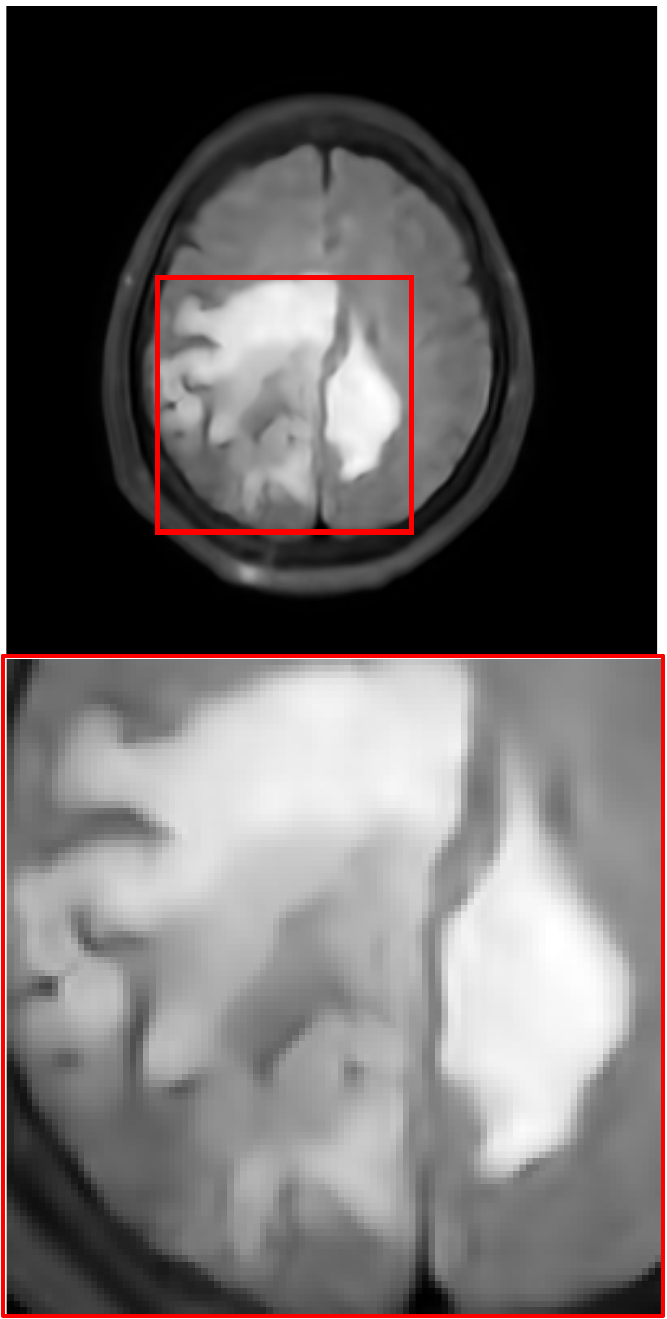}
	\end{minipage}
}
\hfill
    \subfloat[\footnotesize MC-VarNet \cite{lei2023decomposition}]{
	\begin{minipage}[b]{0.1\textwidth}
		\includegraphics[scale=0.17]{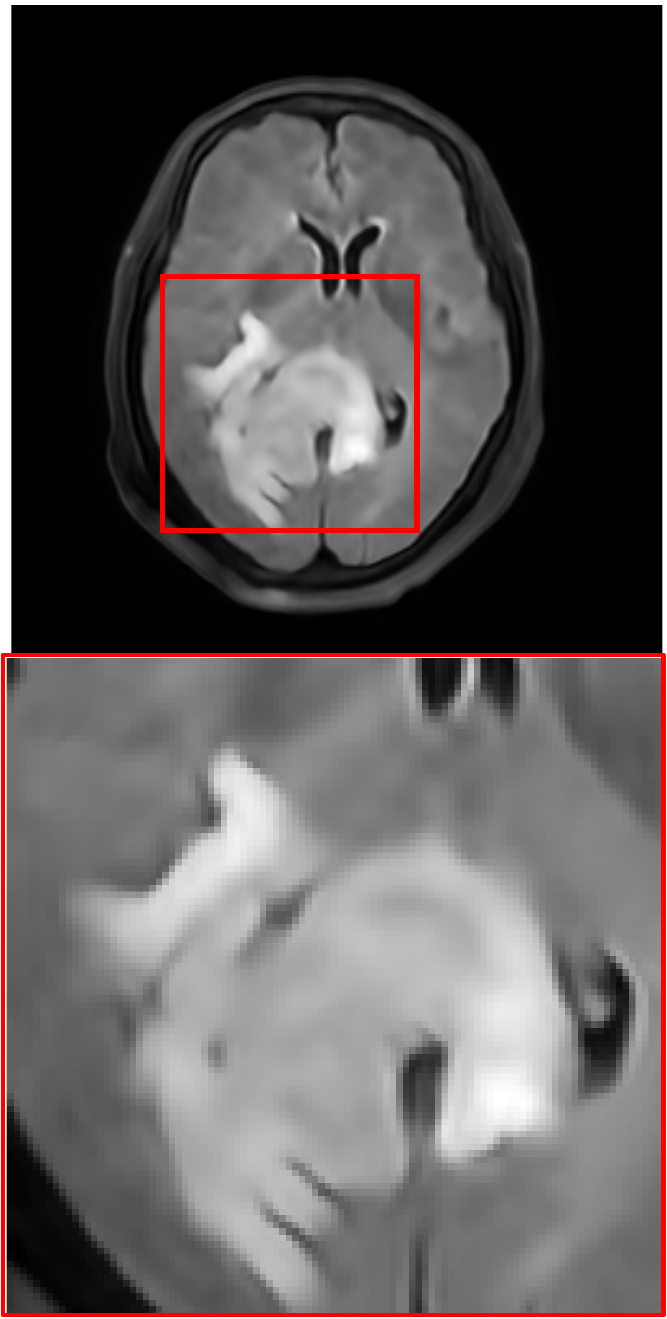}  \\
        \includegraphics[scale=0.17]{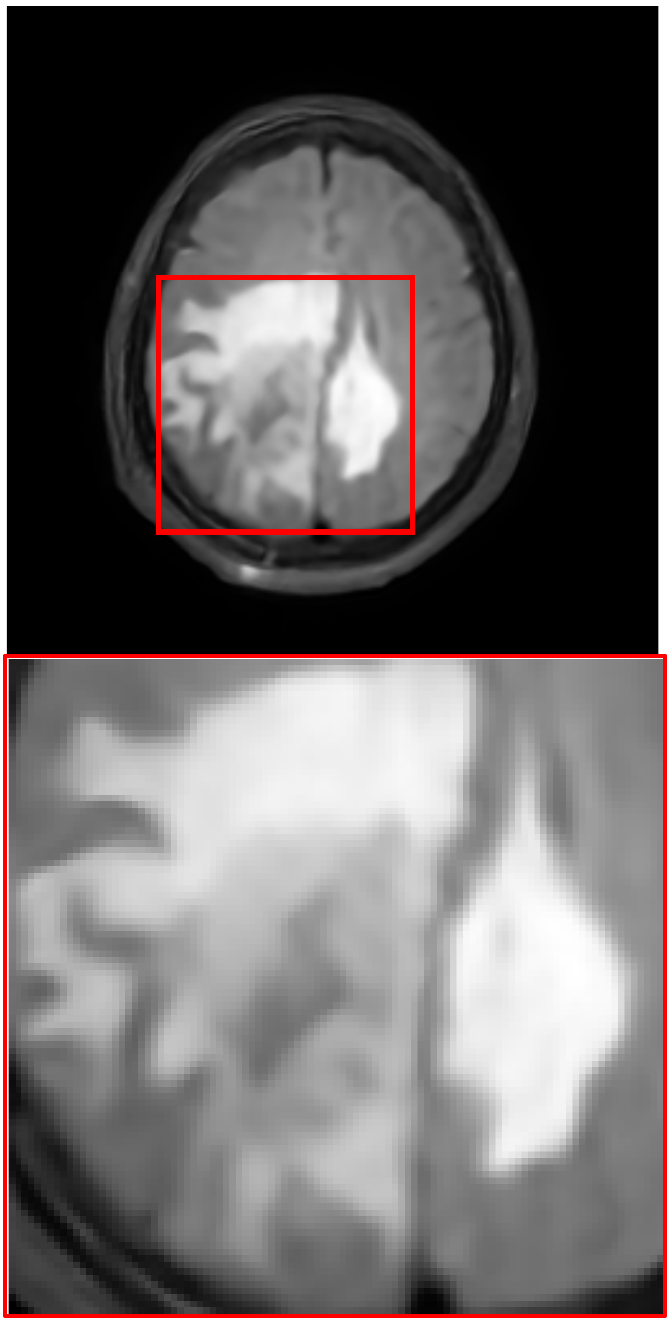}  
	\end{minipage}
}
\hfill
    \subfloat[\footnotesize DisC-Diff \cite{mao2023disc}]{
	\begin{minipage}[b]{0.1\textwidth}
		\includegraphics[scale=0.17]{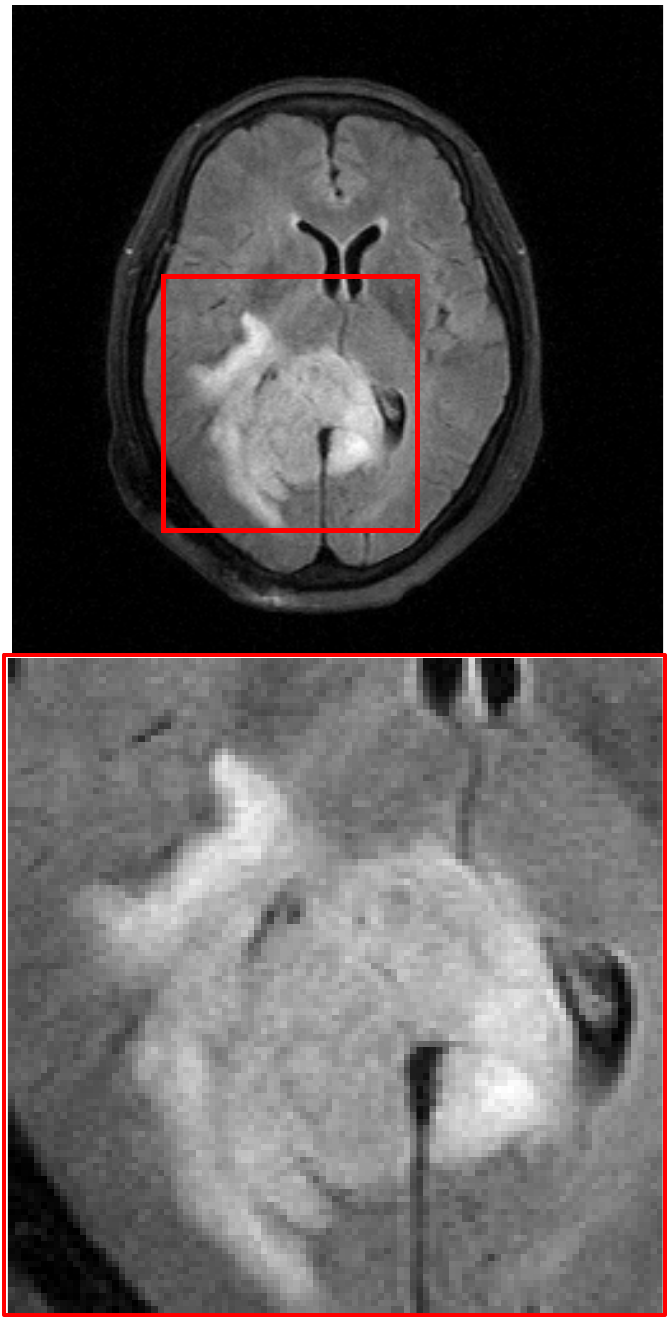} \\
        \includegraphics[scale=0.17]{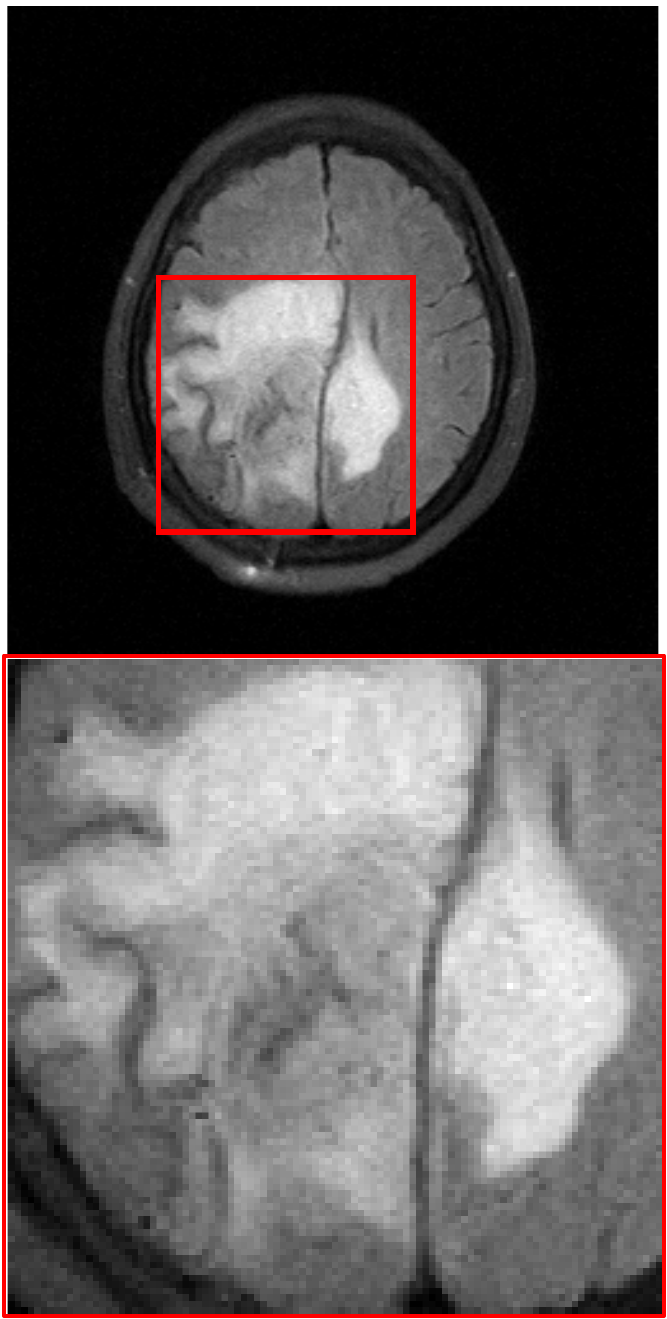}
	\end{minipage}
}
\hfill
    \subfloat[\footnotesize Ours]{
	\begin{minipage}[b]{0.1\textwidth}
		\includegraphics[scale=0.17]{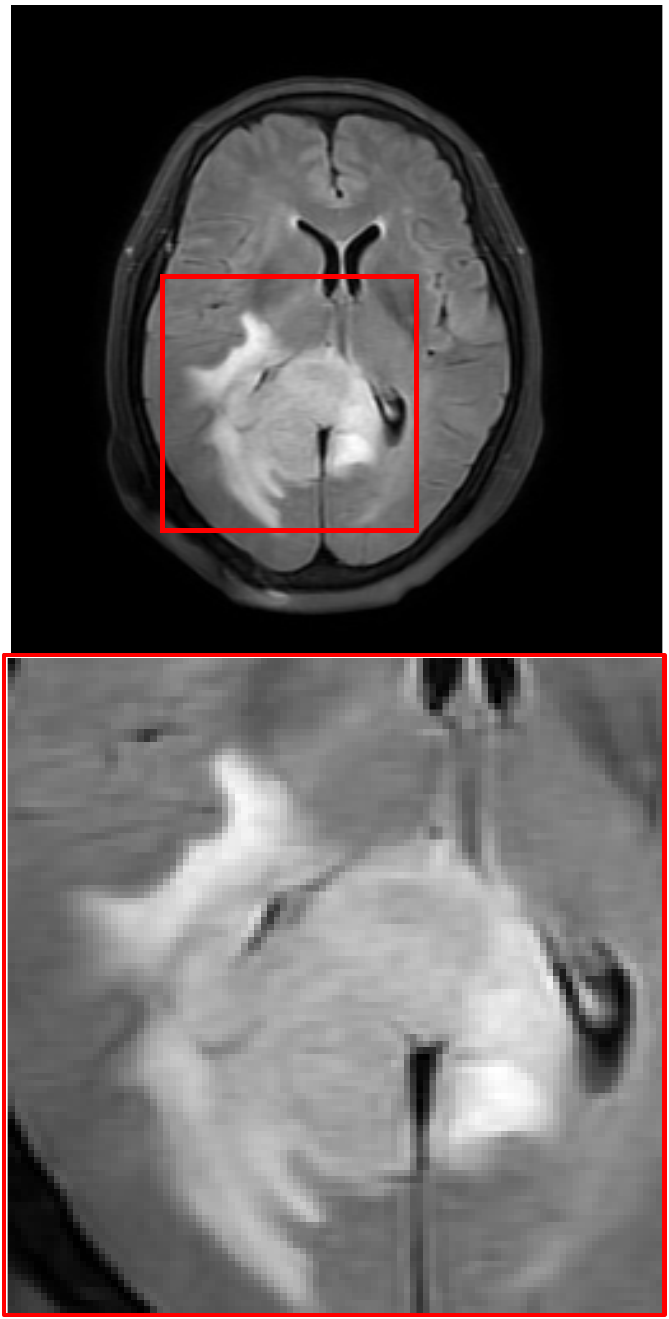} \\
        \includegraphics[scale=0.17]{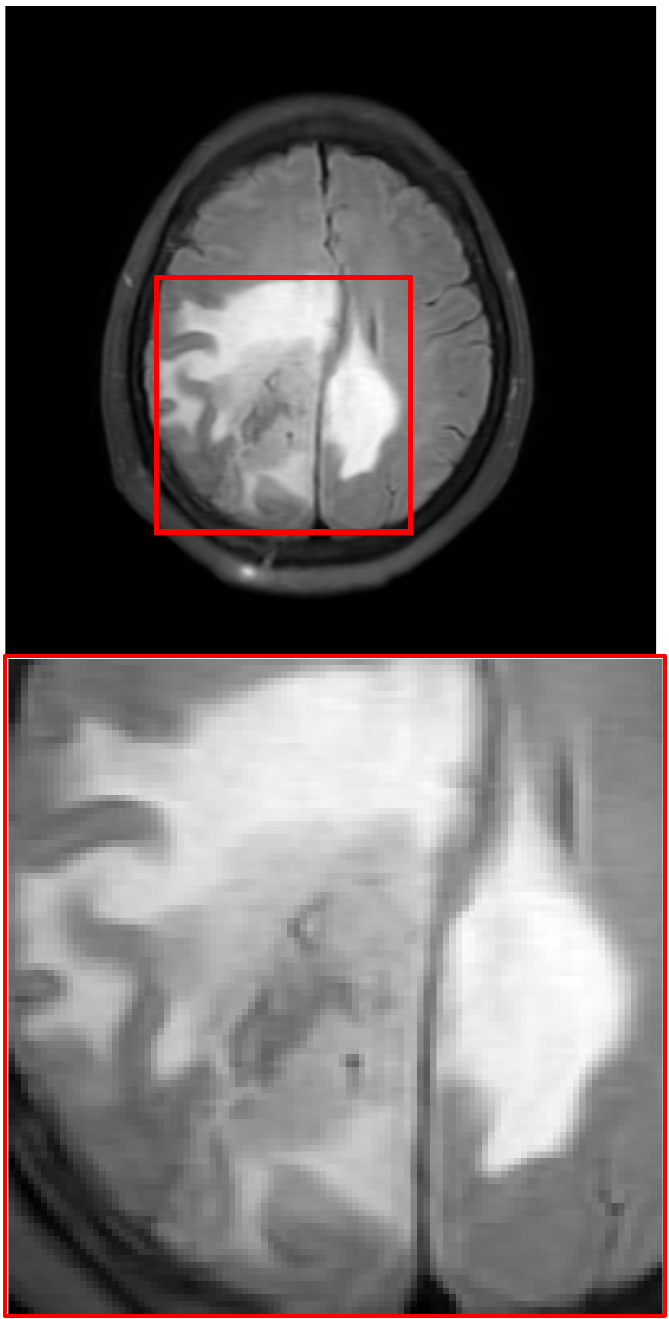}
	\end{minipage}
}
\hfill
    \subfloat[\footnotesize Target HR]{
	\begin{minipage}[b]{0.1\textwidth}
		\includegraphics[scale=0.17]{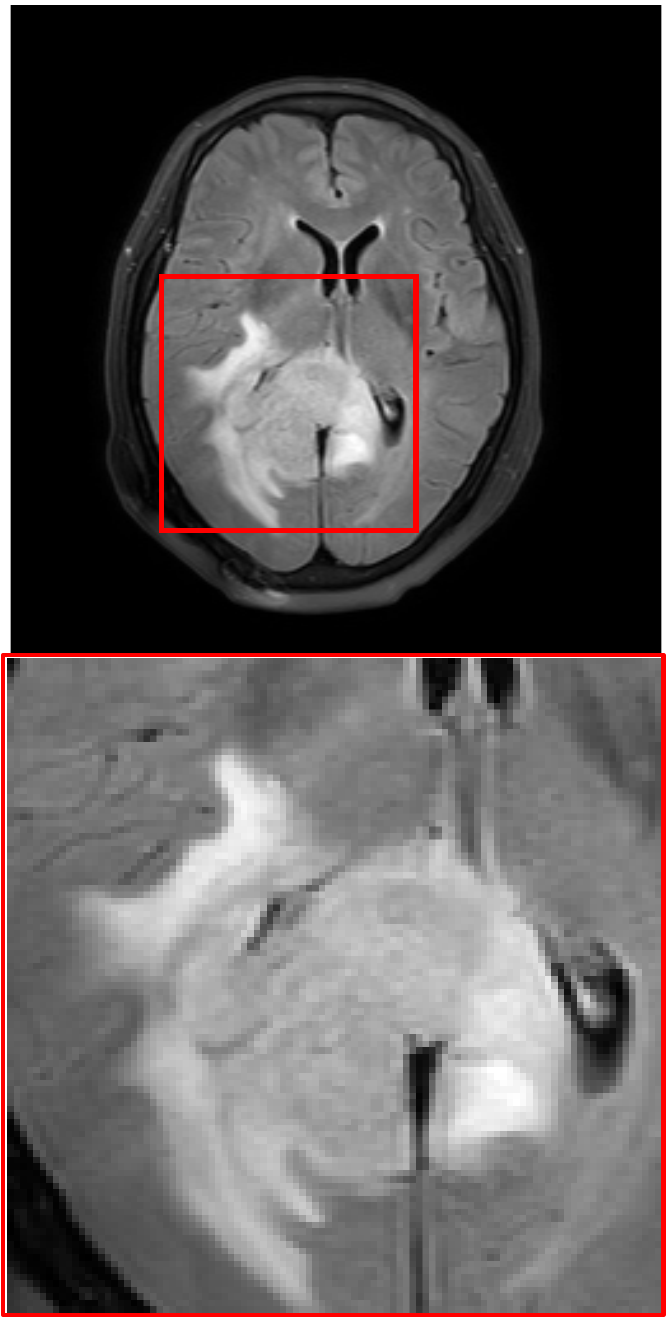} \\
        \includegraphics[scale=0.17]{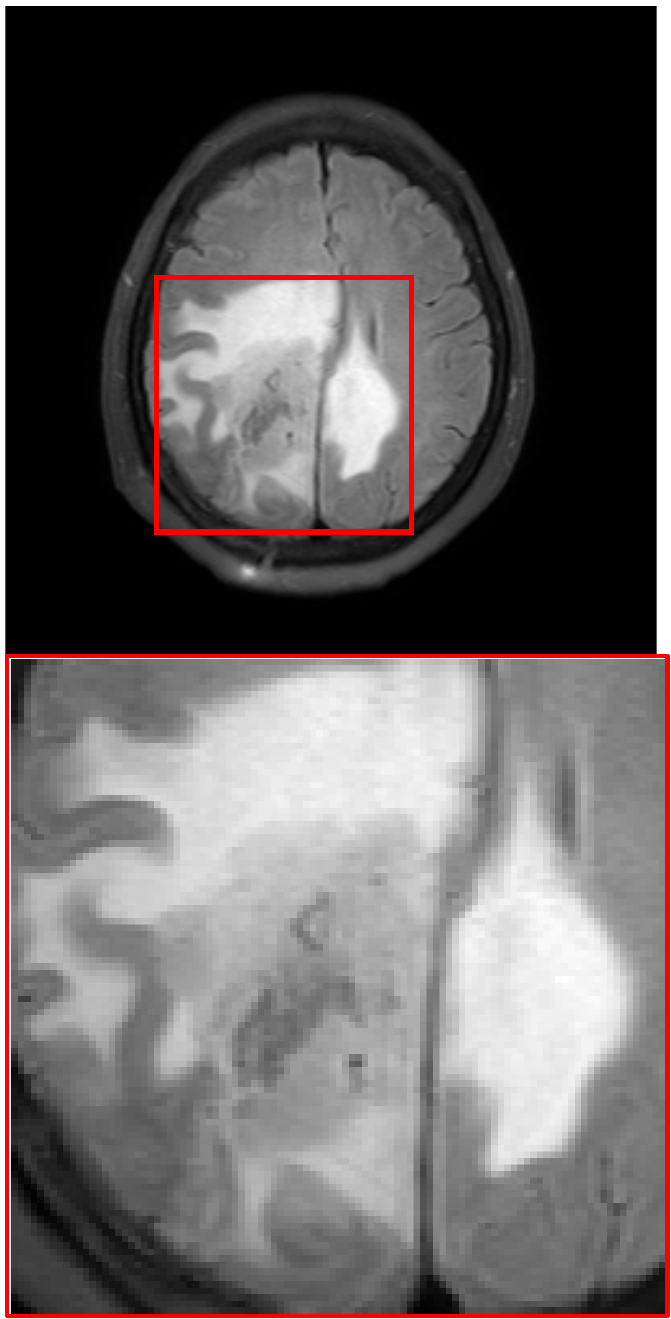}
	\end{minipage}
}
\caption{Qualitative visual comparison of various methods on the clinical tumor dataset ($4\times$).}
\label{fig_tumor}
\end{figure*}

\begin{figure*} [h]
	\centering
	\captionsetup[subfloat]{labelformat=empty}
	\subfloat[\footnotesize MCSR \cite{Lyu2019Multi}]{
	\begin{minipage}[b]{0.1\textwidth}
    	\includegraphics[scale=0.17]{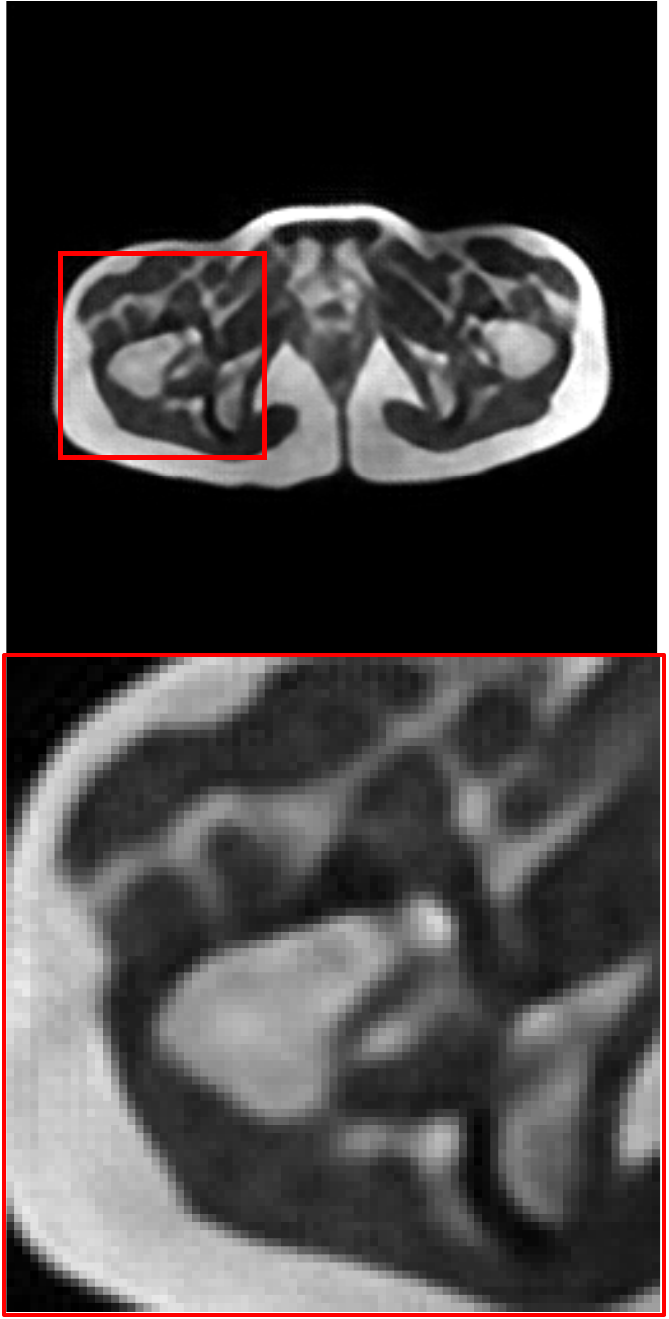} \\
        \includegraphics[scale=0.17]{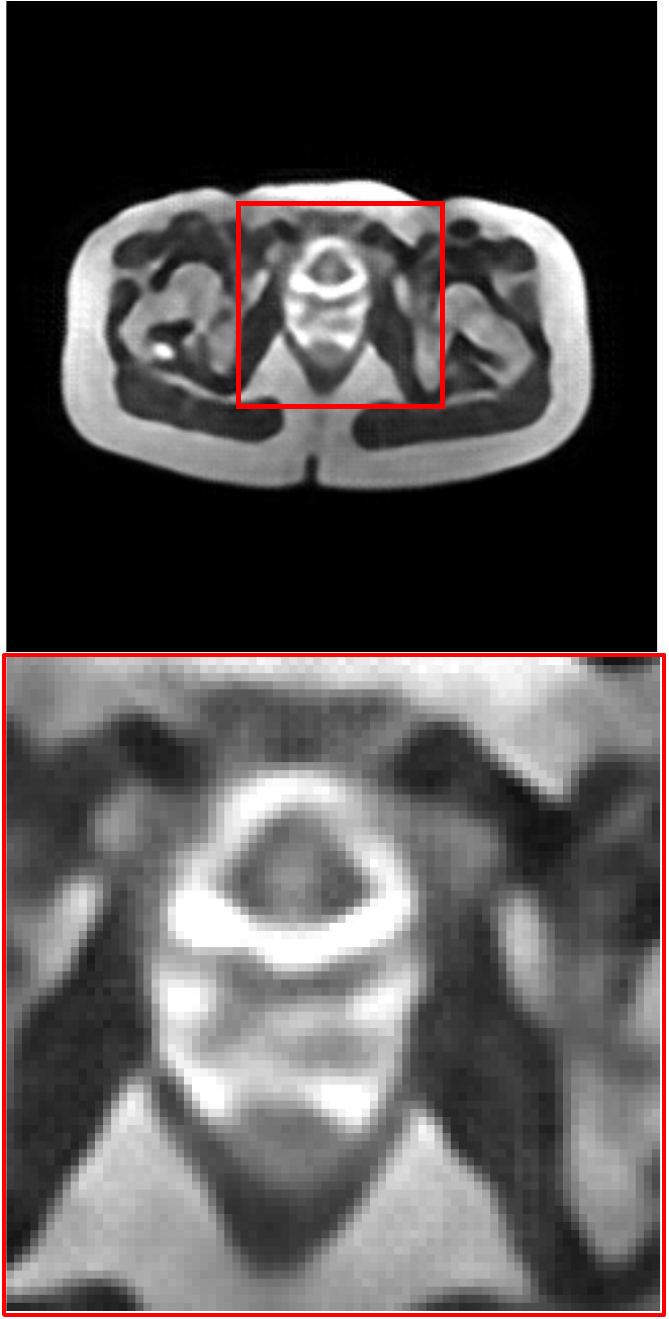}
	\end{minipage}
}
\hfill
	\subfloat[\footnotesize MINet \cite{Feng2021Multi}]{
	\begin{minipage}[b]{0.1\textwidth}
		\includegraphics[scale=0.17]{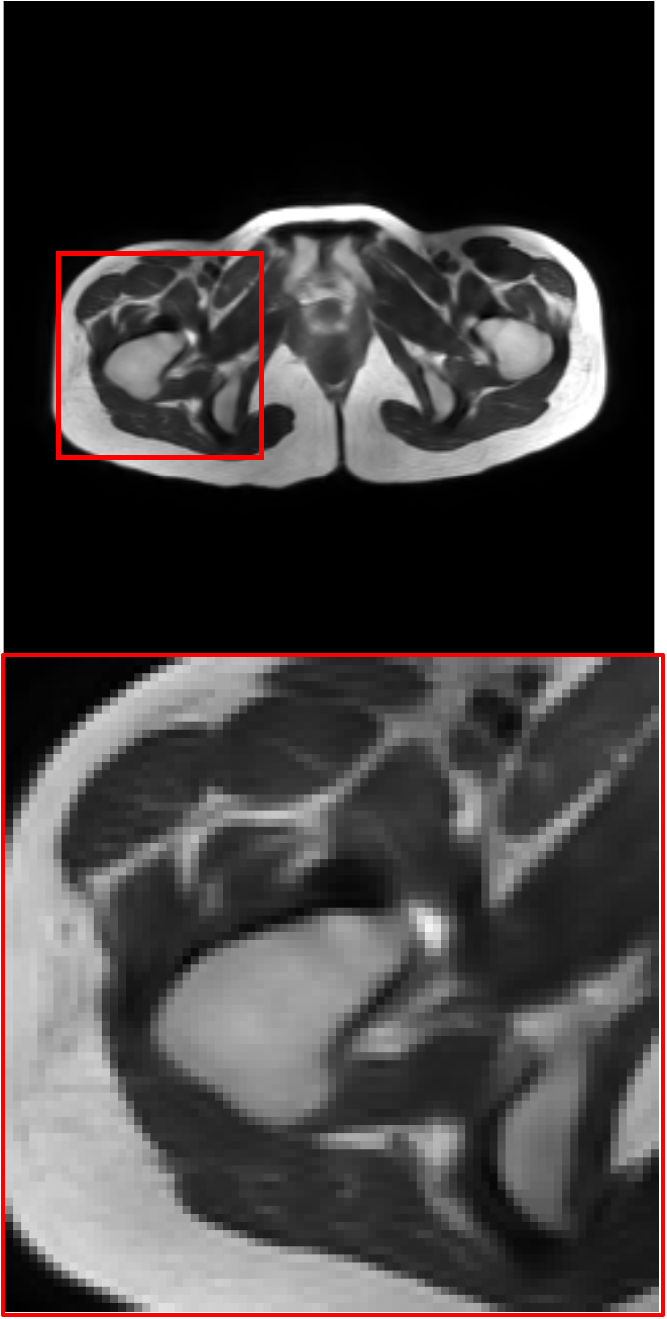} \\
        \includegraphics[scale=0.17]{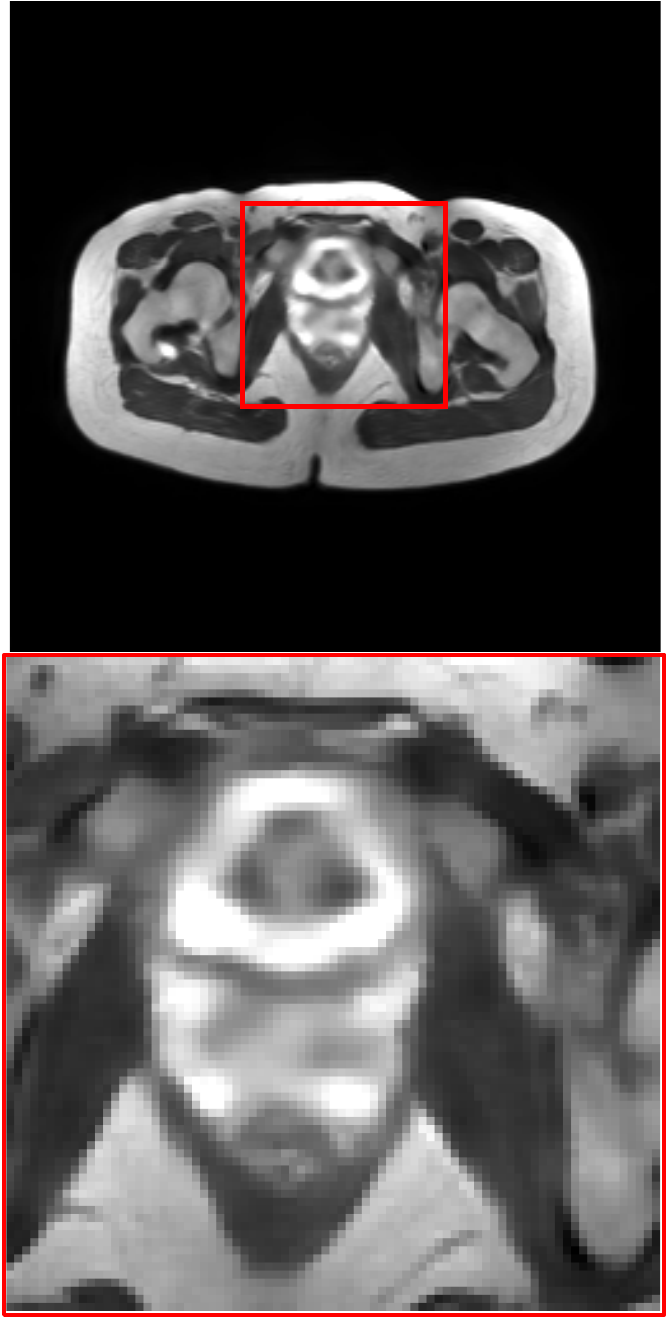}
	\end{minipage}
}
\hfill
	\subfloat[\footnotesize MASA \cite{lu2021masa}]{
	\begin{minipage}[b]{0.1\textwidth}
		\includegraphics[scale=0.17]{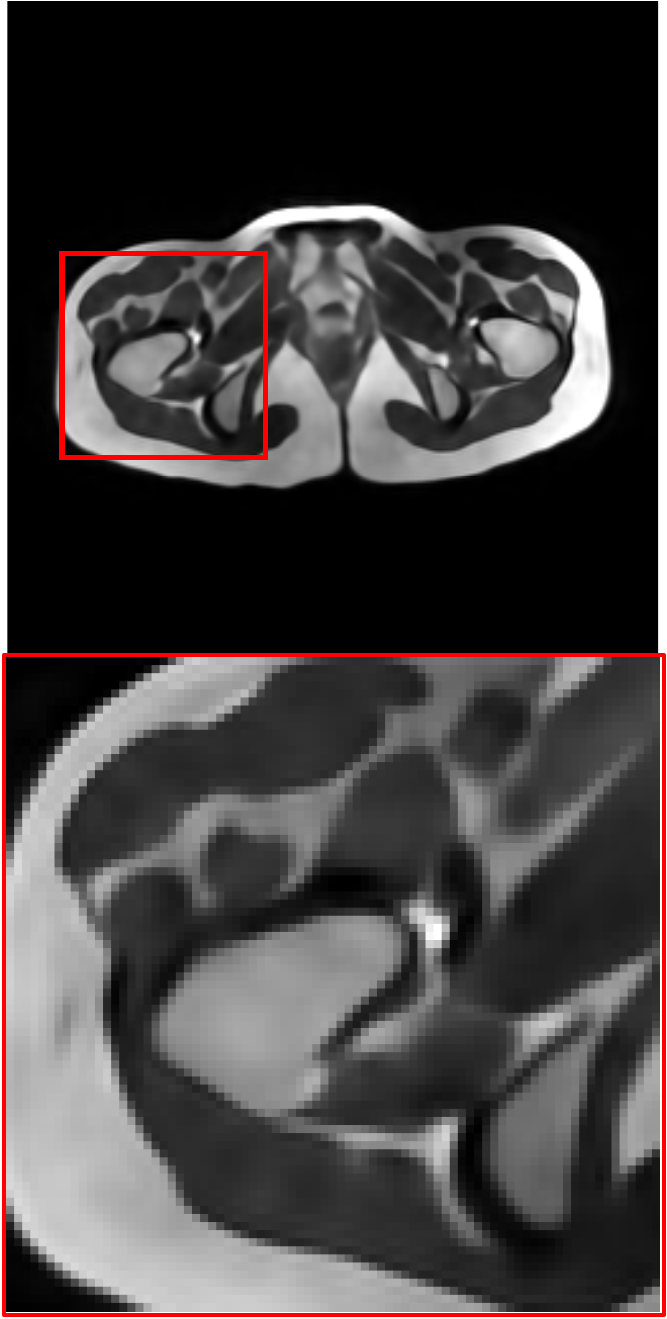} \\
        \includegraphics[scale=0.17]{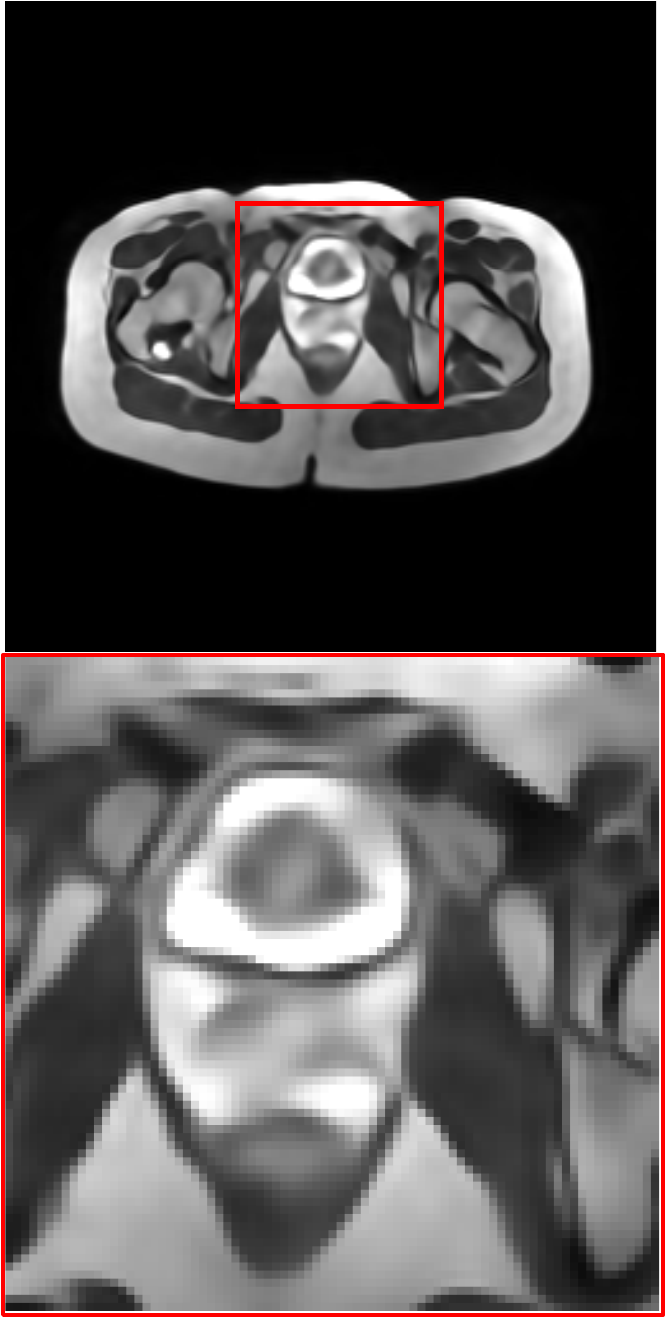}
	\end{minipage}
}
\hfill
    \subfloat[\footnotesize WavTrans \cite{Li2022Wav}]{
	\begin{minipage}[b]{0.1\textwidth}
		\includegraphics[scale=0.17]{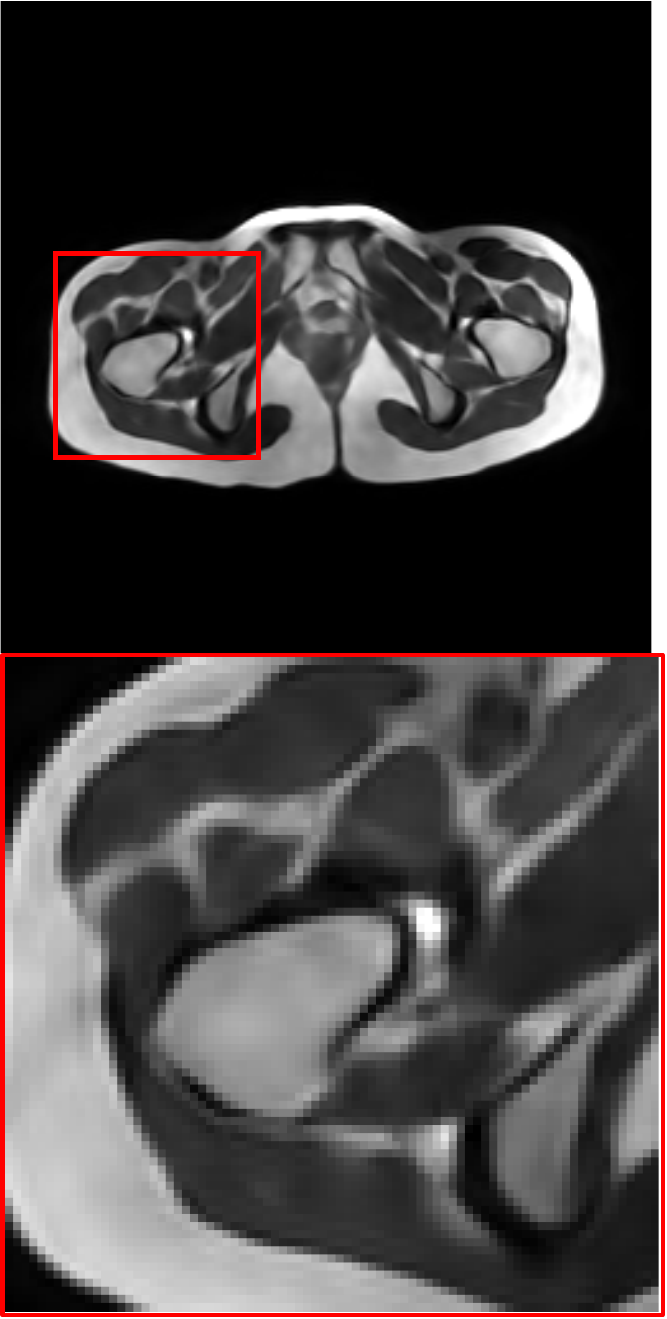} \\
        \includegraphics[scale=0.17]{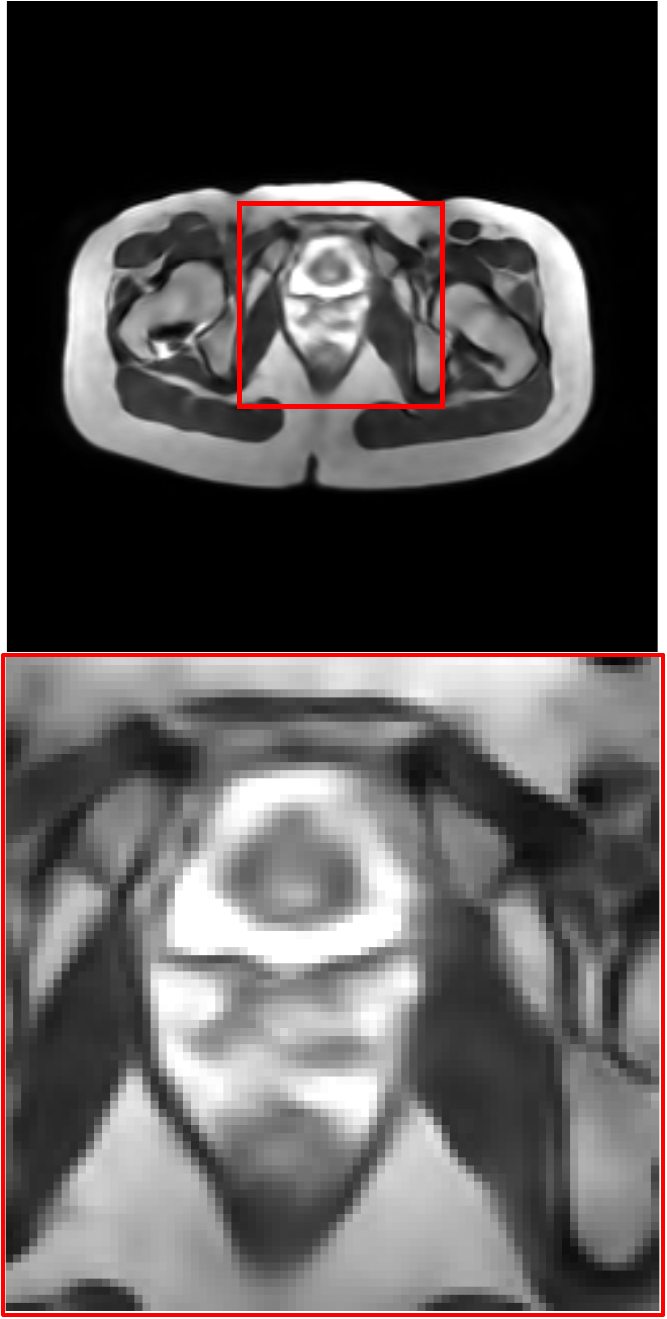}
	\end{minipage}
}
\hfill
    \subfloat[\footnotesize McMRSR \cite{Li2022Trans}]{
	\begin{minipage}[b]{0.1\textwidth}
		\includegraphics[scale=0.17]{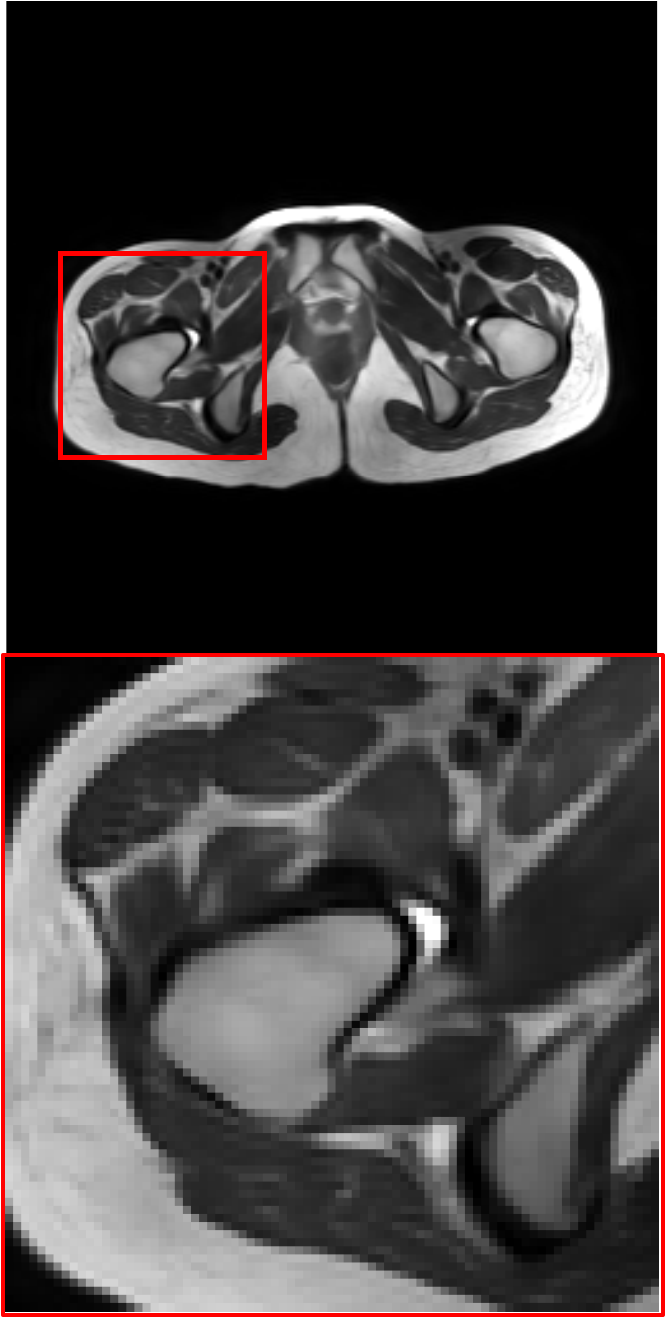} \\
        \includegraphics[scale=0.17]{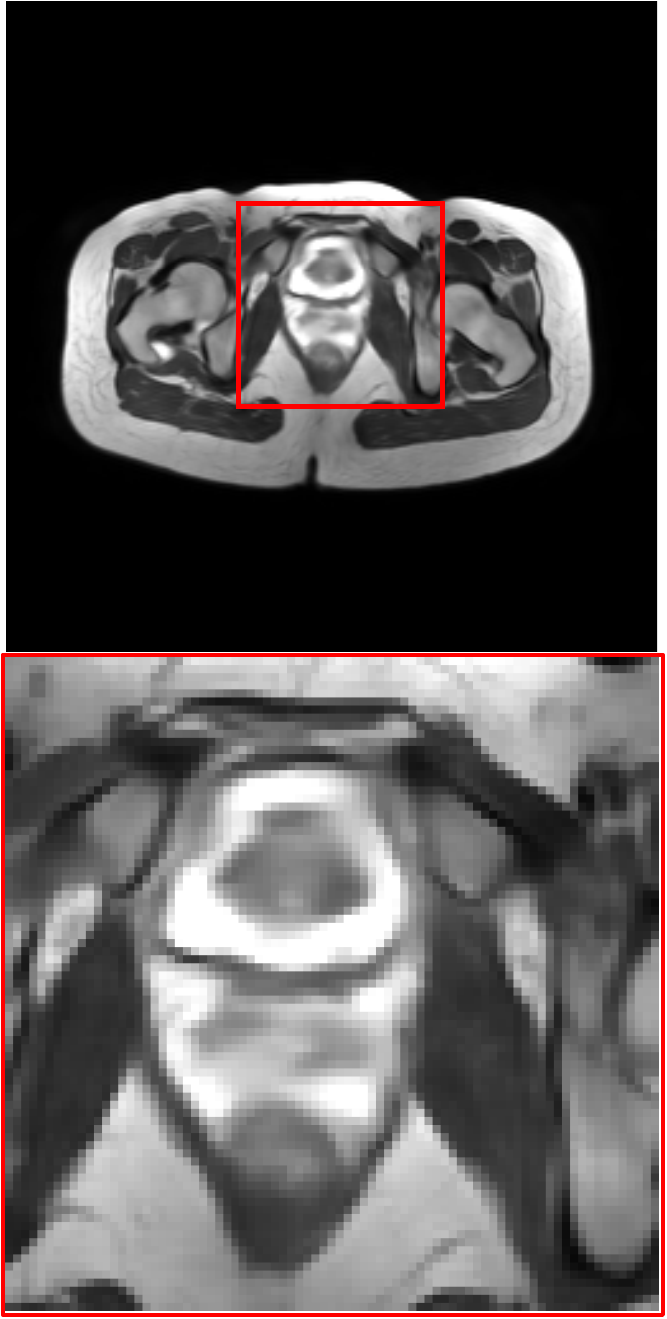}
	\end{minipage}
}
\hfill
    \subfloat[\footnotesize MC-VarNet \cite{lei2023decomposition}]{
	\begin{minipage}[b]{0.1\textwidth}
		\includegraphics[scale=0.17]{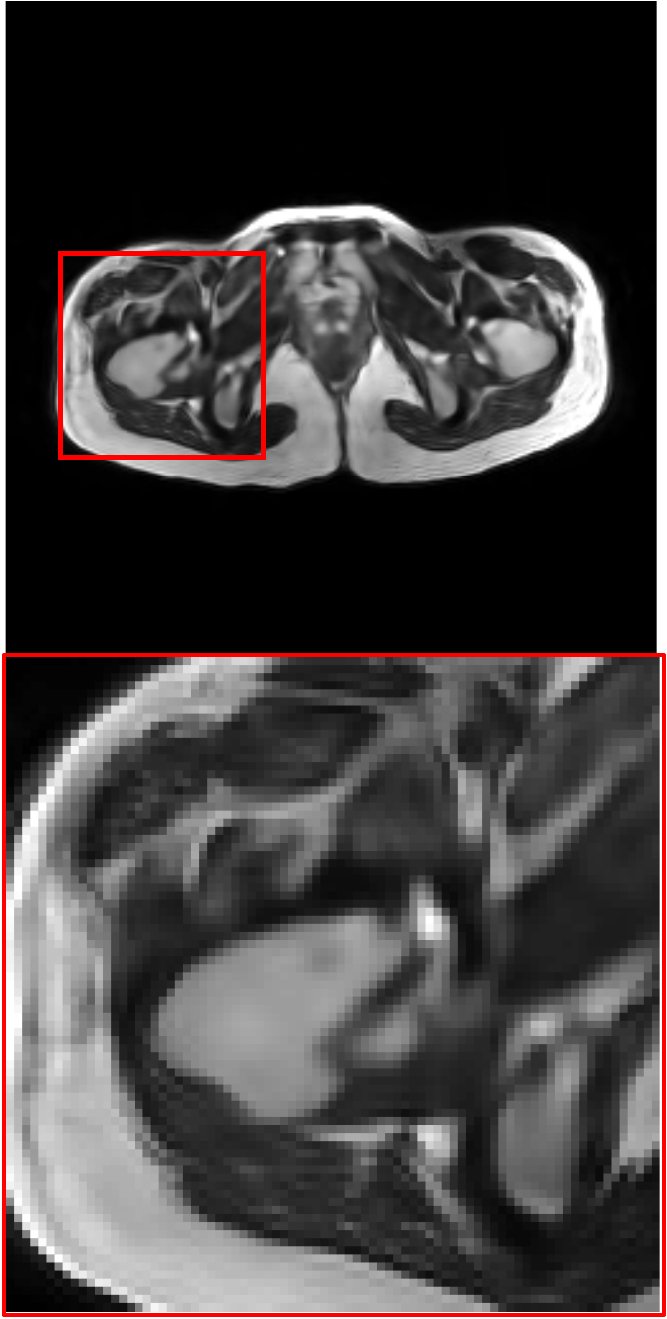}  \\
        \includegraphics[scale=0.17]{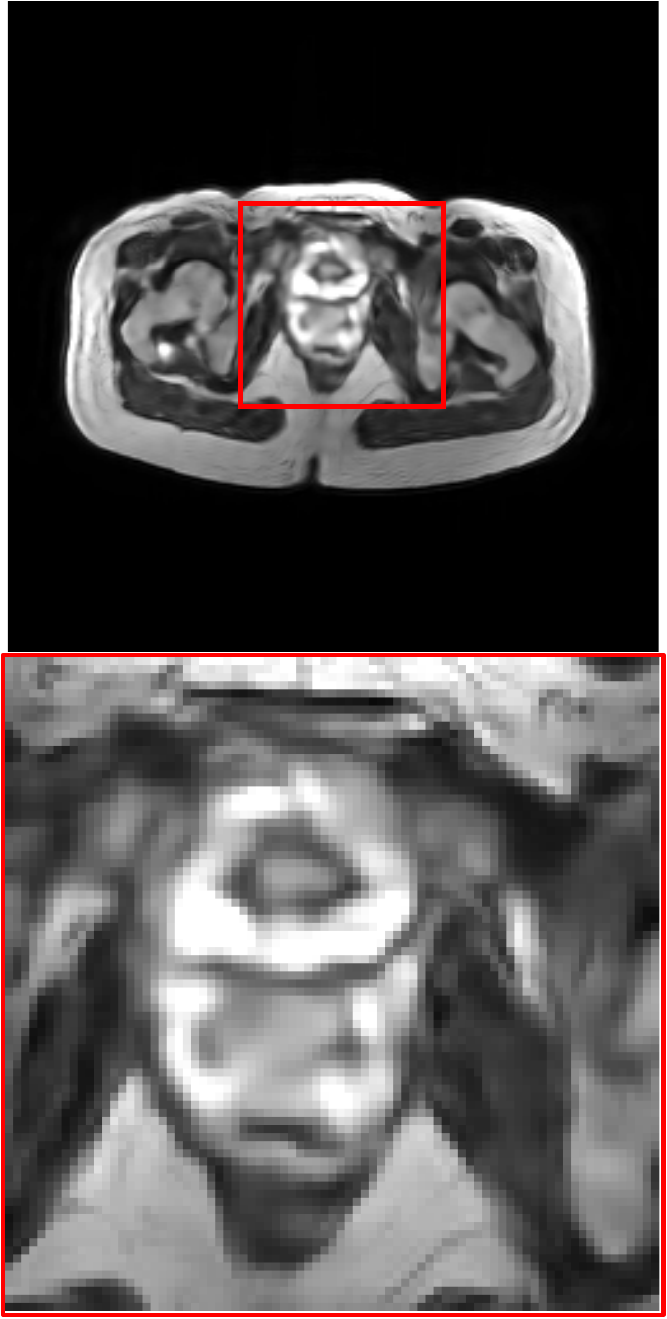}  
	\end{minipage}
}
\hfill
    \subfloat[\footnotesize DisC-Diff \cite{mao2023disc}]{
	\begin{minipage}[b]{0.1\textwidth}
		\includegraphics[scale=0.17]{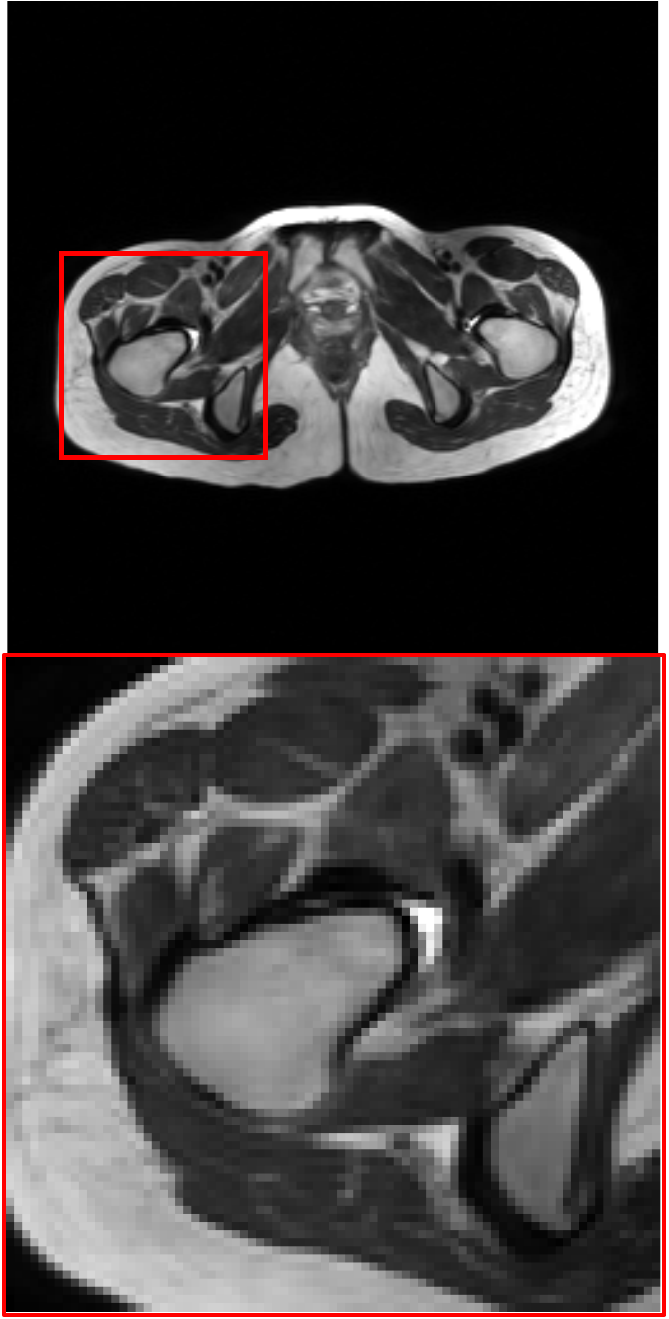} \\
        \includegraphics[scale=0.17]{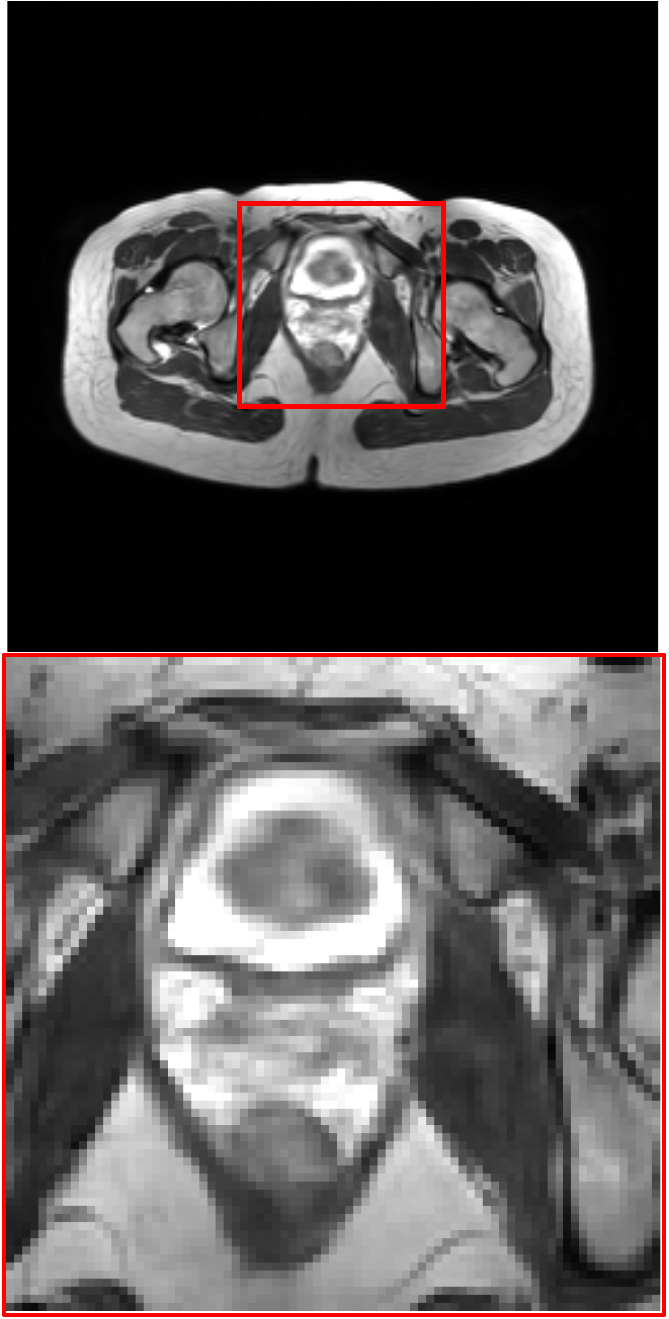}
	\end{minipage}
}
\hfill
    \subfloat[\footnotesize Ours]{
	\begin{minipage}[b]{0.1\textwidth}
		\includegraphics[scale=0.17]{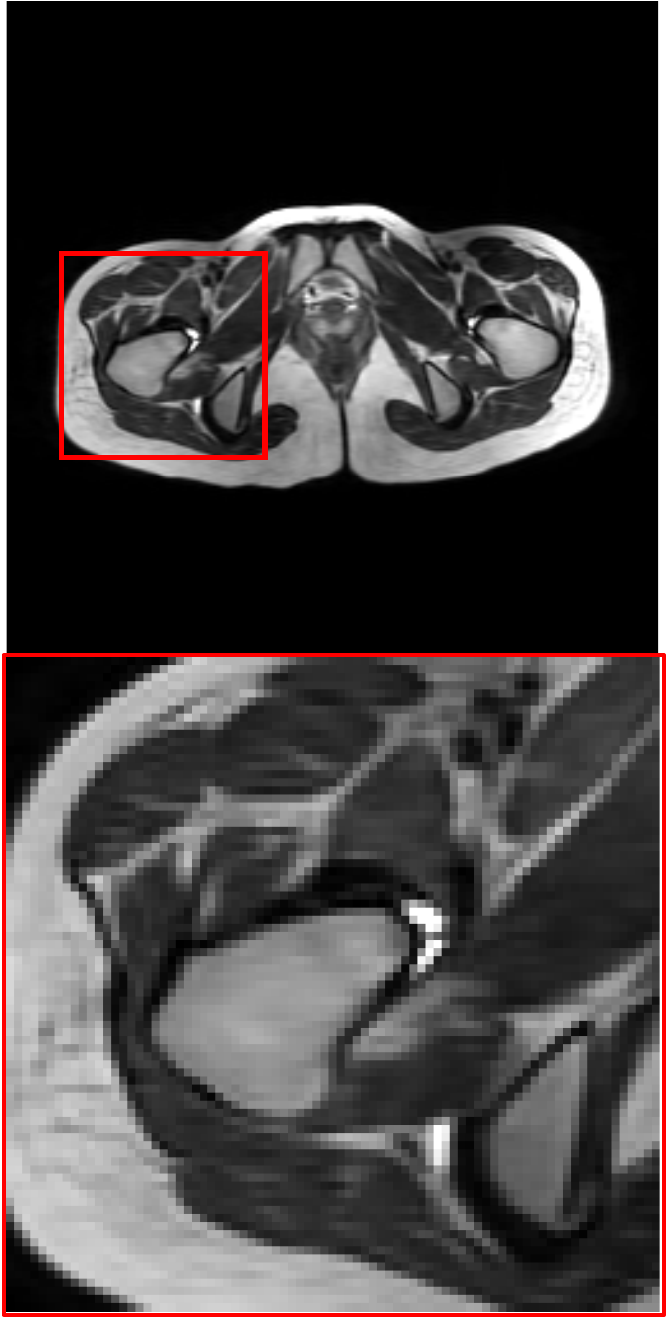} \\
        \includegraphics[scale=0.17]{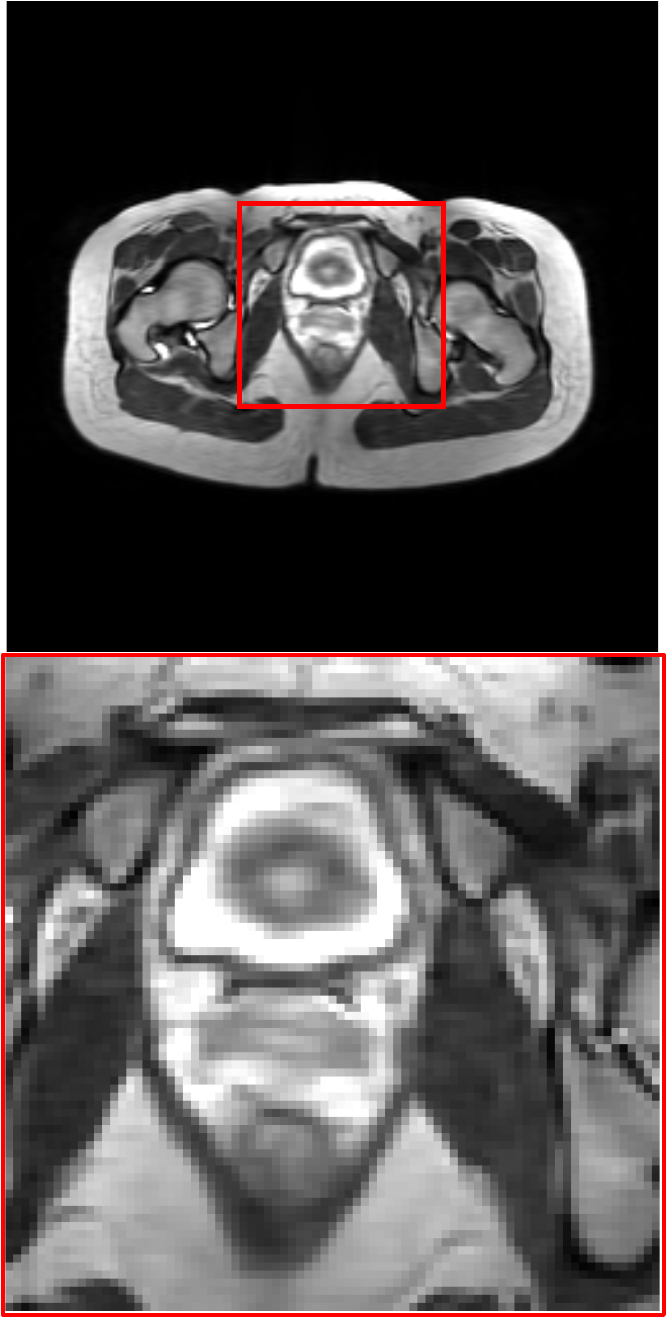}
	\end{minipage}
}
\hfill
    \subfloat[\footnotesize Target HR]{
	\begin{minipage}[b]{0.1\textwidth}
		\includegraphics[scale=0.17]{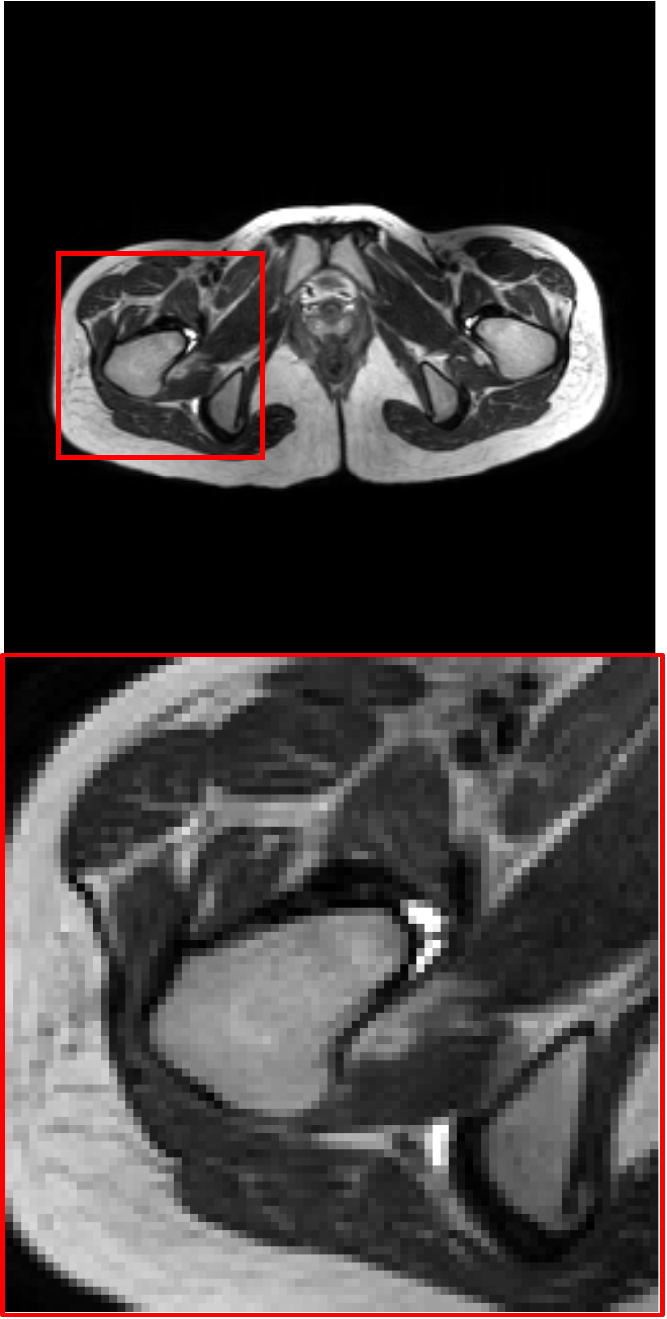} \\
        \includegraphics[scale=0.17]{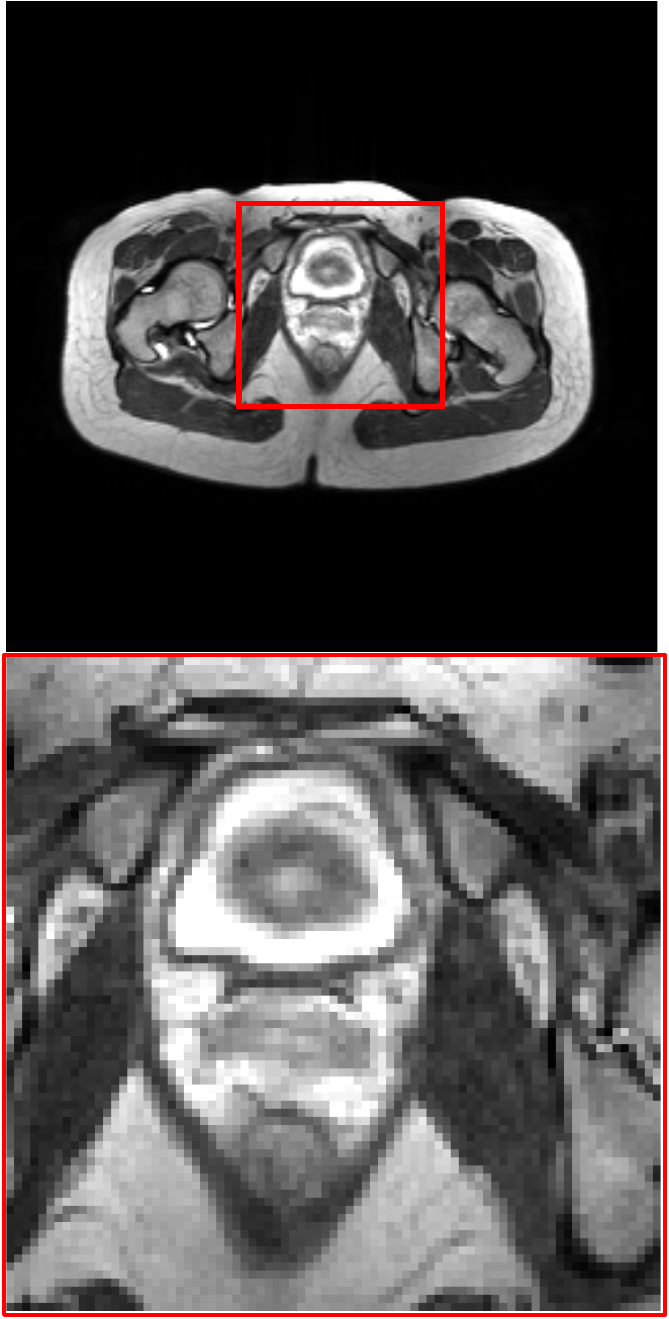}
	\end{minipage}
}
\caption{Qualitative visual comparison of various methods on the clinical pelvic dataset ($4\times$).}
\label{fig_pelvic}
\end{figure*}

\begin{table}[t!]
\small
\centering
\begin{tabular}{l|c|c|c|c}
\hline
\hline
Method & \cellcolor[gray]{0.9} Window Size & \cellcolor[gray]{0.9} PSNR & \cellcolor[gray]{0.9} SSIM & \cellcolor[gray]{0.9} FLOPs  \\  
\hline
\hline
 SwinIR \cite{liang2021swinir} & 8$\times$8  & 30.13  & 0.8107  &  32.99G   \\
\hline
 SwinIR \cite{liang2021swinir} & 16$\times$16  & 30.54  & 0.8183 & 39.79G    \\
\hline
 PLWformer & 16$\times$16 & 30.52 & 0.8180 & \textbf{29.45G}   \\
\hline
PLWformer & 32$\times$32 & \textbf{30.78} & \textbf{0.8242} & 36.25G  \\
\hline
\hline
\end{tabular}
\caption{Performance comparison between SwinIR \cite{liang2021swinir} and PLWformer at different window sizes. The best result is marked in \textbf{bold}.}
\label{tab:size}
\end{table}

\section{Window Size Analyses}
We further show the comparison of window size and computational complexity (\emph{e.g.}, FLOPs) in Table \ref{tab:size}.
For a fair comparison, we conduct single-contrast SR reconstruction employing SwinIR \cite{liang2021swinir} and PLWformer, with the optimization function utilizing the L1 loss. 
Note that in this case, the PLWformer does not utilize the prior knowledge represented by $Z$. The performance metrics, including PSNR, SSIM, and FLOPs, are evaluated with the image size set to 64$\times$64, the upsampling factor at 4, and using the FastMRI dataset.
For SwinIR, expanding the window size can improve the performance of the network, but it also increases computational complexity. 
In contrast, PLWformer employs permutation operations to transfer some spatial information to the channel dimension. Therefore, even with an expanded window size, the computational complexity does not increase significantly. When the window size is 16$\times$16, the FLOPs required for PLWformer are smaller than those for SwinIR.


Besides, we observe that when the window size is the same, the performance of PLWformer is slightly lower than SwinIR, due to the reduction of tokens $K$ and $V$ in PLWformer, which leads to the loss of a small portion of the structural information of the image. However, PLWformer effectively reduces the computational burden, so a slight performance reduction is acceptable.

\section{More Visual Comparisons}
In this section, we present more visual qualitative comparisons. Figures \ref{fig_knee}, \ref{fig_brain}, \ref{fig_tumor}, and \ref{fig_pelvic} show the reconstruction results of each method in FastMRI, clinical brain, clinical tumor, and clinical pelvic, respectively.
As can be seen, although DisC-Diff can reconstruct MR images with high-frequency information, it fails to preserve the structure and content of the original Target HR image effectively, resulting in image distortion. In contrast, our proposed DiffMSR can restore high-frequency information while preserving the structure of the original HR image, indicating the effectiveness of the joint use of DM and PLWformer.

\begin{figure*} [h]
	\centering
	\captionsetup[subfloat]{labelformat=empty}
	\subfloat[\footnotesize DiffuseRecon \cite{peng2022towards}]{
	\begin{minipage}[b]{0.155\textwidth}
    	\includegraphics[scale=0.24]{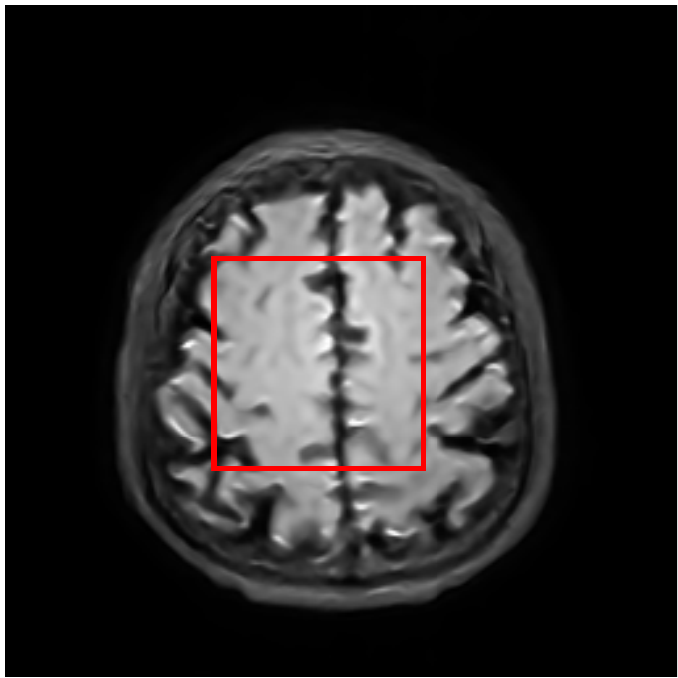} \\
        \includegraphics[scale=0.24]{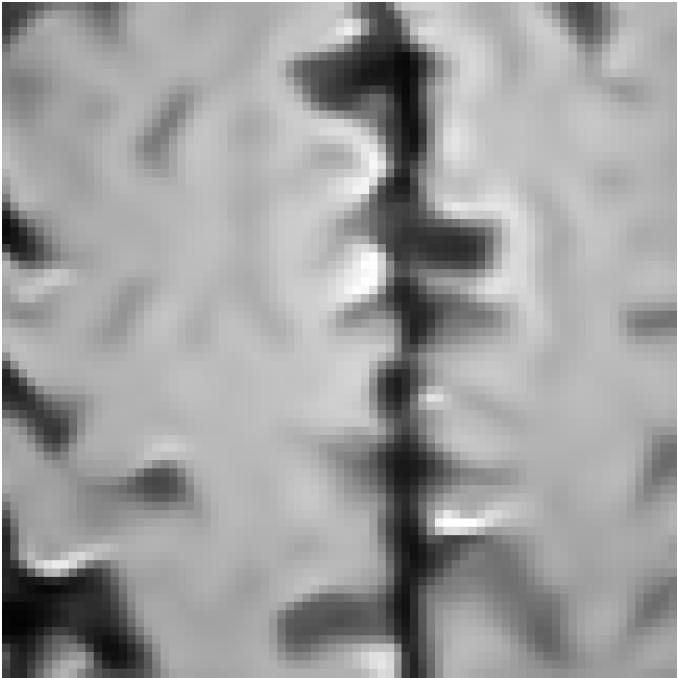}
	\end{minipage}
}
\hfill
	\subfloat[\footnotesize MC-DDPM \cite{xie2022measurement}]{
	\begin{minipage}[b]{0.155\textwidth}
		\includegraphics[scale=0.24]{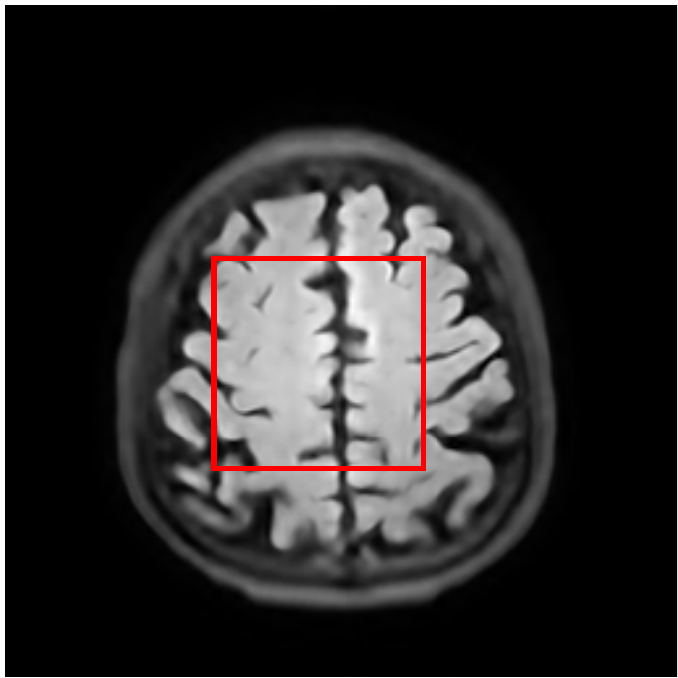} \\
        \includegraphics[scale=0.24]{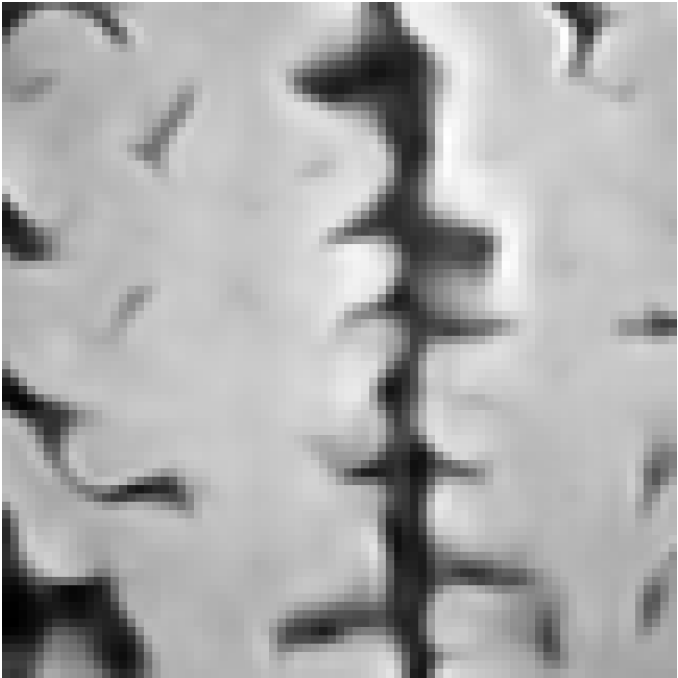}
	\end{minipage}
}
\hfill
	\subfloat[\footnotesize Score-MRI \cite{chung2022score}]{
	\begin{minipage}[b]{0.155\textwidth}
		\includegraphics[scale=0.24]{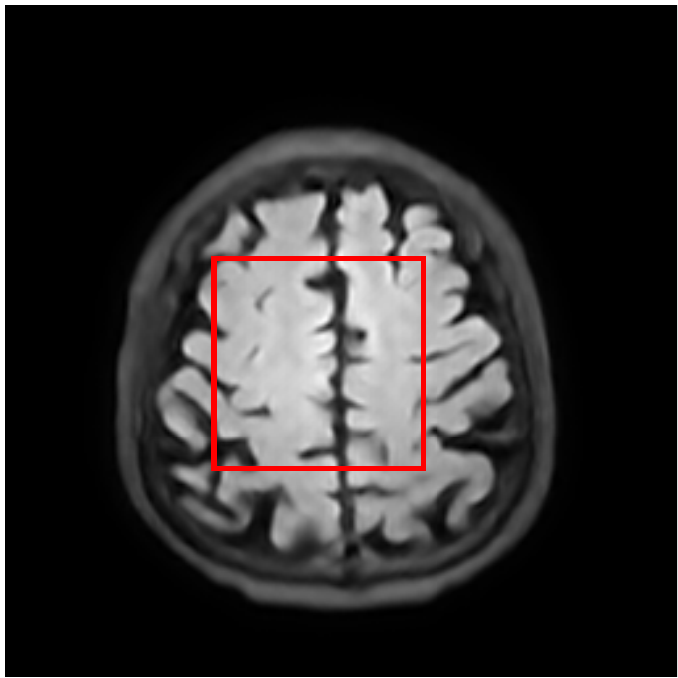} \\
        \includegraphics[scale=0.24]{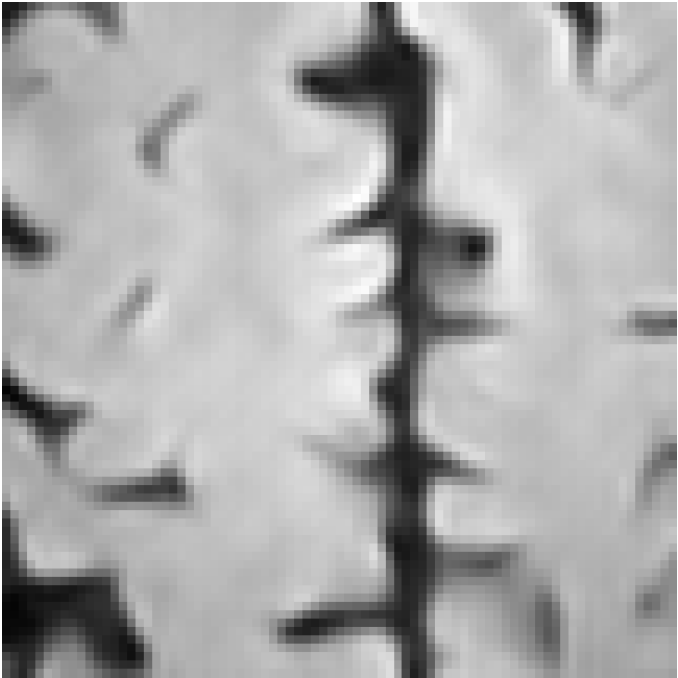}
	\end{minipage}
}
\hfill
    \subfloat[\footnotesize AdaDiff \cite{gungor2023adaptive}]{
	\begin{minipage}[b]{0.155\textwidth}
		\includegraphics[scale=0.24]{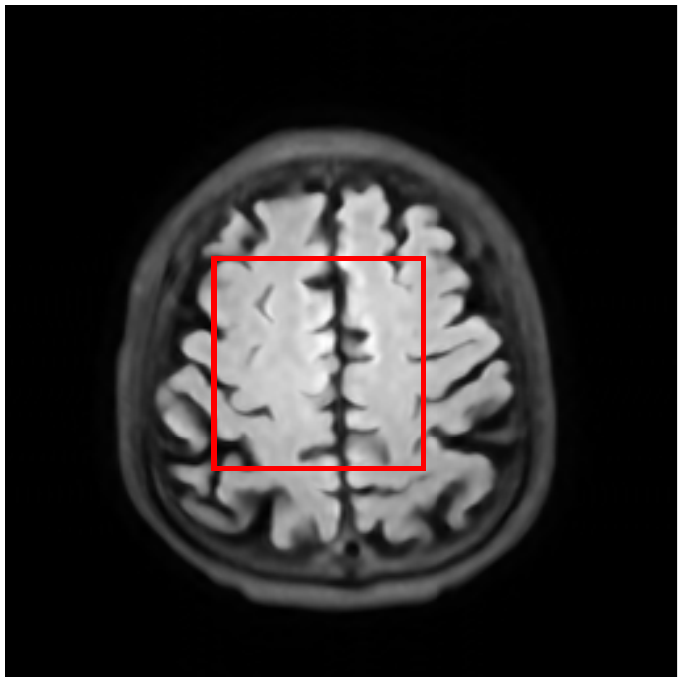} \\
        \includegraphics[scale=0.24]{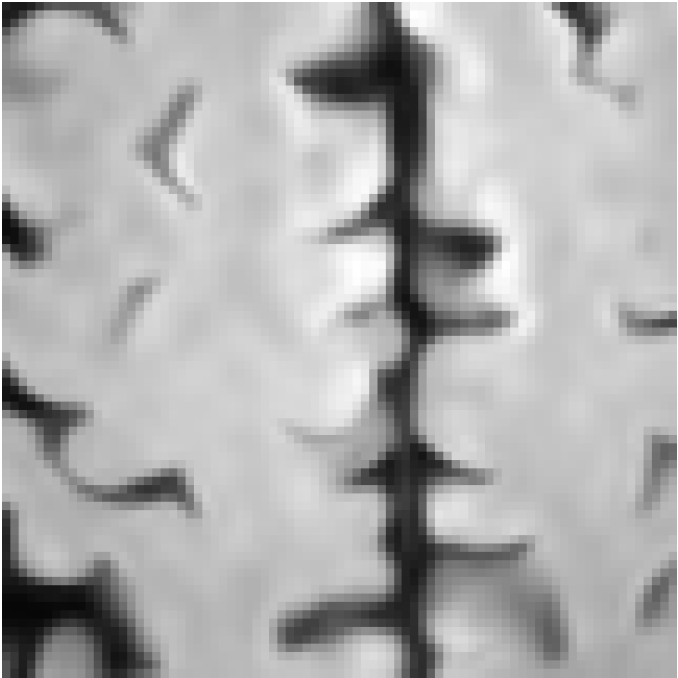}
	\end{minipage}
}
\hfill
    \subfloat[\footnotesize Ours-SCSR ]{
	\begin{minipage}[b]{0.155\textwidth}
		\includegraphics[scale=0.24]{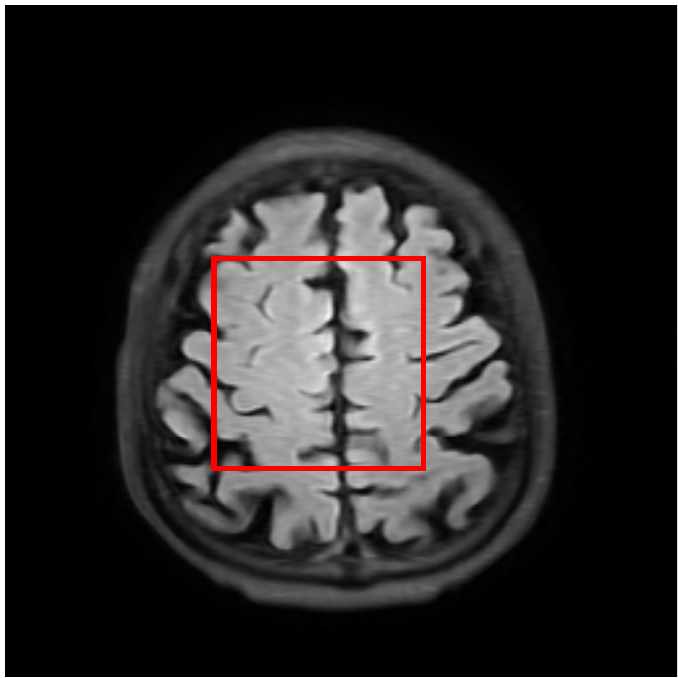} \\
        \includegraphics[scale=0.24]{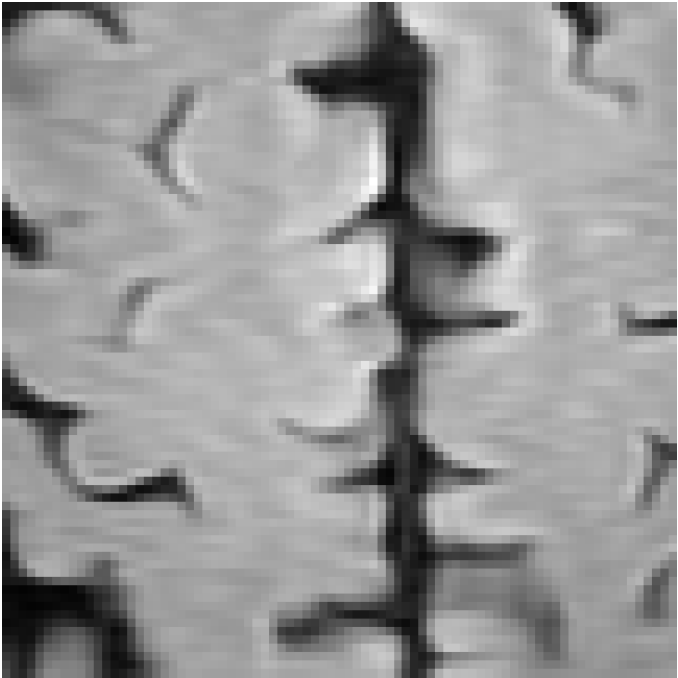}
	\end{minipage}
}
\hfill
    \subfloat[\footnotesize HR]{
	\begin{minipage}[b]{0.155\textwidth}
		\includegraphics[scale=0.24]{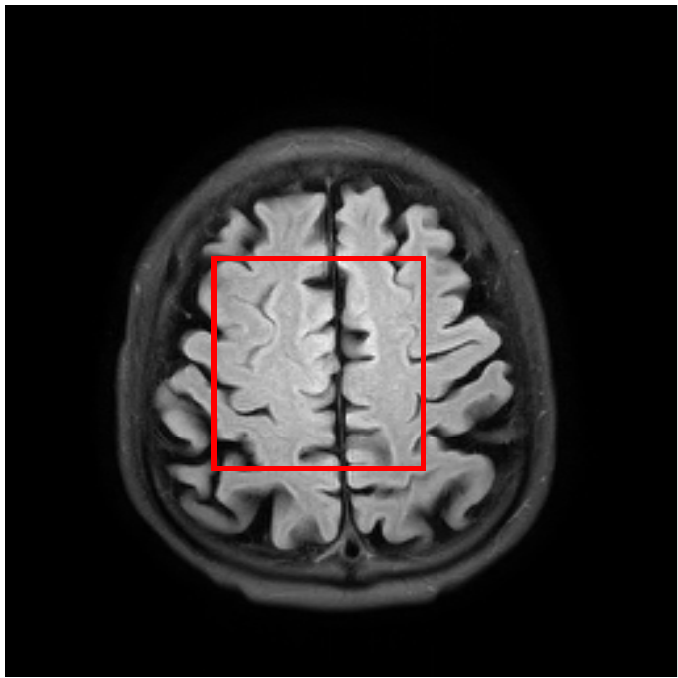}  \\
        \includegraphics[scale=0.24]{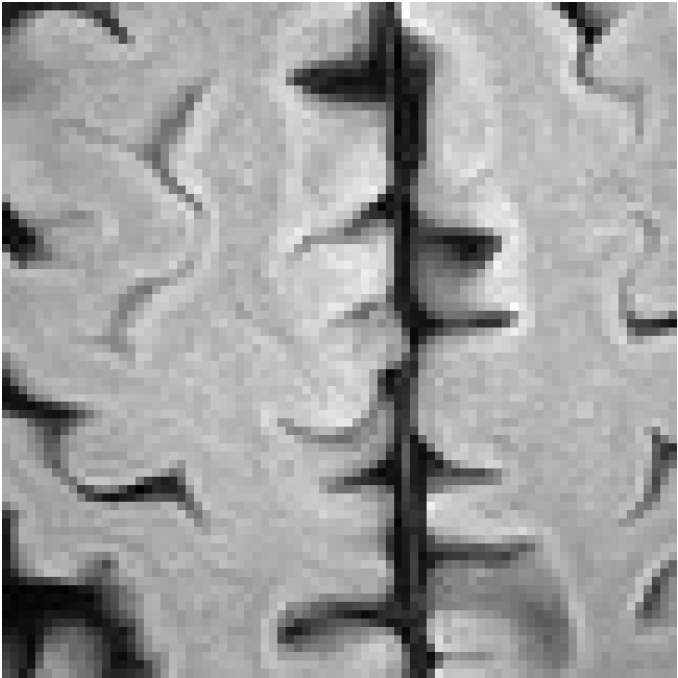}  
	\end{minipage}
}
\caption{Qualitative results of our method versus other DM-based methods under single-contrast MRI super-resolution ($4\times$).}
\label{fig_scsr}
\end{figure*}

Besides, our proposed method can be modified to the Single-Contrast Super-Resolution (SCSR) method by removing the cross-attention Transformer layer.
The visual qualitative results are shown in Figure \ref{fig_scsr}. As can be seen, our method outperforms other DM-based SCSR methods in the SCSR task.

\end{document}